\documentclass[10pt,journal,compsoc]{IEEEtran}
\ifCLASSOPTIONcompsoc
  \usepackage[nocompress]{cite}
\else
  \usepackage{cite}
\fi
\usepackage[pdftex]{graphicx}
\ifCLASSINFOpdf
\else
\fi
\usepackage{algorithm}
\usepackage{algpseudocode}
\usepackage{amsmath}
\usepackage{amssymb}
\usepackage{amsfonts}
\usepackage{multicol}
\usepackage{multirow}
\usepackage{color, graphics}
\usepackage{epsfig}
\usepackage{epstopdf}

\def\blue{\textcolor{black}}

\usepackage{mathrsfs}
\usepackage{epsfig}

\usepackage{caption}

\begin{document}

\title{Adaptive Cost-Sensitive Online Classification}

%
\author{Peilin Zhao, Yifan Zhang, Min Wu, Steven C. H. Hoi, Mingkui Tan, and Junzhou Huang
\IEEEcompsocitemizethanks{
\IEEEcompsocthanksitem \blue{P. Zhao, Y. Zhang and M. Tan are with the South China University of Technology, China. E-mail: peilinzhao@hotmail.com; sezyifan@mail.scut.edu.cn; mingkuitan@scut.edu.cn.}
\IEEEcompsocthanksitem M. Wu is with the Institute for Infocomm Research, Singapore. E-mail: wumin@i2r.a-star.edu.sg.
\IEEEcompsocthanksitem S. C. Hoi is with the Singapore Management University, Singapore. E-mail: chhoi@smu.edu.sg.
\IEEEcompsocthanksitem J. Huang is with Tencent AI Lab, China. Email: joehhuang@tencent.com.
\IEEEcompsocthanksitem \blue{Y. Zhang is the co-first author; M. Tan is the corresponding author.}}
}



\IEEEtitleabstractindextext{
\begin{abstract}
Cost-Sensitive Online Classification has drawn extensive attention in recent years, where the main approach is to directly online optimize two well-known cost-sensitive metrics: (i) weighted sum of sensitivity and specificity; (ii) weighted misclassification cost. However, previous existing methods only considered first-order information of data stream. It is insufficient in practice, since many recent studies have proved that incorporating second-order information enhances the prediction performance of classification models. Thus, we propose a family of cost-sensitive online classification algorithms with adaptive regularization in this paper. We theoretically analyze the proposed algorithms and empirically validate their effectiveness and properties in extensive experiments. Then, for better trade off between the performance and efficiency, we further introduce the sketching technique into our algorithms, which significantly accelerates the computational speed with quite slight performance loss. Finally, we apply our algorithms to tackle several online anomaly detection tasks from real world. Promising results prove that the proposed algorithms are effective and efficient in solving cost-sensitive online classification problems in various real-world domains.
\end{abstract}

}

\maketitle

\IEEEdisplaynontitleabstractindextext

\IEEEpeerreviewmaketitle

\IEEEraisesectionheading{\section{Introduction}\label{Introduction}}
\vspace{-0.1in}
%
%
%
\IEEEPARstart{W}{ith} the rapid growth of datasets, the technologies of machine learning and data mining power many respects of modern society: from content filtering to web searches on social networks, and from goods recommendations to intelligent customer services on e-commerce. Gradually, many real-world large-scale applications make use of a family of techniques called online learning, which has been extensively studied for many years in machine learning and data mining literatures\cite{Rosenblatt1958the,Zhao2011Double,Wang2012Exact,Wu2017Online,Zhao2013Costd,Hoi2018Online}. In general, online learning is a class of efficient and scalable machine learning methods, whose goal is to incrementally learn a model to make correct predictions on a stream of samples. This family of methods provides an opportunity to solve many real-world applications that data arrives sequentially while predictions must be made instantly, such as malicious URL detection\cite{Zhao2013Cost,Ma2011Learning} and portfolio selection\cite{Li2013Confidence}. In addition, online learning is also good at solving large-scale learning tasks, e.g., learning \emph{support vector machine} from billions of data\cite{Shalev-Shwartz2011Pegasos}.

However, although online learning was studied widely, most existing methods were inappropriate to solve cost-sensitive classification problems, because most of them seek performance based on measurable \emph{accuracy} or \emph{mistake rate}, which are obviously cost-insensitive. As a result, these algorithms are difficult to handle numerous real-world problems, where datasets are always class-imbalanced, i.e., the mistake costs of samples are significantly different\cite{Elkan2001The,Veropoulos1999Controling,He2009Learning}. To solve this problem, researchers have suggested to use more meaningful metrics, such as the weighted sum of \emph{sensitivity} and \emph{specificity}\cite{Han2011Data,Brodersen2010the}, and the weighted \emph{misclassification cost}\cite{Elkan2001The,Akbani2004Applying} to replace old ones.
Based on this, many batch classification algorithms are proposed to directly optimize prediction performance for cost-sensitive classification over the past decades\cite{Elkan2001The,Akbani2004Applying}. However, these batch algorithms often suffer from poor scalability and efficiency for large-scale tasks, which make them inappropriate for online classification applications.

Although both \emph{online classification} and \emph{cost-sensitive classification} were studied widely, quite few literatures study cost-sensitive online classification. As results, the Cost-Sensitive Online Classification framework\cite{wang2012cost,wang2014cost} was recently proposed to fill the gap between online learning and cost-sensitive classification. According to this framework, a class of algorithms named as Cost-Sensitive Online Gradient Descend (COG) was proposed to directly optimize predefined cost-sensitive metrics (e.g., weighted sum or weighted misclassification cost) based on online gradient descent technique. Particularly, compared with other traditional online algorithms, COG shows strong empirical performance in solving cost-sensitive online classification problems.

 However, although COG is able to handle the Cost-sensitive online classification tasks, it only takes the first order information of samples (i.e., weighted mean of the gradient). It is obviously insufficient, since many recent studies\cite{Wang2012Exact,Dredze2008Confidence,Crammer2009Adaptive,Crammer2009Exact} have shown that comprehensive consideration with second-order information (i.e., the correlations between features) significantly enhances the performance of online classification.

As an attempt to remedy the limitation of first-order approaches, we propose the Adaptive Regularized Cost-Sensitive Online Gradient Descent algorithms (named ACOG), based on the state-of-the-art Confidence Weighted strategy\cite{Wang2012Exact,Dredze2008Confidence,Crammer2009Adaptive,Crammer2009Exact}. We theoretically analyze their regret bounds\cite{Zinkevich2003Online} and their cost-sensitive metric bounds. Corresponding conclusions confirm the good convergence of ACOG algorithms.

Furthermore, although enjoying the advantage of second-order information, our proposed algorithms are at the cost of higher running time, because the updating process of correlation matrix is time-consuming. As results, it may be inappropriate for some real-world applications with quite high-dimensional datasets. Thus, for better trade off between the efficiency and performance, we further propose an updated version of ACOG algorithms based on sketching techniques\cite{luo2016efficient,woodruff2014sketching,Krummenacher2016Scalable,Wang2014High}, whose running time is linear in the dimensions of samples, just like the first order methods.

Next, we conduct extensive experiments to evaluate the performance and specialities of our proposed algorithms and then apply them to solve online anomaly detection tasks from several real-world domains. Promising results confirm the effectiveness and efficiency of our methods in real-world cost-sensitive online classification problems.

Note that a brief version of this paper had been published in the IEEE ICDM conference\cite{zhao2015cost}. Compared with it, this journal manuscript makes several significant extensions, including (1) an updated variant with sketching methods and some theoretical analyses about its time complexity; (2) an extension of ACOG with an additional loss function and theoretical analyses; (3) more extensive empirical studies to evaluate the proposed algorithms.

The rest of this paper is organized as follows. We present the problem formulation and the proposed algorithms with theoretical analyses in section 2. To save space, we provide theorem proofs and related work in supplemental materials. Next, we propose an efficient version based on sketching techniques in section 3. After that, section 4 empirically evaluates the performance and properties of our algorithms, and section 5 shows an application to real-world anomaly detection tasks. Finally, section 6 concludes the paper.

%
%
\vspace{-0.15in}
\section{Setup and Algorithm} \label{Method}
\vspace{-0.05in}
In this section, we firstly introduce the framework and formulation setting of the Cost-Sensitive Online Classification problem\cite{wang2012cost,wang2014cost}. Then, we present the proposed Adaptively Regularized Cost-Sensitive Online Gradient Descent algorithms (ACOG) in detail.
\vspace{-0.1in}
\subsection{Problem Setting}
\vspace{-0.05in}
Without loss of generality, we consider online binary classification problems here. The main goal is to learn a linear classification model with an updatable predictive vector $w \in \mathbb{R}^d$, based on a stream of training samples $\{(x_1,y_1),(x_2,y_2),...,(x_T,y_T)\}$, where $T$ is the total quantity of samples, $x_t \in \mathbb{R}^d$ is the $d$-dimensional sample at time $t$, and $y_t \in \{-1,1\}$ is the corresponding true class label. In detail, at the $t$-th round of learning, the learner obtains a sample $x_t$ and then predicts its estimated class label $\hat{y}_t = {\rm sign}(w_t^{\top}x_t)$, where $w_t$ is the model predictive vector learnt from the previous $t-1$ samples. Then, the model receives the ground truth of instance $y_t \in \{-1,1\}$, which is the label of true class. If $\hat{y}_t = y_t$, the model makes a correct prediction; otherwise, it makes a mistake and suffers a loss. In the end, the learner updates its predictive vector $w_t$ based on the received painful loss.

For convenience, we define $\mathcal{M} = \{t \ | y_t \neq {\rm sign}(w_t \cdot x_t), \forall t \in [T] \}$ is the mistake index set, $\mathcal{M}_p = \{t \in \mathcal{M}$ and $y_t=+1\}$ is the positive set of mistake index and $\mathcal{M}_n = \{t \in \mathcal{M}$ and $y_t=-1\}$ is the negative one. In addition, we set $M=|\mathcal{M}|$, $M_p= |\mathcal{M}_p|$ and $M_n= |\mathcal{M}_n|$ to denote the number of total mistakes, positive mistakes and negative mistakes. Moreover, we denote the index sets of all positive samples and all negative samples by $\mathcal{I}^p_T = \{i \in [T]| y_i = +1 \}$ and $\mathcal{I}^n_T = \{i \in [T]| y_i = -1 \}$, where $T_p = |\mathcal{I}^p_T|$ and $T_n = |\mathcal{I}^n_T|$ denote the number of positive samples and negative samples.

For performance metrics of this problem, we first assume the positive samples as rare class, i.e., $T_p\leq T_n$. Generally, traditional online classification approaches are eager to maximize accuracy (or minimize mistake rate equivalently):
\begin{equation}
  accuracy = \frac{T-M}{T}. \nonumber
\end{equation}

However, this metric is inappropriate for imbalanced data, because models can easily obtain high accuracy, even simply classifying all imbalanced samples as negative class. So,
a more suitable approach is to measure the $sum$ of weighted $sensitivity$ and $specificity$:
\begin{equation}
  sum =\alpha_p \times \frac{T_p-M_p}{T_p}+ \alpha_n \times \frac{T_n- M_n}{T_n}, \nonumber
\end{equation}
where $\alpha_p, \alpha_n \in [0,1]$ are weight parameters for trade off between sensitivity and specificity, and $\alpha_p + \alpha_n =1$. Note that if $\alpha_p = \alpha_n = 0.5$, the $sum$ metric becomes the famous balanced $accuracy$ metric.

In addition, another metric to measure is the misclassification $cost$ suffered by the model:
\begin{equation}
  cost =c_p \times M_p+ c_n \times M_n, \nonumber
\end{equation}
where $c_p, c_n \in [0,1]$ are misclassification cost parameters for positive and negative instances, and $c_p + c_n =1$. Generally, either the higher of the $sum$ value or the lower of the $cost$ value, the better performance of classification.

Then, we can adjust our focus to maximize $sum$ metric or minimize $cost$ metric. As is known in \cite{wang2012cost,wang2014cost}, both objectives are equivalent to minimizing the following objective:
\begin{equation}\label{1}
  \sum_{y_t=+1}\rho \mathbb{I}_{(y_t w\cdot x_t <0)} +\sum_{y_t=-1} \mathbb{I}_{(y_t w\cdot x_t <0)},
\end{equation}
where $\rho = \frac{\alpha_p T_n}{\alpha_n T_p}$ for weighted $sum$ metric and $\rho = \frac{c_p}{c_n}$ for weighted $cost$ metric.
\vspace{-0.1in}
\subsection{Algorithm}
\vspace{-0.05in}

In this subsection, we present the proposed ACOG algorithms by optimizing the objective from Eq. (1). However, this objective function is non-convex. Thus, to facilitate the optimization, we replace the indicator function with its convex variants (either one of the following two functions):
\begin{equation}\label{2}
  \ell^{I}(w;(x,y))= {\rm max}(0, (\rho * \mathbb{I}_{(y=1)}\small{+}\mathbb{I}_{(y=-1)})\small{-}y(w\cdot x)),
\end{equation}
\begin{equation}\label{3}
  \ell^{II}(w;(x,y))= (\rho * \mathbb{I}_{(y=1)}\small{+}\mathbb{I}_{(y=-1)}) * {\rm max}(0, 1\small{-}y(w\cdot x)).
\end{equation}

For $\ell^{I}(w;(x,y))$, the change of margin yields more "frequent" updates for specific class, compared to the traditional hinge loss; while for $\ell^{II}(w;(x,y))$, the change of the slope causes to more "aggressive" updates for specific class.

Then, our aim is to minimize the regret of learning process\cite{Zinkevich2003Online}, based on either loss functions $\ell^{I}(w;(x,y))$ or $\ell^{II}(w;(x,y))$:
\begin{equation}\label{4}
  Regret := \sum_{t=1}^T \ell(w_t;(x_t,y_t)) - \sum_{t=1}^T \ell(w^*;(x_t,y_t)),
\end{equation}
where $w^* = {\rm arg \ min}_t \sum_{t=1}^T\nabla\ell(w;(x_t,y_t))$. To solve this optimization problem, the cost-sensitive online gradient descent algorithms (COG)\cite{wang2012cost,wang2014cost} were proposed:
 \begin{equation}
  w_{t+1} = w_t- \eta\nabla\ell_t(w_t), \nonumber
\end{equation}
where $\eta$ is the learning rate and $\ell_t(w_t) = \ell(w;(x_t,y_t))$. However, COG algorithms only consider the first order gradient information of the sample stream to update the learner, which is clearly insufficient since many recent studies have shown the significance of incorporating the second order information\cite{Wang2012Exact,Dredze2008Confidence,Crammer2009Adaptive,Crammer2009Exact}. Motivated by this discovery, we propose to introduce adaptive regularization to promote the cost-sensitive online classification.

Let us assume the online model satisfies a multivariate Gaussian distribution, i.e., $w \sim \mathcal{N}(\mu, \Sigma)$ , where $\mu$ is the mean value vector of distribution and $\Sigma$ is the covariance matrix of distribution. Then, we can predict the class label of an sample $x$ based on ${\rm sign}(w^{\top} x)$, when given a definite multivariate Gaussian distribution. In reality, it is more practical to make predictions by simply using distribution mean $\mathbb{E}[w]=\mu$ rather than $w$. So, the rule of model prediction actually adopts ${\rm sign}(\mu^{\top} x)$ in the following. For better understanding, each mean value $\mu_i$ can be regarded as the model's knowledge about the feature $i$; while the diagonal entry of covariance matrix $\Sigma_{i,i}$ is regarded as the confidence of feature $i$. Generally, the smaller of $\Sigma_{i,i}$, the more confidence in the mean weight $\mu_i$ for feature $i$. In addition to diagonal values, other covariance terms $\Sigma_{i,j}$ can be understood as the correlations between two mean weight value $\mu_i$ and $\mu_j$ for feature $i$ and $j$.

Given a multivariate Gaussian distribution, we naturally recast the object functions by minimizing the following unconstraint objective, based on the divergence between empirical distribution and probability distribution:
\begin{equation}
  D_{KL}(\mathcal{N}(\mu, \Sigma)||\mathcal{N}(\mu_t, \Sigma_t)) + \eta \ell_t(\mu) + \frac{1}{2\gamma}x_t^{\top}\Sigma x_t, \nonumber
\end{equation}
where $D_{KL}$ is the Kullback-Leibler divergence, $\eta$ is fitting parameter and $\gamma$ is regularized parameter. Specifically, this objective helps to reach trade off between distribution divergence (first term), loss function (second term) and model confidence (third term). In other word, the objective would like to make the least adjustment at each round to minimize the loss and optimize the confidence of model. To solve this optimization problem, we first depict the Kullback-Leibler divergence explicitly:
\begin{align}
  D_{KL}\big{(}&\mathcal{N}(\mu, \Sigma)||\mathcal{N}(\mu_t, \Sigma_t)\big{)}    \nonumber  \\
  =&\frac{1}{2}{\rm log}\Big{(}\frac{{\rm det}\Sigma_t}{{\rm det}\Sigma}\Big{)}+ \frac{1}{2}{\rm Tr}(\Sigma^{-1}_t \Sigma)+\frac{1}{2}||\mu_t-\mu||_{\Sigma^{-1}_t}^2 -\frac{d}{2}.  \nonumber
\end{align}

However, this optimization function dose not have the closed-form solution. Thus, we change the loss term $\ell_t(\mu)$ with its first order Taylor expansion $\ell_t(\mu_t)+g_t^{\top}(\mu-\mu_t)$, where $g_t=\partial \ell_t (\mu_t)$. Now, we obtain the final optimization objective by removing constant terms:
\begin{equation}\label{5}
  f_t(\mu,\Sigma) = D_{KL}(\mathcal{N}(\mu, \Sigma)||\mathcal{N}(\mu_t, \Sigma_t)) \small{+} \eta g_t^{\top}\mu  \small{+} \frac{1}{2\gamma}x_t^{\top}\Sigma x_t,
\end{equation}
which is much easier to be solved.

A simple method to solve this objective function is to decompose it into two parts depending on $\mu$ and $\Sigma$, respectively. Then, the updates of mean vector $\mu$ and covariance matrix $\Sigma$ can be performed independently:

$\bullet$ \hspace{2ex} Update the mean parameter:
\begin{equation}
  \mu_{t+1} = {\rm arg} \ \min_{\mu} f_t(\mu,\Sigma);   \nonumber
\end{equation}

$\bullet$ \hspace{2ex} If $\ell_t(\mu_t) \neq 0$, update the covariance matrix:
\begin{equation}
  \Sigma_{t+1} = {\rm arg} \  \min_{\Sigma} f_t(\mu,\Sigma).   \nonumber
\end{equation}

For the update of mean parameter, setting the derivative of $\partial_{\mu}f_t(\mu_{t+1},\Sigma)$ as zero will give:
\begin{align}
  \Sigma_t^{-1}(\mu_{t+1}-\mu_t) + \eta g_t=0 \hspace{1ex} \Longrightarrow \hspace{1ex} \mu_{t+1} = \mu_t- \eta \Sigma_t g_t,  \nonumber
\end{align}
while for covariance matrix, setting the derivative of $\partial_{\Sigma}f_t(\mu,\Sigma_{t+1})$ as zero will result in:
\begin{align}
  -\Sigma_{t+1}^{-1}+\Sigma_{t}^{-1}+\frac{x_tx_t^{\top}}{\gamma}=0 \hspace{1ex} \Longrightarrow \hspace{1ex}  \Sigma_{t+1}^{-1} = \Sigma_{t}^{-1}+\frac{x_tx_t^{\top}}{\gamma}, \nonumber
\end{align}
where adopting the Woodbury identity\cite{Horn1985matrix} will give:
\begin{align}\label{6}
  \Sigma_{t+1} = \Sigma_{t} - \frac{\Sigma_{t}x_tx_t^\top\Sigma_{t}}{\gamma+x_t^\top\Sigma_{t}x_t}.
\end{align}

Note that the update of mean parameter $\mu$ relies on the confidence parameter $\Sigma$, we thus propose to update $\mu$ based on the updated covariance matrix $\Sigma_{t+1}$ instead of the old one $\Sigma_t$, which should be more accurate:
\begin{align}\label{7}
  \mu_{t+1} = \mu_t- \eta \Sigma_{t+1} g_t.
\end{align}

This is different from AROW\cite{Crammer2009Adaptive}, where the updating rule of $\mu_t$ based on the old matrix $\Sigma_t$. To intuitively understand this change, let us assume $\Sigma_{t+1}$ as a diagonal matrix. Then, we can find that the updating process actually assigns the updating value of each dimension with different self-adaptive learning rates. So, it is more appropriate to update $\mu$, with the learning rate that considers the current sample. In other words, the more unconfident of the weight, the more aggressive of its updates. Then, we summarize the proposed Adaptive Regularized Cost-Sensitive Online Gradient Descent (ACOG) in Alogrithm 1.

\begin{algorithm} \label{algorithm 1}
    \caption{Adaptive Regularized Cost-Sensitive Online Gradient Descent (ACOG)}
    \begin{algorithmic}[1]
        \Require learning rate $\eta$; regularized parameter $\gamma$; bias parameter $\rho=\frac{\alpha_p*T_n}{\alpha_n*T_p}$ for ``sum`` and $\rho=\frac{c_p}{c_n}$ for ``cost``.
        \Ensure $\mu_1=0$, $\Sigma_1=I$.
        \For{$t = 1 \to T$}
            \State Receive sample $x_t;$
            \State Compute $\ell_t(\mu_t) \small{=}\ell^{*}(\mu_{t};(x_t,y_t)), where \ *\in \{I, II\} ;$
            \If {$\ell_t(\mu_t) > 0$}
                \State $\Sigma_{t+1} = \Sigma_{t} - \frac{\Sigma_{t}x_tx_t^\top\Sigma_{t}}{\gamma+x_t^\top\Sigma_{t}x_t};$
                \State $\mu_{t+1} = \mu_{t} - \eta\Sigma_{t+1}g_t, \ where \ g_t=\partial_{\mu}\ell_t(\mu_t);$
            \Else
                \State $\mu_{t+1} = \mu_{t}, \Sigma_{t+1}= \Sigma_{t};$
            \EndIf
        \EndFor
    \end{algorithmic}
\end{algorithm}

For simplification, we ignore the sample numbers $T$ in the analyses of algorithms efficiency. Thus the time complexity for the updates of $\Sigma_{t+1}$ and $\mu$ are both $\mathcal{O}(d^2)$, so the overall time complexity for ACOG is $\mathcal{O}(d^2)$, which is quite slower than the first order COG algorithms, especially for high-dimensional datasets. To reduce the time complexity, We propose to use the diagonal version of ACOG (i.e., ACOG$_{diag}$), which accelerates the speed of ACOG algorithms to $\mathcal{O}(d)$. Specifically, only a diagonal version $\Sigma_t$ would be maintained and updated at round $t$, which can improve computational efficiency and save memory cost.
%
%

\vspace{1ex}
\textbf{Remark.} In ACOG algorithms, one practical concern is the setting of the value of $\rho$, when optimizing the weighted $sum$ performance. Normally, $\rho$ is denoted as $\rho = \frac{\alpha_pT_n}{\alpha_nT_p}$ for $sum$ metric. However, the value of $T_p$ and $T_n$  might be unknown in advance on real-world online classification tasks. A practical method is to approximate the ratio $\frac{T_n}{T_p}$ according to the empirical distribution of the past training instances, and adaptively update $\frac{T_n}{T_p}$ during the online learning process. In addition, we would empirical examine this problem in experiments.

\vspace{-0.05in}
\subsection{Theoretical Analysis}
\vspace{-0.05in}

In this subsection, we theoretically analyze the proposed ACOG algorithms in terms of two cost-sensitive metrics. Before that, we first prove an important theorem, which gives the regret bounds for algorithms that contributes to later theoretical analyses.

\textbf{Theorem 1.} \ \emph{Let $(x_1,y_1), (x_2,y_2), ..., (x_T,y_T)$ be a sequence of samples, where $x_t \in \mathbb{R}^d, y_t \in \{-1,1\}$. Then for any $\mu \in \mathbb{R}^d$, by setting $\eta=\sqrt{\frac{{\rm max}_{t\leq T}||\mu_t-\mu||^2{\rm Tr}(\Sigma_{T+1}^{-1})}{\gamma{\rm log}(|\Sigma_{T+1}^{-1}|)}}$, the proposed ACOG-I satisfies:}
\begin{align}
  Regret \leq D_{\mu}\sqrt{\gamma{\rm Tr}(\Sigma_{T+1}^{-1}){\rm log}(|\Sigma_{T+1}^{-1}|)}, \nonumber
\end{align}
\emph{where $D_{\mu} = {\rm max}_t||\mu_t-\mu||$}. \emph{In addition, by setting $\eta=\sqrt{\frac{{\rm max}_{t\leq T}||\mu_t-\mu||^2{\rm Tr}(\Sigma_{T+1}^{-1})}{\rho^2\gamma{\rm log}(|\Sigma_{T+1}^{-1}|)}}$, ACOG-II satisfies:}
\begin{equation}
  Regret \leq \rho D_{\mu}\sqrt{\gamma{\rm Tr}(\Sigma_{T+1}^{-1}){\rm log}(|\Sigma_{T+1}^{-1}|)}. \nonumber
\end{equation}

\textbf{Remark. } Let us suppose $||x_t||\leq 1$, it is easy to discover ${\rm Tr}(\Sigma_{T+1}^{-1}) \leq O(T/\gamma)$, which means the regrets of ACOG are in the order of $O(\sqrt{T})$. This order of regret is the optimal, since the loss function is not strongly convex\cite{Abernethy2008Optimal}.

\textbf{Theorem 2.} \ \emph{Under the same assumptions in the Theorem 1, by setting $\rho=\frac{\alpha_pT_n}{\alpha_nT_p}$, for any $\mu \in \mathbb{R}^d$ the ACOG-I satisfies:}
\begin{align}
   sum \geq 1 \small{-} \frac{\alpha_n}{T_n}[\sum_{t=1}^T\ell_t(\mu)\small{+}D_{\mu}\sqrt{\gamma{\rm Tr}(\Sigma_{T+1}^{-1}){\rm log}(|\Sigma_{T+1}^{-1}|)}], \nonumber
\end{align}
\emph{and the ACOG-II satisfies:}
\begin{align}
   sum \geq 1 \small{-} \frac{\alpha_n}{T_n}[\sum_{t=1}^T\ell_t(\mu)\small{+} \rho D_{\mu}\sqrt{\gamma{\rm Tr}(\Sigma_{T+1}^{-1}){\rm log}(|\Sigma_{T+1}^{-1}|)}]. \nonumber
\end{align}

\textbf{Remark. } It is easy to verify that $\sum_{t=1}^T \ell_t(\mu)$ is a convex estimate of $\rho M_p+M_n$ for $\mu$, so $\frac{\alpha_n}{T¡ª_n} \sum_{t=1}^T  \ell_t(\mu)$ is an estimate of $\alpha_p\frac{M_p}{T_p}+\alpha_n\frac{M_n}{T_n}$. In addition, it is worthy noting that $\alpha_n$ cannot be set as zero, since $\rho=\frac{\alpha_pT_n}{\alpha_nT_p}$. However, one limitation here is that we may not know $\frac{T_n}{T_p}$ in advance for a real-world online learning task. To solve this issue, an alternative approach is to consider the $cost$ metric, which does not need the $\frac{T_n}{T_p}$ term in advance because $\rho = \frac{c_p}{c_n}$.

\textbf{Theorem 3.} \ \emph{Under the same assumptions in the Theorem 1, by setting $\rho=\frac{c_p}{c_n}$, for any $\mu \in \mathbb{R}^d$, the ACOG-I satisfies:}
\begin{align}
   cost \leq c_n[\sum_{t=1}^T\ell_t(\mu)\small{+}D_{\mu}\sqrt{\gamma{\rm Tr}(\Sigma_{T+1}^{-1}){\rm log}(|\Sigma_{T+1}^{-1}|)}], \nonumber
\end{align}
\emph{and the ACOG-II satisfies:}
\begin{align}
   cost \leq c_n[\sum_{t=1}^T\ell_t(\mu)\small{+} \rho D_{\mu}\sqrt{\gamma{\rm Tr}(\Sigma_{T+1}^{-1}){\rm log}(|\Sigma_{T+1}^{-1}|)}]. \nonumber
\end{align}

\textbf{Remark. } For the $cost$ metric, $\sum_{t=1}^T \ell_t(\mu)$ is a convex estimate of $\frac{c_p}{c_n}M_p+M_n$, and so $c_n\sum_{t=1}^T \ell_t(\mu)$ is an estimate of $c_pM_p+c_nM_n$. Moreover, one should note that $c_n$ cannot be set as zero because of $\rho=\frac{c_p}{c_n}$.

\vspace{-0.1in}
\section{Enhanced Algorithm with Sketching} \label{Method}
\vspace{-0.05in}
As mentioned above, the time complexity of ACOG is $O(d^2)$ and its diagonal version is $O(d)$. However, the diagonal ACOG cannot enjoy the correlation information between different dimensions of samples. When instances have low \emph{effective rank}, the regret bound of diagonal ACOG may be much worse than its full-matrix version due to the lack of enough dependance on the data dimensionality\cite{Krummenacher2016Scalable}. Unfortunately, real-world high-dimensional datasets are common to have such low rank settings with abundant correlations between features. So for those real-world datasets, it is more appropriate to choose the full matrix version. However, ACOG has one limitation that it will take a large amount of time, when receiving quite high-dimensional samples. To better balance the performance and the running time, we propose an enhanced version of our algorithms, named Sketched Adaptive Regularized Cost-Sensitive Online Gradient Descent (SACOG).
\vspace{-0.1in}
\subsection{Sketched Algorithm}
\vspace{-0.05in}
In this subsection, we will present the enhanced version of ACOG via \textbf{Oja's sketch method}\cite{luo2016efficient,Oja1982Simplified,oja1985stochastic}, which is designed to accelerate computation efficiency when the second order matrix of sequential data is low rank.

In detail, the main idea of SACOG is to approximate the second covariance matrix $\Sigma$ by a small number of carefully selective directions, called as a $sketch$.

According to Eq. (6-7), we know the updating rule of model parameter $\mu$:
\begin{align}
  \mu_{t+1}= \mu_{t}-\eta\Sigma_{t+1}g_{t}, \nonumber
\end{align}
and the incremental formula of covariance matrix:
\begin{align}
  \Sigma_{t+1}^{-1}= \Sigma_{t}^{-1}+ \frac{x_t x_t^{\top}}{\gamma},\nonumber
\end{align}
which can be expressed in another way:
\begin{align}\label{8}
  \Sigma_{t+1}^{-1}= I_d + \sum_{i=1}^{t}\frac{x_i x_i^{\top}}{\gamma},
\end{align}
where $d$ is the dimensionality of instance.

Let $X_t\in \mathbb{R}^{t\times d}$ be a matrix, whose $t$-th row is $\hat{x}_t^{\top}$, where we define $\hat{x}_t = \frac{x_t}{\sqrt{\gamma}}$  as the $to$-$sketch$ $vector$. Then, the Eq.(8) can be written as:
\begin{align}
  \Sigma_{t+1}^{-1} =I_d + X_t^{\top}X_t.\nonumber
\end{align}

Now, we define $S_t\in \mathbb{R}^{m\times d}$ as sketch matrix to approximate $X_t$, where the sketch size $m\ll d$ is a small constant.

When $m$ is chosen so that $X_t^{\top}X_t$ can be approximated by $S_t^{\top}S_t$ well, the Eq.(8) can be redefined as:
\begin{align}
  \Sigma_{t+1}^{-1} = I_d + S_t^{\top}S_t.  \nonumber
\end{align}

Then by the Woodbury identity\cite{Horn1985matrix}, we have:
\begin{align}\label{9}
  \Sigma_{t+1} = I_d - S_t^{\top}H_t S_t ,
\end{align}
where $H_t = (I_m + S_tS_t^{\top})^{-1} \in \mathbb{R}^{m\times m}$. Then, we rewrite the updating rule of parameter $\mu$:
\begin{align}\label{10}
  \mu_{t+1}= \mu_{t}-\eta(g_{t}-S_t^{\top}H_t S_t g_t).
\end{align}

Based on above, we summarize Sketched ACOG in Algorithm 2.
\begin{algorithm} \label{algorithm 2}
    \caption{Sketched Adaptive Regularized Cost-Sensitive Online Gradient Descent (SACOG)}
    \begin{algorithmic}[1]
        \Require learning rate $\eta$; regularized parameter $\gamma$; sketch size $m$; bias $\rho=\frac{\alpha_p*T_n}{\alpha_n*T_p}$ for ``sum`` and $\rho=\frac{c_p}{c_n}$ for ``cost``.
        \Ensure $\mu_1=0$, sketch$(S_0,H_0)\leftarrow$ \textbf{SketchInit}$(m)$.
        \For{$t = 1 \to T$}
            \State Receive sample $x_t;$
            \State Compute $\ell_t(\mu_t) \small{=}\ell^{*}(\mu_{t};(x_t,y_t)), where \ *\in \{I, II\} ;$
            \State Compute the $t$-$sketch$ $vector$ $\hat{x}_t= \frac{x_t}{\sqrt{\gamma}};$
            \State $(S_t,H_t) \leftarrow$ \textbf{SketchUpdate}$(\hat{x});$
            \If {$\ell_t(\mu_t) > 0$}

                \State $\mu_{t+1}= \mu_{t}\small{-}\eta(g_{t}\small{-}S_t^{\top}H_t S_t g_t), \ where \ g_t \small{=}\partial_{\mu}\ell_t(\mu_t);$
            \Else
                \State $\mu_{t+1} = \mu_{t}.$
            \EndIf
        \EndFor
    \end{algorithmic}
\end{algorithm}

Then we discuss how to maintain the matrices $S_t$ and $H_t$ efficiently via sketching technique, where we compute eigenvalues and eigenvectors of sequential data through online gradient descent with $to$-$sketch$ $vector$ $\hat{x}_t$ as input.

In detail, let the diagonal matrix $\Lambda_t \in \mathbb{R}^{m\times m}$ contain the approximated eigenvalues and $V_t \in \mathbb{R}^{m\times d}$ be the estimated eigenvectors at round $t$. According to Oja's algorithm\cite{Oja1982Simplified,oja1985stochastic}, the updating rules of $\Lambda_t$ and $V_t$ are defined as:
\begin{align}\label{11-2}
  \Lambda_{t}=&(I_m - \Gamma_{t})\Lambda_{t-1} + \Gamma_{t} diag\{V_{t-1}\hat{x}_t\}^2 , \\
  &V_{t} \xleftarrow{orth} V_{t-1} + \Gamma_{t}V_{t-1} \hat{x}_t \hat{x}_t^{\top},
\end{align}
where learning rate $\Gamma_t = \frac{1}{t}I_m \in \mathbb{R}^{m\times m}$ is a diagonal matrix, and $\xleftarrow{orth}$ represents an orthonormalizing step\footnote{For sake of simplicity, $V_t+ \Gamma_{t+1}V_t\hat{x}_t\hat{x}_t^{\top}$ is assumed as full rank with rows all the way, so that the $\xleftarrow{orth}$ operation always keeps the same dimensionality of $V_t$.}. Then, the sketch matrices can be obtained by:
\begin{align}\label{13}
  &S_t = (t\Lambda)^{\frac{1}{2}}V_t, \\
H_t = diag\{&\frac{1}{1+t\Lambda_{1,1}},...,\frac{1}{1+t\Lambda_{m,m}}\}.   \nonumber
\end{align}

Since the rows of $S_t$ are always orthogonal, $H_t$ is an efficiently maintainable diagonal matrix all the way. We summarize the Oja's sketching technique in Algorithm 3.

\begin{algorithm} \label{algorithm 3}
    \caption{Oja's Sketch for SACOG}
    \begin{algorithmic}
        \Require  $m$, $\hat{x}$ and stepsize matrix $\Gamma_t$.
        \State \hspace{-2.7ex} \textbf{Internal State} \ $t$, $\Lambda$, $V$ and $H.$
        \State \hspace{-2.2ex}\textbf{SketchInit}$(m)$
        \State \hspace{1ex}1: Set $t=0, S = 0_{m \times d},  H = I_m , \Lambda = 0_{m\times m}$
        \State \hspace{3ex} and $V$ to any $ m \times d$ matrix with orthonormal rows$;$
        \State \hspace{1ex}2: Return $(S,H).$
        \\

        \State \hspace{-2.6ex} \textbf{SketchUpdate}$(\hat{x})$
        \State \hspace{1ex}1: Update $t \leftarrow t+1;$
        \State \hspace{1ex}2: Update $\Lambda=(I_m - \Gamma_{t})\Lambda + \Gamma_{t} diag\{V\hat{x}\}^2 ;$
        \State \hspace{1ex}3: Update $V \xleftarrow{orth} V + \Gamma_{t}V \hat{x} \hat{x}^{\top};$
        \State \hspace{1ex}4: Set $S = (t\Lambda)^{\frac{1}{2}}V;$
        \State \hspace{1ex}5: Set $H = diag\{\frac{1}{1+t\Lambda_{1,1}},...,\frac{1}{1+t\Lambda_{m,m}}\};$
        \State \hspace{1ex}6: Return $(S,H).$
    \end{algorithmic}
\end{algorithm}

\textbf{Remark.} The time complexity of this algorithm is $O(m^2d)$ per round because of the orthonormalizing operation, and one can update the sketch every $m$ rounds to improve time complexity to $O(md)$\cite{Hardt2014The}. Another concern is the regret guarantee, which is not clear now because existing analysis for Oja's algorithm is only for the stochastic situation\cite{luo2016efficient}. However, SACOG provides good empirical performance.
\vspace{-0.25in}
\subsection{Sparse Sketched Algorithm}
\vspace{-0.05in}
However, even via sketching, SACOG algorithms are still quite slower than most online first order methods, because they cannot enjoy the sparse information of samples while first-order algorithms can. The question is that in many real-world applications, the samples are normally high sparse that the number of nonzero elements satisfies $||x||_0 \leq s$ with some small constants $s \ll d$.

As results, many first order methods can enjoy a per-round running time depending on $s$ rather than $d$. But for SACOG, even when samples are sparse, the sketch matrix $S_t$ still becomes dense quickly, because of the orthonormalizing updating of $V_t$. For this reason, the updates of $\mu_t$ cannot enjoy the sparsity of samples. To handle this question, we propose an enhanced sparse version of SACOG to achieve a purely sparsity-dependent time cost.

The main idea is that we adjust the formulations of eigenvector $V_t$ and predictive vector $\mu_t $, so that the updates of them are always sparse. In detail, there are two key modifications for SACOG: \ (1) The Eigenvectors $V_t$ are modified as $V_t = F_tZ_t$, where $F_t \in \mathbb{R}^{m\times m}$ is an orthonormalizing matrix so that $F_tZ_t$ is orthonormal, and $Z_t \in \mathbb{R}^{m\times d}$ is a sparsely updatable direction. \ (2) The weights $\mu_t$ fall into two parts $\mu_t = w_t+Z_{t-1}^{\top}b_t$, where $w_t\in \mathbb{R}^{d}$ captures the sparsely updating weights on the complementary subspace, and $b_t \in \mathbb{R}^{m}$ captures the weights on the subspace form $V_{t-1}$ (same as $Z_{t-1}$). Then, we describe how to sparsely update two weight parts $w_t$ and $b_t$. Firstly, from Eq. (13), we know $S_t = (t\Lambda)^{\frac{1}{2}}V_t= (t\Lambda)^{\frac{1}{2}}F_tZ_t$. Then, we have:
\begin{footnotesize}
    \begin{align}
        \mu_{t+1} =& \mu_t - \eta (I_d -S_t^{\top}H_tS_t)g_t  \nonumber \\
        = &w_t +Z_{t-1}^{\top}b_t - \eta g_t+  \eta Z_t^{\top} F_t^{\top}(t\Lambda H_t)F_t Z_t g_t \nonumber \\
        = &[ \underbrace{w_t - \eta g_t- (Z_t-Z_{t-1})^{\top}b_t}_{w_{t+1}}] + Z_t^{\top}[ \underbrace{b_t + \eta F_t^{\top}(t\Lambda H_t)F_t Z_t g_t}_{b_{t+1}}]. \nonumber
    \end{align}
\end{footnotesize}

According to this, we can define the updating rule of $w_t$:
\begin{align}\label{14}
        w_{t+1} = &w_t - \eta g_t- (Z_t-Z_{t-1})^{\top}b_t \nonumber \\
        =&w_t- \eta g_t-\hat{x}_t \delta_t^{\top} b_t,
\end{align}
where $Z_t \small{=} Z_{t-1}\small{+}\delta_t \hat{x}_t^{\top}$, and define the updating rule of $b_t$:
\begin{align}\label{15}
    b_{t+1} = b_t+ \eta  F_t^{\top}(t\Lambda_t H_t)F_t Z_t g_t.
\end{align}

Based on above, we summarize the sparse SACOG in Algorithm 4.
\begin{algorithm} \label{algorithm 4}
    \caption{Sparse Sketched Adaptive Regularized Cost-Sensitive Online Gradient Descent (SACOG)}
    \begin{algorithmic}[1]
        \Require learning rate $\eta$; regularized parameter $\gamma$; sketch size $m$; bias $\rho=\frac{\alpha_p*T_n}{\alpha_n*T_p}$ for ``sum`` and $\rho=\frac{c_p}{c_n}$ for ``cost``.
        \Ensure $w_1=0_{d\times 1}$, $b_1=0_{m \times 1};$
        \Ensure Sketch $(\Lambda_0, F_0, Z_0 ,H_0)\leftarrow$ \textbf{SketchInit}$(m);$
        \For{$t = 1 \to T$}
            \State Receive sample $x_t;$
            \State Compute $\ell_t(\mu_t) \small{=}\ell^{*}(\mu_{t};(x_t,y_t)), where \ *\in \{I, II\} ;$
            \State Compute the $t$-$sketch$ $vector$ $\hat{x}_t= \frac{x_t}{\sqrt{\gamma}};$
            \State $(\Lambda_t, F_t, Z_t ,H_t,\delta_t) \leftarrow$ \textbf{SketchUpdate}$(\hat{x});$
            \If {$\ell_t(\mu_t) > 0$}
                \State $w_{t+1} = w_t -\eta g_t -\hat{x}_t \delta_t^{\top} b_t;$
                \State $b_{t+1} = b_t + \eta F_t^{\top}(t\Lambda_t H_t)F_tZ_tg_t;$
                \State $\mu_{t+1}= w_{t+1} + Z_t^{\top}b_{t+1};$
            \Else
                \State $\mu_{t+1} = \mu_{t}$, \ $w_{t+1}= w_{t}$, \ $b_{t+1}= b_t.$
            \EndIf
        \EndFor
    \end{algorithmic}
\end{algorithm}

Next, we describe how to update $\Lambda_t$, $F_t$ and $Z_t$. First, we rewrite the updating rule of eigenvalues $\Lambda_t$ from Eq. (11):
\begin{align}\label{16}
    \Lambda_{t}=(I_m - \Gamma_{t})\Lambda_{t-1} + \Gamma_{t} diag\{F_{t-1}Z_{t-1}\hat{x}_t\}^2.
\end{align}

Then from Eq. (12), we have:
\begin{align}\label{17}
        F_tZ_t \xleftarrow{orth} & F_{t-1}Z_{t-1} + \Gamma_tF_{t-1}Z_{t-1}\hat{x}_t \hat{x}_t^{\top},  \nonumber \\
        = &F_{t-1}(Z_{t-1} + F_{t-1}^{-1}\Gamma_tF_{t-1}Z_{t-1}\hat{x}_t\hat{x}_t^{\top}).
\end{align}
Here, $Z_t = Z_{t-1}+\delta_t \hat{x}_t^{\top}$, where $\delta_t = F_{t-1}^{-1}\Gamma_tF_{t-1}Z_{t-1}\hat{x}_t$ (note that $F_t$ is always invertible because of Footnote 1). Now, it is easy to note that $Z_t- Z_{t-1}$ is a sparse rank-one matrix, which represents the update of $w_t$ is efficient.

Finally, for the update of $F_t$ so that $F_tZ_t$ is also orthonormalizing, we apply the Gram-Schmidt algorithm to $F_{t-1}$ in a Banach space, where the inner product is defined as $ \langle a,b \rangle=a^{\top}K_tb$ and $K_t=Z_tZ_t^{\top}$ is the Gram matrix (See Algorithm 6). Then, we can update $K_t$ efficiently based on the update of $Z_t$:
 \begin{align}\label{18}
        K_t = & Z_tZ_t^{\top},  \nonumber \\
        = &(Z_{t-1}+\delta_t\hat{x}_t^{\top})(Z_{t-1}+\delta_t\hat{x}_t^{\top})^{\top},  \nonumber \\
        = &K_{t-1} +Z_{t-1} \hat{x}_t \delta_t^{\top}+ \delta_t\hat{x}_t^{\top}Z_{t-1}^{\top}+ \delta_t \hat{x}_t^{\top} \hat{x}_t \delta_t^{\top}.
\end{align}

we summarize the Sparse Oja's algorithm for SACOG in Algorithm 5.

\begin{algorithm} \label{algorithm 5}
    \caption{Sparse Oja's Sketch for SACOG}
    \begin{algorithmic}
        \Require $m$, $\hat{x}$ and stepsize matrix $\Gamma_t$.
        \State \hspace{-2.7ex} \textbf{Internal State} \ $t$, $\Lambda$, $F$, $Z$, $K$ and $H.$
        \State \hspace{-2.8ex} \textbf{SketchInit}$(m)$
        \State \hspace{1ex}1: Set $t=0, F = K= H = I_m , \Lambda = 0_{m\times m}$
        \State \hspace{3ex} and $Z$ to any $ m \times d$ matrix with orthonormal rows$;$
        \State \hspace{1ex}2: Return $(\Lambda,F,Z,H).$ \\

        \State \hspace{-2.7ex} \textbf{SketchUpdate}$(\hat{x})$
        \State \hspace{1ex}1: Update $t \leftarrow t+1;$
        \State \hspace{1ex}2: $\Lambda=(I_m - \Gamma_{t})\Lambda+ \Gamma_{t} diag\{F Z\hat{x}\}^2;$
        \State \hspace{1ex}3: Set $\delta = F^{-1}\Gamma_tFZ        \hat{x}^{\top};$
        \State \hspace{1ex}4: $K \leftarrow K +Z \hat{x} \delta^{\top}+ \delta\hat{x}^{\top}Z^{\top}+ \delta \hat{x}^{\top} \hat{x} \delta^{\top};$
        \State \hspace{1ex}5: $Z \leftarrow Z + \delta \hat{x}^{\top};$
        \State \hspace{1ex}6: $(L,Q) \leftarrow$ Decompose$(F, K)$,
        \State \hspace{3ex} where $LQZ=FZ$ and $QZ$ is orthogonal$;$
        \State \hspace{1ex}7: Set $F=Q;$
        \State \hspace{1ex}8: Set $H = diag\{\frac{1}{1+t\Lambda_{1,1}},...,\frac{1}{1+t\Lambda_{m,m}}\};$
        \State \hspace{1ex}9: Return $(\Lambda, F, Z ,H,\delta).$
    \end{algorithmic}
\end{algorithm}

\begin{algorithm} \label{algorithm 6}
    \caption{Decompose$(F, K)$}
    \begin{algorithmic}[1]
        \Require $F \in \mathbb{R}^{m\times m}$ and Gram matrix $K=ZZ^{\top} \in \mathbb{R}^{m\times m};$
        \Ensure $L= 0_{m \times m}$ and $Q = 0_{m \times m};$
        \For{$i = 1 \to m$}
            \State Let $f^{\top}$ be the $i$-th row of $F;$
            \State Compute $\alpha = QKf$, $\beta = f - Q^{\top}\alpha$ and $c=\sqrt{\beta^{\top}K\beta};$
            \If {$c \neq 0$}
                \State Insert $\frac{1}{c}\beta^{\top}$ to the $i$-th row of $Q;$
            \EndIf
            \State Set the $i$-th entry of $\alpha$ to be $c;$
            \State Insert $\alpha$ to the $i$-th row of $L;$
            \EndFor
            \State Delete the all-zero columns of $L$ and all-zero rows of $Q;$
            \State Return $(L,Q).$

    \end{algorithmic}
\end{algorithm}

\textbf{Remark.} Note that the most time-consuming step is the update of $F_t$ (See line 3 in Algorithm 6), which is $O(m^3)$. In addition, the time complexity for update of $w_t$ is $O(ms)$ and that of $b_t$ is $O(m^2 + ms)$. Thus, the overall time complexity of sparse ACOG per round is $O(m^3 + ms)$. One can improve the running time per round to $O(m^2+ms)$ by only updating the sketch every $m$ rounds. To the best of our knowledge, this is the first time that sparse Oja's sketch method is applied to the cost-sensitive online classification problem.

\vspace{-0.1in}
\section{Experiments}
\vspace{-0.05in}
In this section, we first evaluate the performance and characteristics of the original algorithms (i.e., ACOG and its diagonal version). After that, we further evaluate the effectiveness and efficiency of sketched variants (i.e., SACOG and its sparse version).

\vspace{-0.1in}
\subsection{Experimental Testbed and Setup}
\vspace{-0.05in}
At the beginning, we compare ACOG and its diagonal variant, with several famous standard online learning algorithms as follows: (1) Perceptron Algorithm\cite{Rosenblatt1958the,Freund1999Large}; (2) Relaxed Online Maximum Margin Algorithm\cite{Li2000The} (``ROMMA``); (3) Passive-Aggressive algorithm\cite{Crammer2006Online} (``PA-I`` and ``CPA-PB``); (4) Perceptron Algorithm with Uneven Margin\cite{Li2002The} (``PAUM``); (5) Adaptive Regularization of Weight Vector\cite{Crammer2009Adaptive} (``AROW``); (6) Cost-Sensitive Online Gradient Descent\cite{wang2012cost,wang2014cost} (``COG-I`` and ``COG-II``), from which ACOG was derived. All algorithms were evaluated on 4 benchmark datasets as listed in Table 1, which are obtained from LIBSVM\footnote{https://www.csie.ntu.edu.tw/~cjlin/libsvmtools/datasets/}.

For data preprocessing, all samples are normalized by $x_t \leftarrow \frac{x_t}{\|x_t\|_2}$, which is extensively used in online learning, since samples are obtained sequentially.

For a valid comparison, all algorithms used the same experimental settings. We set $\alpha_p=\alpha_n=0.5$ for $sum$, and $c_p=0.9$ and $c_n=0.1$ for $cost$. The value of $\rho$ was set to $\frac{\alpha_p*T_n}{\alpha_n*T_p}$ for $sum$ and $\frac{c_p}{c_n}$ for $cost$, respectively. For CPA$_{PB}$ algorithm, $\rho(-1,1)$ was set to 1, and $\rho(1,-1)$ was $\rho$. For PAUM, the uneven margin was set to $\rho$.  In addition, the parameter of $C$ for PA-I, learning rate $\lambda$ for COG and learning rate $\eta$ for all our proposed algorithms were selected from $[10^{-5},10^{-4},...,10^{5}]$. The regularized parameter $\gamma$ for AROW and all our algorithms were set as 1.

On each dataset, experiments were conducted over 20 random permutations of instances. Results are reported through the average performance of 20 runs and evaluated by 4 metrics: $sensitivity, \ specificity$, the weighted $sum$ of sensitivity and specificity, and the weighted $cost$ of misclassification. All algorithms were implemented in MATLAB on a 3.40GHz Winodws machine.
\vspace{-0.05in}
\begin{table}[h]
	\caption{\small{List of Binary Datasets in Experiments}}
    \vspace{-0.1in}
    \begin{center}
	\begin{tabular}{|l|c|c|c|}\hline
		Dataset &\#Examples & \#Features & \#Pos:\#Neg  \\\hline \hline
        covtype & 581012 & 54 & 1:1  \\
        german&1000&24&1:2.3\\
        a9a&48842&123&1:3.2\\
        ijcnn1&141691&22&1:9.4\\\hline
	\end{tabular}
    \end{center}

\end{table}

\vspace{-0.3015in}
\subsection{Evaluation with Sum Metrics}
\subsubsection{Evaluation of Weighted Sum Performance}
\vspace{-0.05in}

First of all, we aim to evaluate the weighted $sum$ performance of ACOG and its diagonal version. Table 2 summaries the experimental results on 4 datasets, and Fig. 1 shows the development of online average $sum$ performance on all datasets, respectively.

From Fig. 1 and Table 2, we can find that second-order algorithms (i.e., our proposed ACOG algorithms and regular AROW algorithm) outperform first-order algorithms on almost all datasets. This confirms the effectiveness of introducing the second order information into online classification. At the same time, ACOG algorithms significantly outperform all other online learning algorithms including AROW on all datasets, which confirms the superiority of combination between the second order information and cost-sensitive online classification.

Then by evaluating both $sensitivity$ and $specificity$ metrics, our proposed algorithms not only achieve the best $sensitivity$ on all datasets, but also produce a fairly good $specificity$ for most datasets. This implies the proposed ACOG approaches are effective in improving prediction accuracy for rare class samples.

Moreover, while ACOG$_{diag}$ algorithms achieve smaller $sum$ than ACOG algorithms, their computations are faster. This indicates the diagonal ACOG algorithms have ability to balance the effectiveness and efficiency.

\begin{figure}
    \begin{minipage}{0.5\linewidth}
      \centerline{\includegraphics[width=4.5cm]{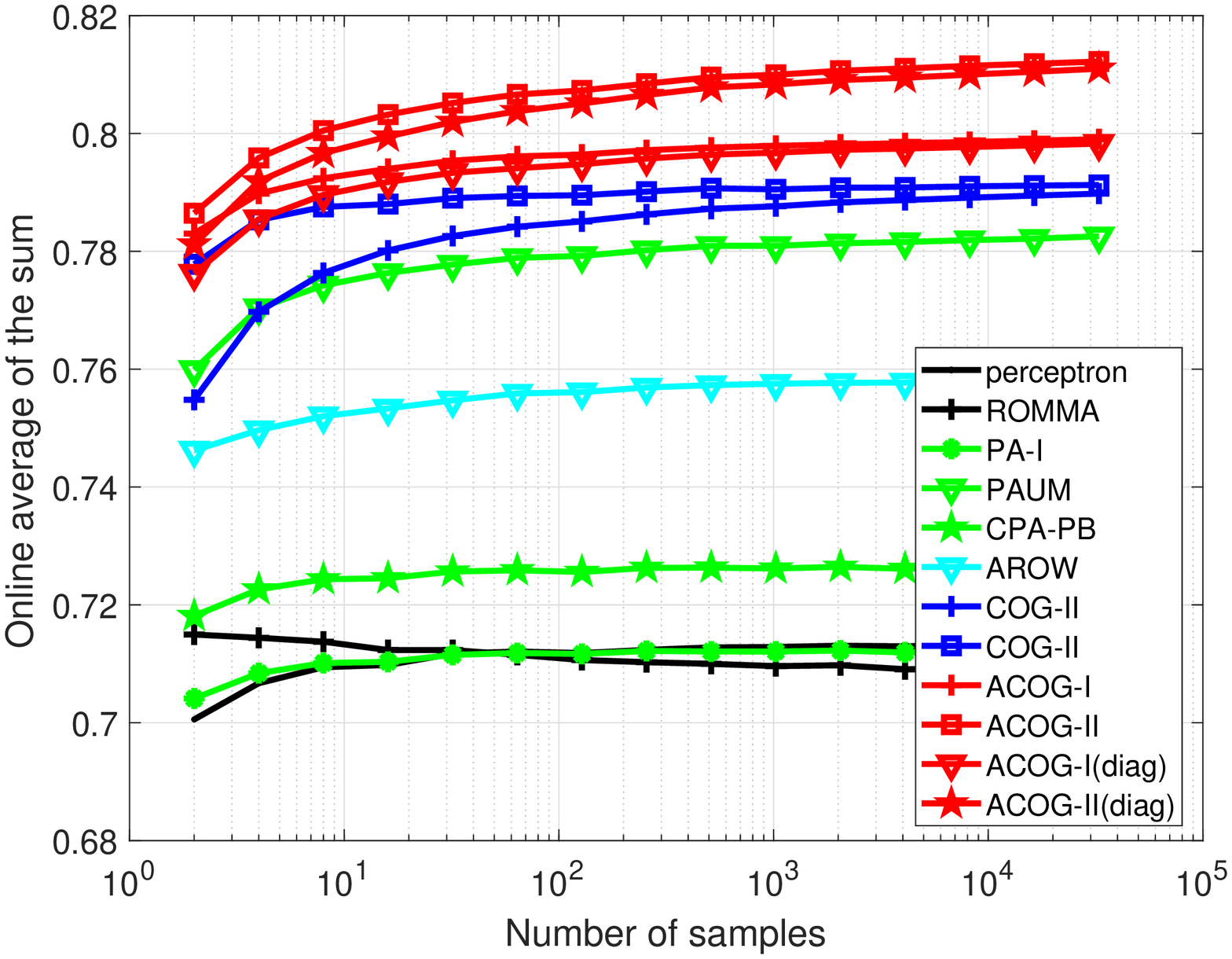}}
      \centerline{(a) a9a}
    \end{minipage}
    \hfill
    \begin{minipage}{0.5\linewidth}
      \centerline{\includegraphics[width=4.5cm]{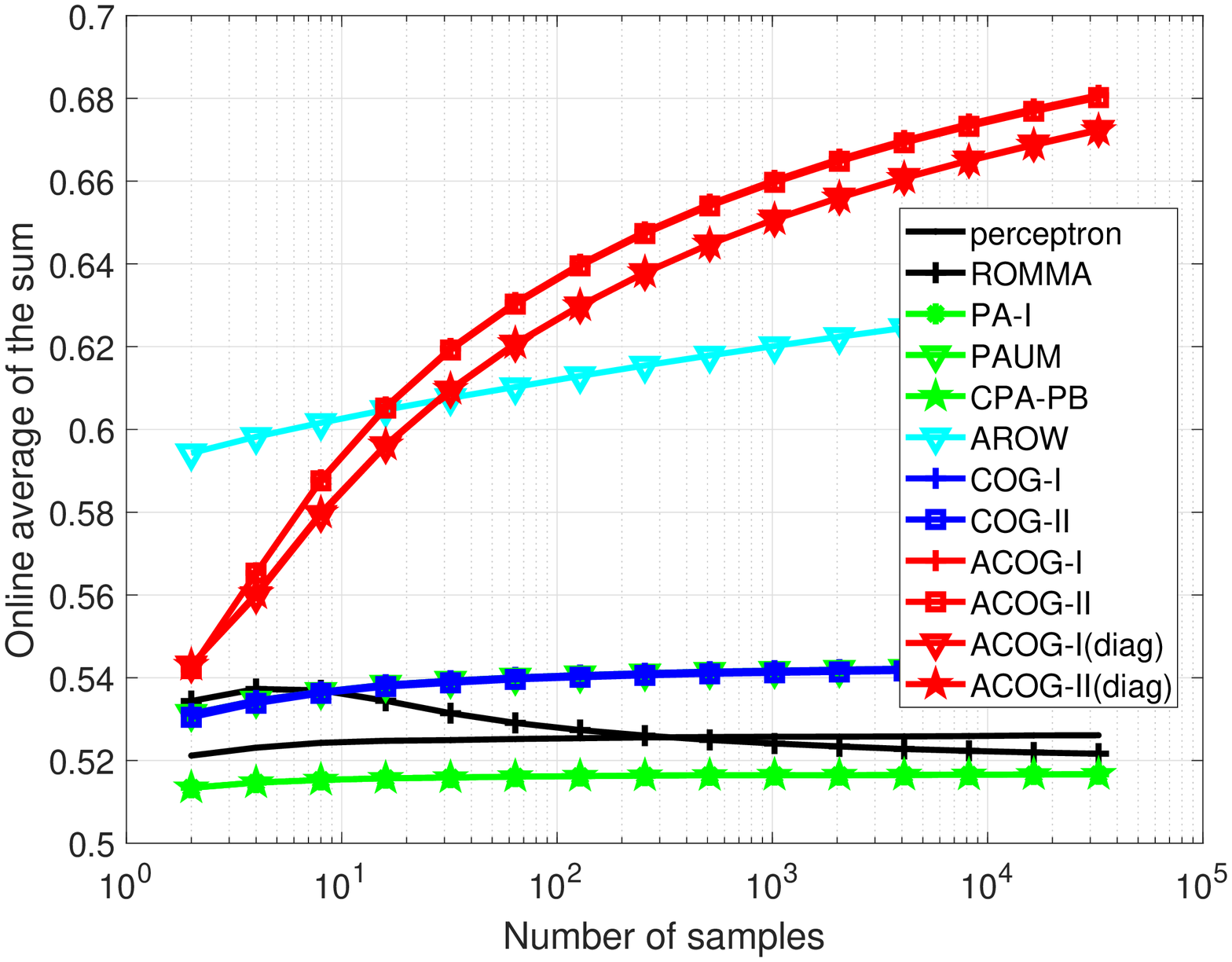}}
      \centerline{(b) covtype}
    \end{minipage}
    \vfill
    \begin{minipage}{0.5\linewidth}
      \centerline{\includegraphics[width=4.5cm]{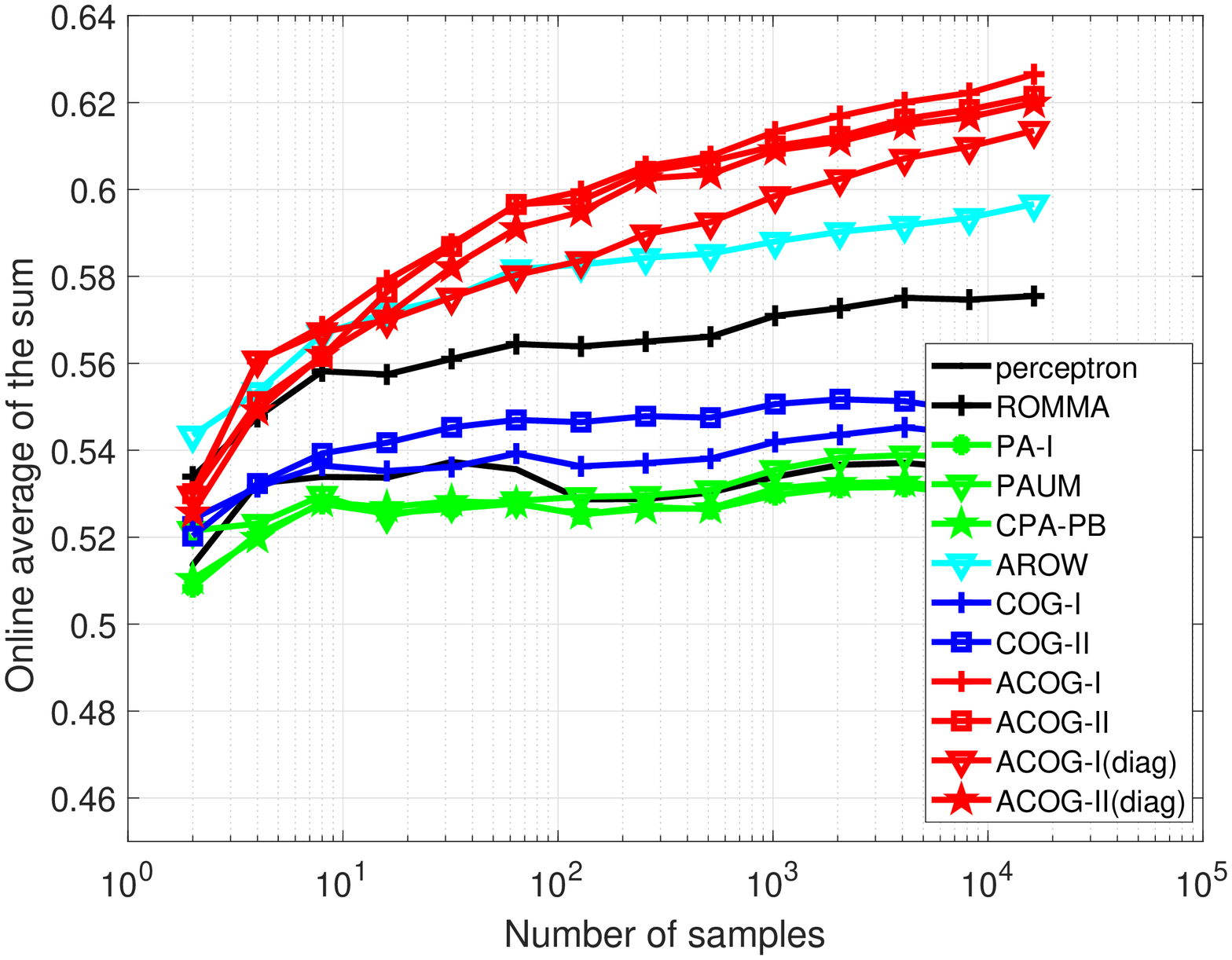}}
      \centerline{(c) german}
    \end{minipage}
    \hfill
    \begin{minipage}{0.5\linewidth}
      \centerline{\includegraphics[width=4.5cm]{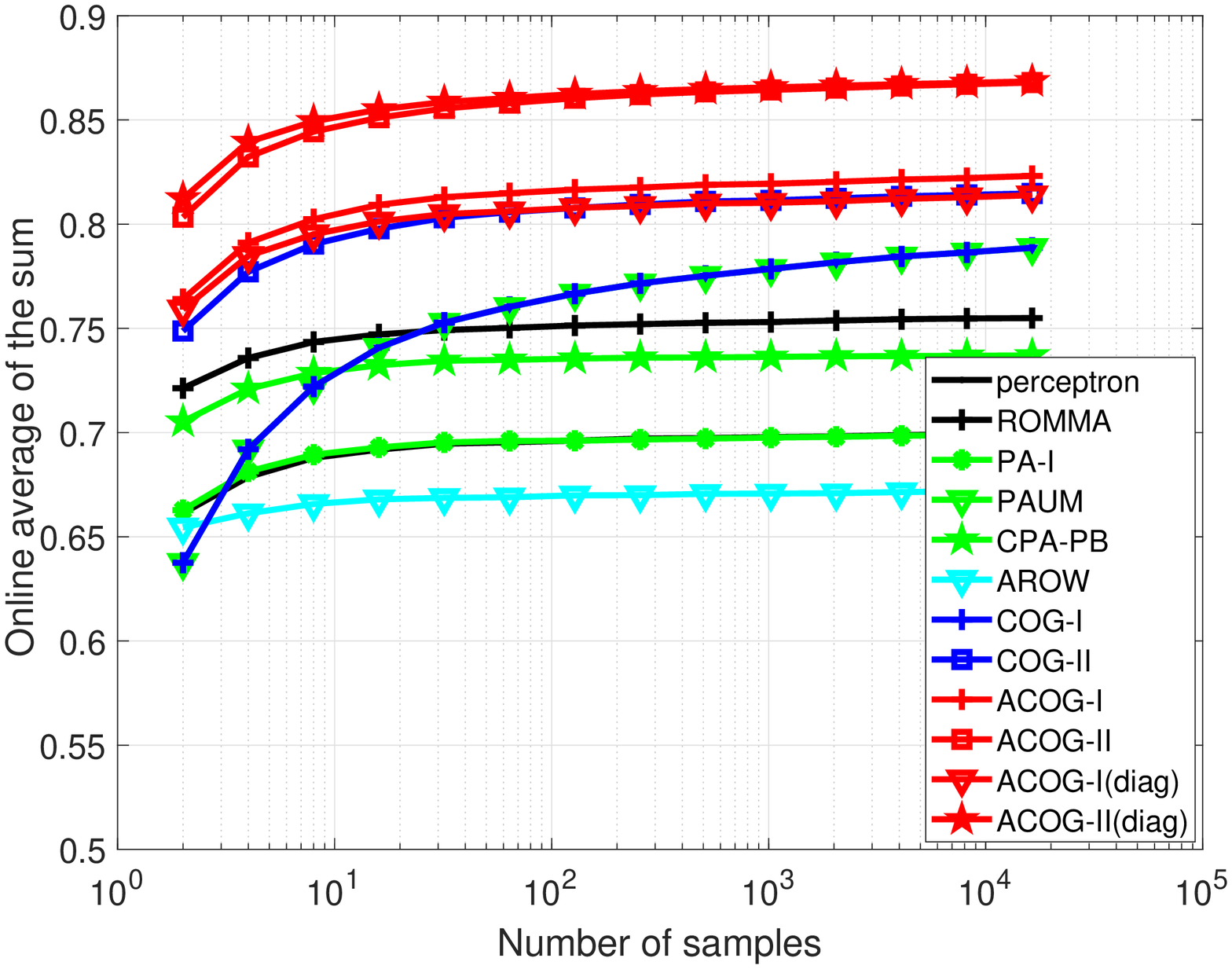}}
      \centerline{(d) ijcnn1}
    \end{minipage}
    \caption{Evaluation of online ``sum`` performance of the proposed algorithms on public datasets.}
    \label{sum}
\end{figure}

\begin{table*}
	\caption{Evaluation of the Cost-Sensitive Classification Performance of ACOG and Other Algorithms}
    \vspace{-0.1in}
    \begin{center}
    \scalebox{0.86}{
	\begin{tabular}{|l|c|c|c|c|c|c|c|c|}\hline
	
        \multirow{2}{*}{Algorithm}&\multicolumn{4}{c|}{$``sum``$ on a9a} &  \multicolumn{4}{c|}{$``cost``$ on a9a}\cr\cline{2-9}
        &Sum(\%)&Sensitivity(\%)&Specificity  (\%)&Time(s)   & Cost($10^2$)&Sensitivity(\%)&Specificity  (\%)&Time(s)\cr
        \hline \hline

        Perceptron   &71.343 	$\pm$ 0.215 	& 56.406 	$\pm$ 0.327 	& 86.280 	$\pm$ 0.102 	& 0.196 	 &50.951 	$\pm$ 0.382 	& 56.406 	$\pm$ 0.327 	& 86.280 	$\pm$ 0.102 	& 0.191 	 \\
        ROMMA        &70.904 	$\pm$ 0.239 	& 57.918 	$\pm$ 0.493 	& 83.889 	$\pm$ 0.262 	& 0.225 	  &50.184 	$\pm$ 0.361 	& 57.989 	$\pm$ 0.346 	& 83.863 	$\pm$ 0.227 	& 0.224 	 \\
        PA-I         &71.274 	$\pm$ 0.169 	& 56.310 	$\pm$ 0.277 	& 86.237 	$\pm$ 0.113 	& 0.212 	    &51.068 	$\pm$ 0.311 	& 56.310 	$\pm$ 0.277 	& 86.237 	$\pm$ 0.113 	& 0.212 	 \\
        PAUM         &78.255 	$\pm$ 0.155 	& 70.868 	$\pm$ 0.345 	& 85.643 	$\pm$ 0.116 	& 0.192 	   &35.976 	$\pm$ 0.346 	& 70.868 	$\pm$ 0.345 	& 85.643 	$\pm$ 0.116 	& 0.197 	 \\
        CPA$_{PB}$  &72.678 	$\pm$ 0.209 	& 62.818 	$\pm$ 0.345 	& 82.537 	$\pm$ 0.145 	& 0.254 	   &42.517 	$\pm$ 0.326 	& 66.818 	$\pm$ 0.285 	& 79.503 	$\pm$ 0.132 	& 0.246 	 \\
        AROW  &75.854 	$\pm$ 0.188 	& 58.858 	$\pm$ 0.510 	& \textbf{92.849 	$\pm$ 0.153} 	& 5.591 	 &45.931 	$\pm$ 0.486 	& 58.858 	$\pm$ 0.510 	& \textbf{92.849 	$\pm$ 0.153} 	& 5.104 	 \\
        COG-I  &78.978 	$\pm$ 0.128 	& 71.967 	$\pm$ 0.264 	& 85.990 	$\pm$ 0.137 	& 0.192 	 &28.632 	$\pm$ 0.263 	& 79.390 	$\pm$ 0.241 	& 81.284 	$\pm$ 0.107 	& 0.190 	 \\
        COG-II  &79.126 	$\pm$ 0.103 	& 81.038 	$\pm$ 0.243 	& 77.213 	$\pm$ 0.168 	& 0.201 	  &25.527 	$\pm$ 0.182 	& 89.013 	$\pm$ 0.171 	& 62.398 	$\pm$ 0.243 	& 0.193 	 \\
        ACOG-I  &79.903 	$\pm$ 0.109 	& 73.561 	$\pm$ 0.347 	& 86.244 	$\pm$ 0.162 	& 3.080 	  &26.760 	$\pm$ 0.291 	& 81.129 	$\pm$ 0.340 	& 81.398 	$\pm$ 0.219 	& 2.837 	 \\
        ACOG-II  &\textbf{81.220 	$\pm$ 0.108} 	& \textbf{85.269 	$\pm$ 0.219} 	& 77.171 	$\pm$ 0.134 	& 3.344 	&\textbf{20.307 	$\pm$ 0.169} 	& \textbf{94.079 	$\pm$ 0.136} 	& 62.107 	$\pm$ 0.185 	& 3.612 	 \\
        ACOG-I$_{diag}$  &79.827 	$\pm$ 0.094 	& 73.361 	$\pm$ 0.245 	& 86.293 	$\pm$ 0.103 	& 0.202 	  &26.917 	$\pm$ 0.253 	& 80.990 	$\pm$ 0.282 	& 81.369 	$\pm$ 0.147 	& 0.205 	 \\
        ACOG-II$_{diag}$  &\textbf{81.098 	$\pm$ 0.083} 	& \textbf{84.705 	$\pm$ 0.227} 	& 77.491 	$\pm$ 0.152 	& 0.216 	  &\textbf{20.661 	$\pm$ 0.110} 	& \textbf{93.352 	$\pm$ 0.126} 	& 63.212 	$\pm$ 0.238 	& 0.213 	 \\

        \hline
        \hline

        \multirow{2}{*}{Algorithm}&\multicolumn{4}{c|}{$``sum``$ on covtype} &  \multicolumn{4}{c|}{$``cost``$ on covtype}\cr\cline{2-9}
        &Sum(\%)&Sensitivity(\%)&Specificity  (\%)&Time(s)   & Cost($10^2$)&Sensitivity(\%)&Specificity  (\%)&Time(s)\cr
        \hline \hline

        Perceptron   &52.609 	$\pm$ 0.057 	& 51.364 	$\pm$ 0.058 	& 53.854 	$\pm$ 0.057 	& 1.649 	&1377.464 	$\pm$ 1.638 	& 51.364 	$\pm$ 0.058 	& 53.854 	$\pm$ 0.057 	& 1.662 	 \\
        ROMMA        &52.164 	$\pm$ 0.674 	& 50.819 	$\pm$ 0.731 	& 53.509 	$\pm$ 0.647 	& 2.233 	  &1391.250 	$\pm$ 19.560 	& 50.860 	$\pm$ 0.702 	& 53.541 	$\pm$ 0.614 	& 2.295 	 \\
        PA-I         &51.666 	$\pm$ 0.056 	& 50.324 	$\pm$ 0.061 	& 53.008 	$\pm$ 0.063 	& 1.869 	   &1406.500 	$\pm$ 1.675 	& 50.324 	$\pm$ 0.061 	& 53.008 	$\pm$ 0.063 	& 1.913 	 \\
        PAUM         &54.268 	$\pm$ 0.059 	& 52.588 	$\pm$ 0.089 	& 55.949 	$\pm$ 0.066 	& 1.693 	   &1340.022 	$\pm$ 2.311 	& 52.588 	$\pm$ 0.089 	& 55.949 	$\pm$ 0.066 	& 1.709 	 \\
        CPA$_{PB}$  &51.667 	$\pm$ 0.057 	& 50.552 	$\pm$ 0.063 	& 52.781 	$\pm$ 0.065 	& 2.135 	  &1194.433 	$\pm$ 1.911 	& 59.661 	$\pm$ 0.070 	& 44.275 	$\pm$ 0.072 	& 2.199 	 \\
        AROW  &63.036 	$\pm$ 0.033 	& 60.158 	$\pm$ 0.244 	& \textbf{65.914 	$\pm$ 0.213} 	& 22.640 	  &687.696 	$\pm$ 3.148 	& 76.580 	$\pm$ 0.137 	& \textbf{69.579 	$\pm$ 0.134} 	& 22.556 	 \\
        COG-I  &54.268 	$\pm$ 0.059 	& 52.588 	$\pm$ 0.089 	& 55.949 	$\pm$ 0.066 	& 1.637 	 &631.834 	$\pm$ 1.721 	& 84.036 	$\pm$ 0.070 	& 24.494 	$\pm$ 0.062 	& 1.710 	 \\
        COG-II  &54.208 	$\pm$ 0.051 	& 54.038 	$\pm$ 0.096 	& 54.377 	$\pm$ 0.055 	& 1.643 	 &426.122 	$\pm$ 0.834 	& 94.088 	$\pm$ 0.031 	& 7.501 	$\pm$ 0.107 	& 1.657 	 \\
        ACOG-I  &\textbf{68.077 	$\pm$ 0.038} 	& \textbf{70.565 	$\pm$ 0.073} 	& \textbf{65.588 	$\pm$ 0.082} 	& 13.782 	  &466.376 	$\pm$ 1.190 	& 90.693 	$\pm$ 0.049 	& 23.054 	$\pm$ 0.038 	& 18.988 	 \\
        ACOG-II  &\textbf{68.020 	$\pm$ 0.030} 	& \textbf{71.265 	$\pm$ 0.070} 	& 64.774 	$\pm$ 0.068 	& 13.528 	 &\textbf{305.056 	$\pm$ 0.355} 	& \textbf{98.969 	$\pm$ 0.021} 	& 6.365 	$\pm$ 0.163 	& 13.232 	 \\
        ACOG-I$_{diag}$  &67.247 	$\pm$ 0.060 	& 69.183 	$\pm$ 0.076 	& \textbf{65.311 	$\pm$ 0.082} 	& 1.824 	 &469.701 	$\pm$ 1.377 	& 90.594 	$\pm$ 0.090 	& 22.782 	$\pm$ 0.370 	& 1.971 	 \\
        ACOG-II$_{diag}$  &67.225 	$\pm$ 0.062 	& 69.913 	$\pm$ 0.096 	& 64.537 	$\pm$ 0.086 	& 1.805 	 &\textbf{308.987 	$\pm$ 7.944} 	& \textbf{98.739 	$\pm$ 0.367} 	& 7.015 	$\pm$ 0.507 	& 1.828 	 \\

        \hline
        \hline
        \multirow{2}{*}{Algorithm}&\multicolumn{4}{c|}{$``sum``$ on german} &  \multicolumn{4}{c|}{$``cost``$ on german}\cr\cline{2-9}
        &Sum(\%)&Sensitivity(\%)&Specificity  (\%)&Time(s)   & Cost($10^2$)&Sensitivity(\%)&Specificity  (\%)&Time(s)\cr
        \hline \hline

        Perceptron   &53.760 	$\pm$ 1.655 	& 35.133 	$\pm$ 2.343 	& 72.386 	$\pm$ 0.977 	& 0.003 	&1.945 	$\pm$ 0.070 	& 35.133 	$\pm$ 2.343 	& 72.386 	$\pm$ 0.977 	& 0.003 	 \\
        ROMMA        &57.625 	$\pm$ 2.943 	& 43.550 	$\pm$ 4.496 	& 71.700 	$\pm$ 1.710 	& 0.004 	 &1.721 	$\pm$ 0.128 	& 43.650 	$\pm$ 4.372 	& 71.536 	$\pm$ 1.932 	& 0.004 	 \\
        PA-I         &53.043 	$\pm$ 1.902 	& 34.000 	$\pm$ 2.818 	& 72.086 	$\pm$ 1.128 	& 0.003 	    &1.977 	$\pm$ 0.083 	& 34.000 	$\pm$ 2.818 	& 72.086 	$\pm$ 1.128 	& 0.003 	 \\
        PAUM         &54.145 	$\pm$ 1.335 	& 26.483 	$\pm$ 3.633 	& 81.807 	$\pm$ 1.341 	& 0.003 	    &2.112 	$\pm$ 0.091 	& 26.483 	$\pm$ 3.633 	& 81.807 	$\pm$ 1.341 	& 0.003 	 \\
        CPA$_{PB}$  &53.185 	$\pm$ 1.948 	& 37.883 	$\pm$ 2.925 	& 68.486 	$\pm$ 1.144 	& 0.004 	  &1.759 	$\pm$ 0.082 	& 44.317 	$\pm$ 2.883 	& 63.464 	$\pm$ 1.287 	& 0.004 	 \\
        AROW  &59.948 	$\pm$ 1.295 	& 26.367 	$\pm$ 3.893 	& \textbf{93.529 	$\pm$ 1.630} 	& 0.014 	  &1.610 	$\pm$ 0.082 	& 43.867 	$\pm$ 3.364 	& \textbf{86.571 	$\pm$ 1.543} 	& 0.016 	 \\
        COG-I  &54.424 	$\pm$ 1.474 	& 36.083 	$\pm$ 2.203 	& 72.764 	$\pm$ 0.807 	& 0.003 	 &1.770 	$\pm$ 0.081 	& 42.933 	$\pm$ 3.010 	& 67.200 	$\pm$ 0.990 	& 0.003 	 \\
        COG-II  &54.952 	$\pm$ 1.359 	& 54.833 	$\pm$ 1.318 	& 55.071 	$\pm$ 1.442 	& 0.003 	   &1.035 	$\pm$ 0.033 	& 81.067 	$\pm$ 0.799 	& 25.200 	$\pm$ 1.983 	& 0.003 	 \\
        ACOG-I  &\textbf{63.150 	$\pm$ 1.02}5 	& 49.050 	$\pm$ 1.932 	& 77.250 	$\pm$ 1.489 	& 0.008 	   &1.232 	$\pm$ 0.049 	& 62.750 	$\pm$ 2.017 	& 67.671 	$\pm$ 1.394 	& 0.010 	 \\
        ACOG-II  &62.511 	$\pm$ 1.190 	& \textbf{63.000 	$\pm$ 2.052} 	& 62.021 	$\pm$ 1.408 	& 0.008 	 &\textbf{0.875 	$\pm$ 0.044} 	& \textbf{86.883 	$\pm$ 2.264} 	& 25.564 	$\pm$ 4.099 	& 0.011 	 \\
        ACOG-I$_{diag}$  &61.765 	$\pm$ 1.195 	& 47.517 	$\pm$ 2.610 	& 76.014 	$\pm$ 1.022 	& 0.003 	 &1.330 	$\pm$ 0.064 	& 58.967 	$\pm$ 2.362 	& 68.300 	$\pm$ 0.901 	& 0.003 	 \\
        ACOG-II$_{diag}$  &62.281 	$\pm$ 1.428 	& 62.883 	$\pm$ 1.852 	& 61.679 	$\pm$ 1.576 	& 0.003 	  &\textbf{0.912 	$\pm$ 0.045} 	& \textbf{84.733 	$\pm$ 0.876} 	& 28.629 	$\pm$ 4.046 	& 0.003 	 \\

        \hline
        \hline
        \multirow{2}{*}{Algorithm}&\multicolumn{4}{c|}{$``sum``$ on ijcnn1} &  \multicolumn{4}{c|}{$``cost``$ on ijcnn1}\cr\cline{2-9}
        &Sum(\%)&Sensitivity(\%)&Specificity  (\%)&Time(s)   & Cost($10^2$)&Sensitivity(\%)&Specificity  (\%)&Time(s)\cr
        \hline \hline

       Perceptron   &69.988 	$\pm$ 0.252 	& 45.926 	$\pm$ 0.455 	& 94.051 	$\pm$ 0.050 	& 0.112 	   &26.303 	$\pm$ 0.221 	& 45.926 	$\pm$ 0.455 	& 94.051 	$\pm$ 0.050 	& 0.114 	 \\
        ROMMA        &75.547 	$\pm$ 0.229 	& 57.689 	$\pm$ 0.439 	& 93.405 	$\pm$ 0.111 	& 0.124 	   &21.467 	$\pm$ 0.207 	& 57.666 	$\pm$ 0.459 	& 93.404 	$\pm$ 0.108 	& 0.128 	 \\
        PA-I         &69.980 	$\pm$ 0.312 	& 45.542 	$\pm$ 0.579 	& 94.418 	$\pm$ 0.083 	& 0.119 	     &26.305 	$\pm$ 0.274 	& 45.542 	$\pm$ 0.579 	& 94.418 	$\pm$ 0.083 	& 0.124 	 \\
        PAUM         &79.066 	$\pm$ 0.275 	& 64.377 	$\pm$ 0.590 	& 93.755 	$\pm$ 0.092 	& 0.112 	    &18.378 	$\pm$ 0.239 	& 64.377 	$\pm$ 0.590 	& 93.755 	$\pm$ 0.092 	& 0.118 	 \\
        CPA$_{PB}$  &73.745 	$\pm$ 0.209 	& 57.328 	$\pm$ 0.371 	& 90.161 	$\pm$ 0.091 	& 0.155 	   &23.096 	$\pm$ 0.200 	& 57.215 	$\pm$ 0.407 	& 90.233 	$\pm$ 0.094 	& 0.160 	 \\
        AROW  &67.258 	$\pm$ 0.460 	& 36.208 	$\pm$ 0.980 	& \textbf{98.308 	$\pm$ 0.074} 	& 0.450 	  &28.626 	$\pm$ 0.401 	& 36.208 	$\pm$ 0.980 	& \textbf{98.308 	$\pm$ 0.074} 	& 0.465 	 \\
        COG-I  &79.066 	$\pm$ 0.275 	& 64.377 	$\pm$ 0.590 	& 93.755 	$\pm$ 0.092 	& 0.109 	   &18.441 	$\pm$ 0.236 	& 64.171 	$\pm$ 0.590 	& 93.814 	$\pm$ 0.096 	& 0.116 	 \\
        COG-II  &81.520 	$\pm$ 0.232 	& 81.940 	$\pm$ 0.363 	& 81.100 	$\pm$ 0.182 	& 0.112 	   &16.398 	$\pm$ 0.197 	& 81.683 	$\pm$ 0.311 	& 81.394 	$\pm$ 0.205 	& 0.116 	 \\
        ACOG-I  &82.375 	$\pm$ 0.230 	& 71.010 	$\pm$ 0.607 	& 93.740 	$\pm$ 0.178 	& 0.212 	  &15.197 	$\pm$ 0.123 	& 71.996 	$\pm$ 0.352 	& 93.429 	$\pm$ 0.102 	& 0.218 	 \\
        ACOG-II  &\textbf{86.872 	$\pm$ 0.174} 	& \textbf{88.924 	$\pm$ 0.323} 	& 84.820 	$\pm$ 0.218 	& 0.288 	  &\textbf{12.279 	$\pm$ 0.149} 	& \textbf{87.626 	$\pm$ 0.293} 	& 84.770 	$\pm$ 0.165 	& 0.298 	 \\
        ACOG-I$_{diag}$  &81.468 	$\pm$ 0.225 	& 69.007 	$\pm$ 0.502 	& 93.929 	$\pm$ 0.092 	& 0.114 	  &15.681 	$\pm$ 0.227 	& 70.680 	$\pm$ 0.624 	& 93.631 	$\pm$ 0.127 	& 0.122 	 \\
        ACOG-II$_{diag}$  &\textbf{86.929 	$\pm$ 0.124} 	& \textbf{88.205 	$\pm$ 0.266} 	& 85.652 	$\pm$ 0.107 	& 0.120 	 &\textbf{12.016 	$\pm$ 0.111} 	& \textbf{87.164 	$\pm$ 0.300} 	& 85.801 	$\pm$ 0.138 	& 0.122 	 \\

        \hline

	\end{tabular}}
    \end{center}
    \vspace{-0.1in}
\end{table*}

\subsubsection{Evaluation of Sum under Varying Weights}
\vspace{-0.05in}
\begin{figure}
    \begin{minipage}{0.5\linewidth}
      \centerline{\includegraphics[width=4.5cm]{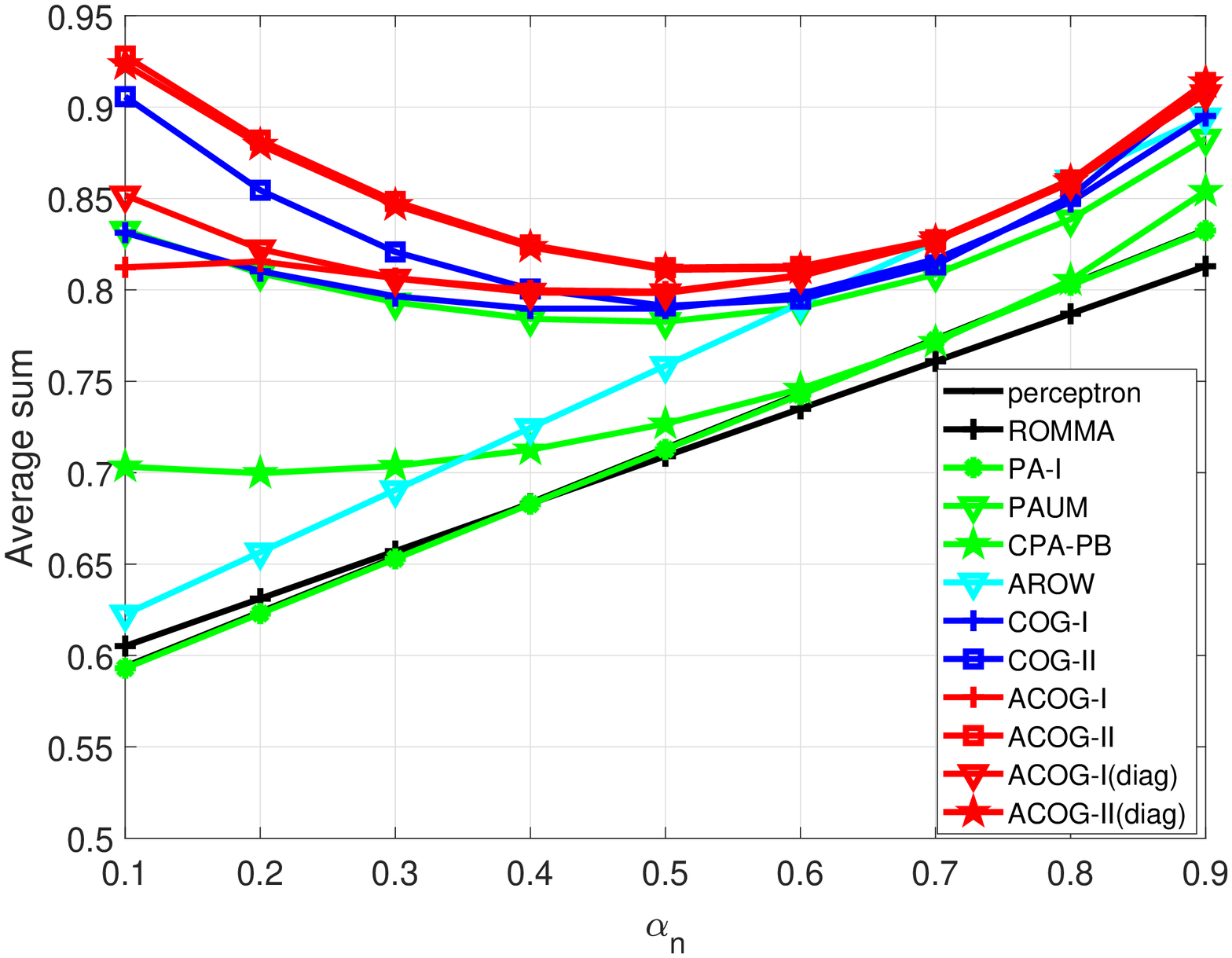}}
      \centerline{(a) a9a}
    \end{minipage}
    \hfill
    \begin{minipage}{0.5\linewidth}
      \centerline{\includegraphics[width=4.5cm]{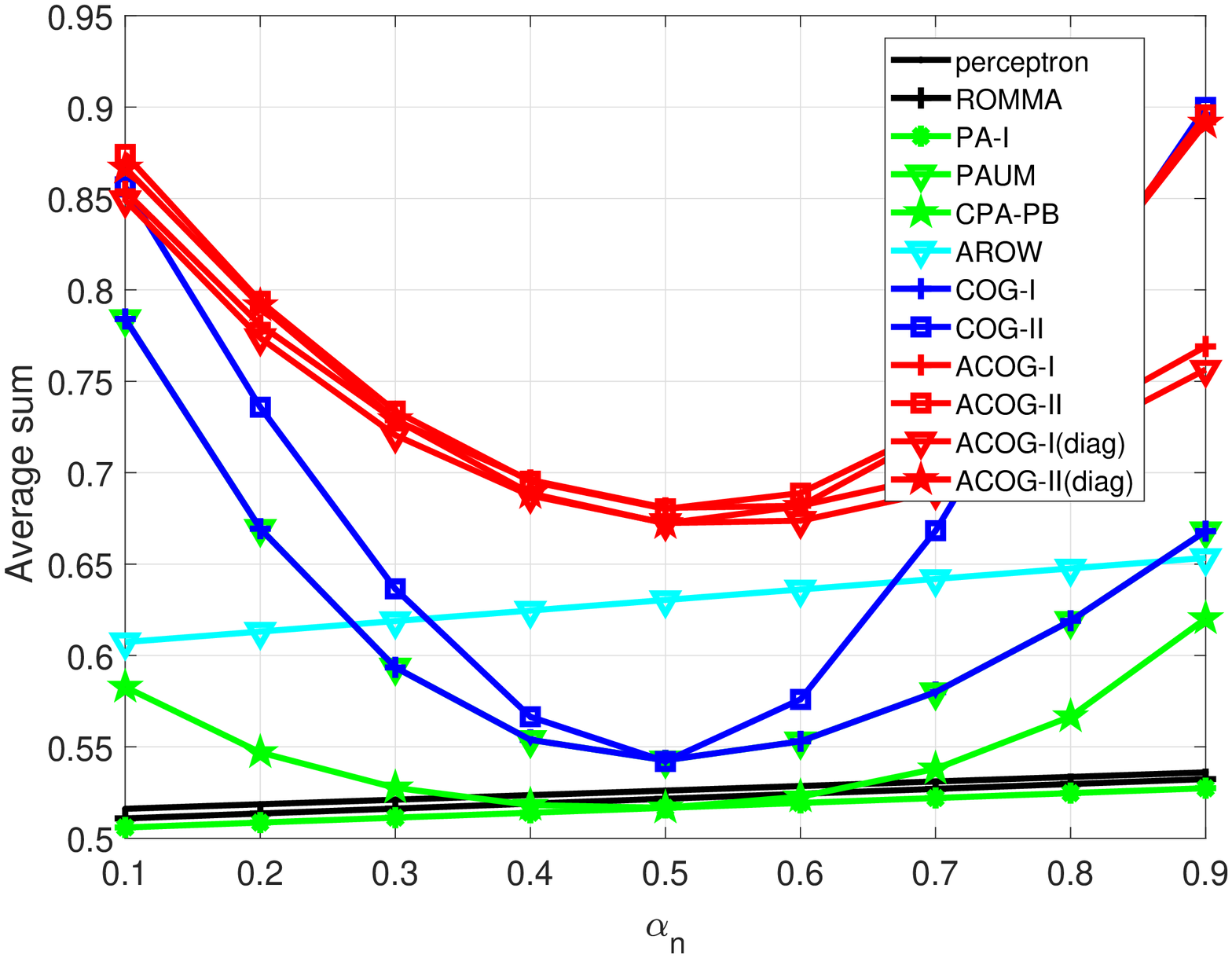}}
      \centerline{(b) covtype}
    \end{minipage}
    \vfill
    \begin{minipage}{0.5\linewidth}
      \centerline{\includegraphics[width=4.5cm]{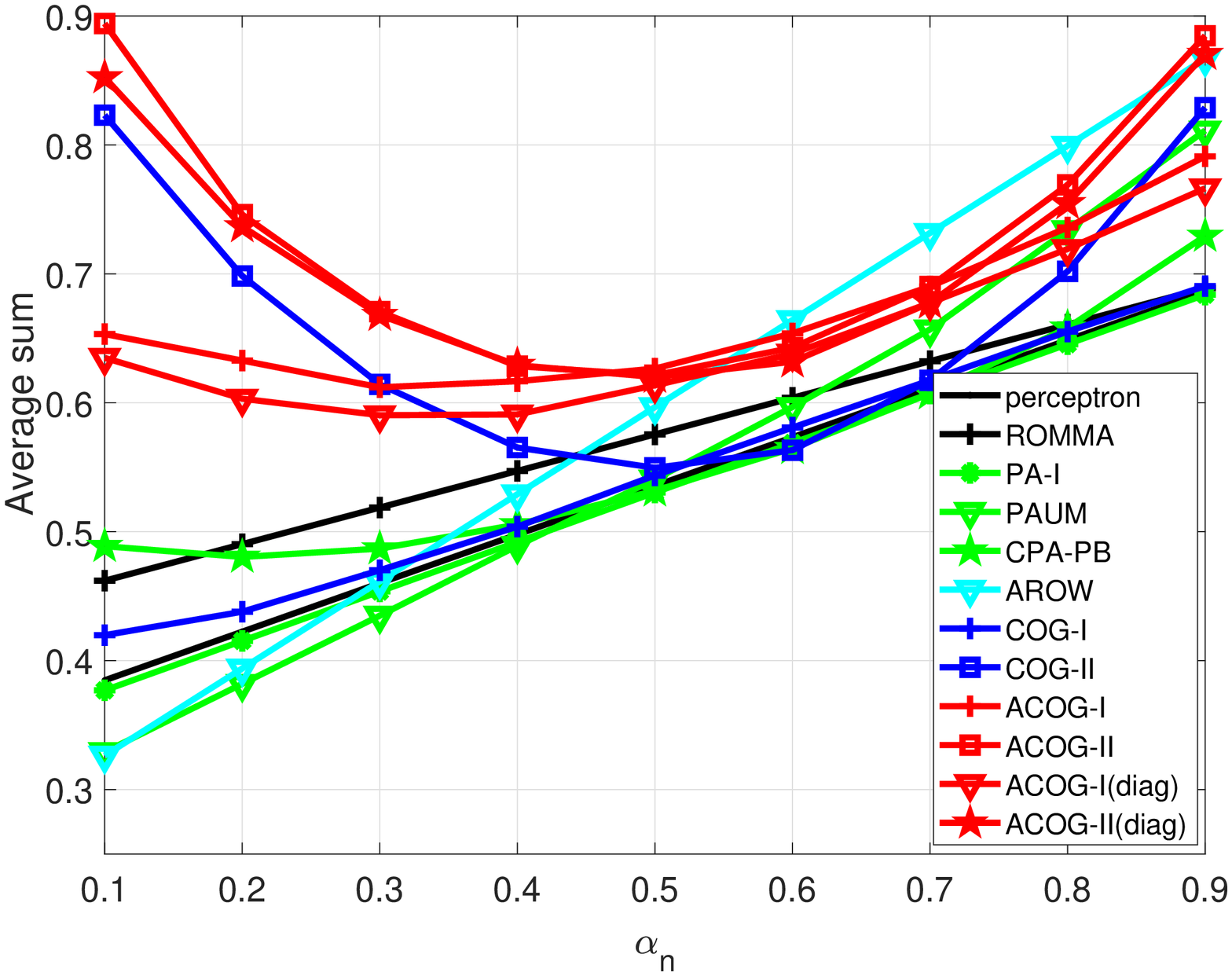}}
      \centerline{(c) german}
    \end{minipage}
    \hfill
    \begin{minipage}{0.5\linewidth}
      \centerline{\includegraphics[width=4.5cm]{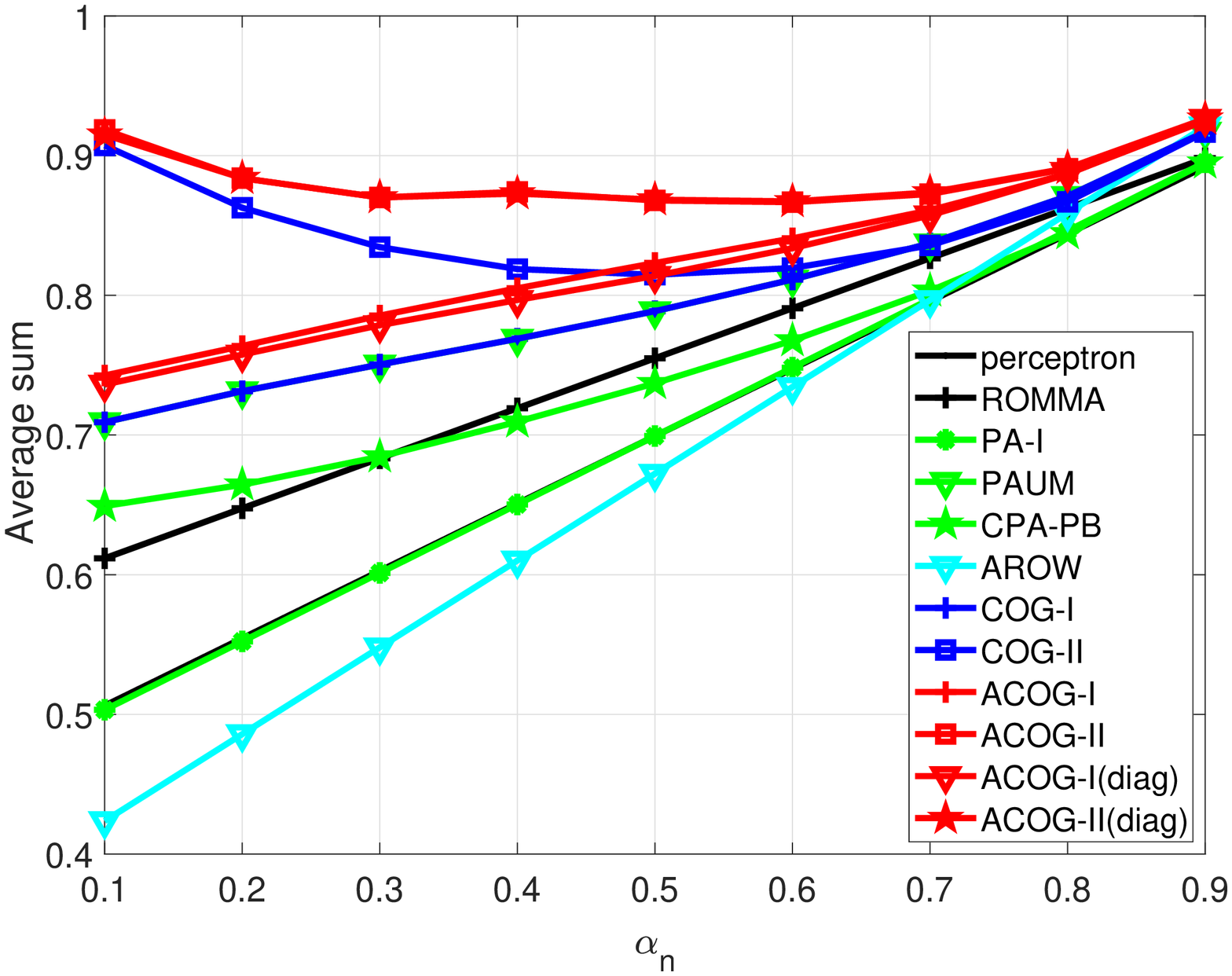}}
      \centerline{(d) ijcnn1}
    \end{minipage}

    \caption{Evaluation of weighted ``sum`` performance under varying weights of sensitivity and specificity.}
    \label{weight_sum}
\end{figure}

In this subsection, we would like to evaluate the $sum$ of proposed methods under different cost-sensitive weights. Fig. 2 shows the empirical results under different weights of $\alpha_n$ and $\alpha_p$. We find that our proposed algorithms consistently outperform all other algorithms under different values of weight on almost all datasets. This further validates the effectiveness of the proposed methods.

\vspace{-0.1in}
\subsection{Evaluation with Cost Metrics}
\subsubsection{Evaluation of Weighted Cost Performance}
\vspace{-0.05in}
Table 2 summaries the experimental performance of the ACOG$_{cost}$ on 4 datasets in terms of $cost$ metrics, and Fig. 3 illustrates the development of online $cost$ performance at each iteration.

By evaluating the $cost$ performance in Fig. 3 and Table 2, our proposed methods achieve much lower misclassification $cost$ than other methods among all cases. For example, the overall $cost$ of ACOG is about less than half of $cost$ made by all regular first-order algorithms (i.e., perceptron, ROMMA, PA-I, PAUM and CPA$_{PB}$). This implies that introducing the second order information is beneficial to the decrease of misclassification $cost$.

In addition, by examining both $sensitivety$ and $specificity$ metrics, we observe that our proposed methods often achieve the best $sensitivity$ result on all datasets, and attain a relatively good $specificity$ among all cases.

Moreover, the diagonal ACOG$_{diag}$ methods achieve higher $cost$ value than ACOG methods, but their running time are lower. This is similar with the situation based on $sum$ metric. Thus, the ACOG$_{diag}$ methods can be regarded as a choice to balance the performance and efficiency.

\begin{figure}
    \begin{minipage}{0.5\linewidth}
      \centerline{\includegraphics[width=4.5cm]{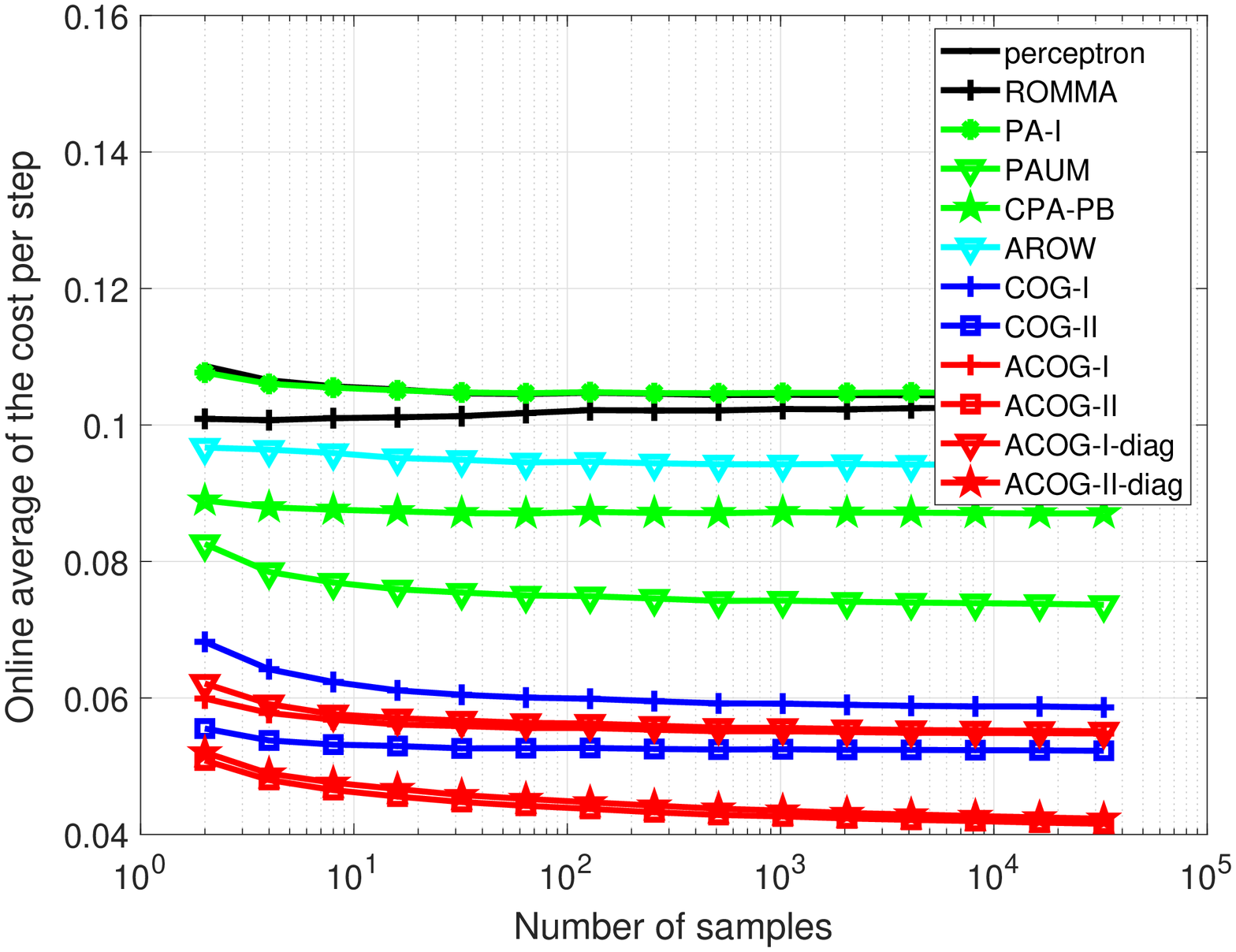}}
      \centerline{(a) a9a}
    \end{minipage}
    \hfill
    \begin{minipage}{0.5\linewidth}
      \centerline{\includegraphics[width=4.5cm]{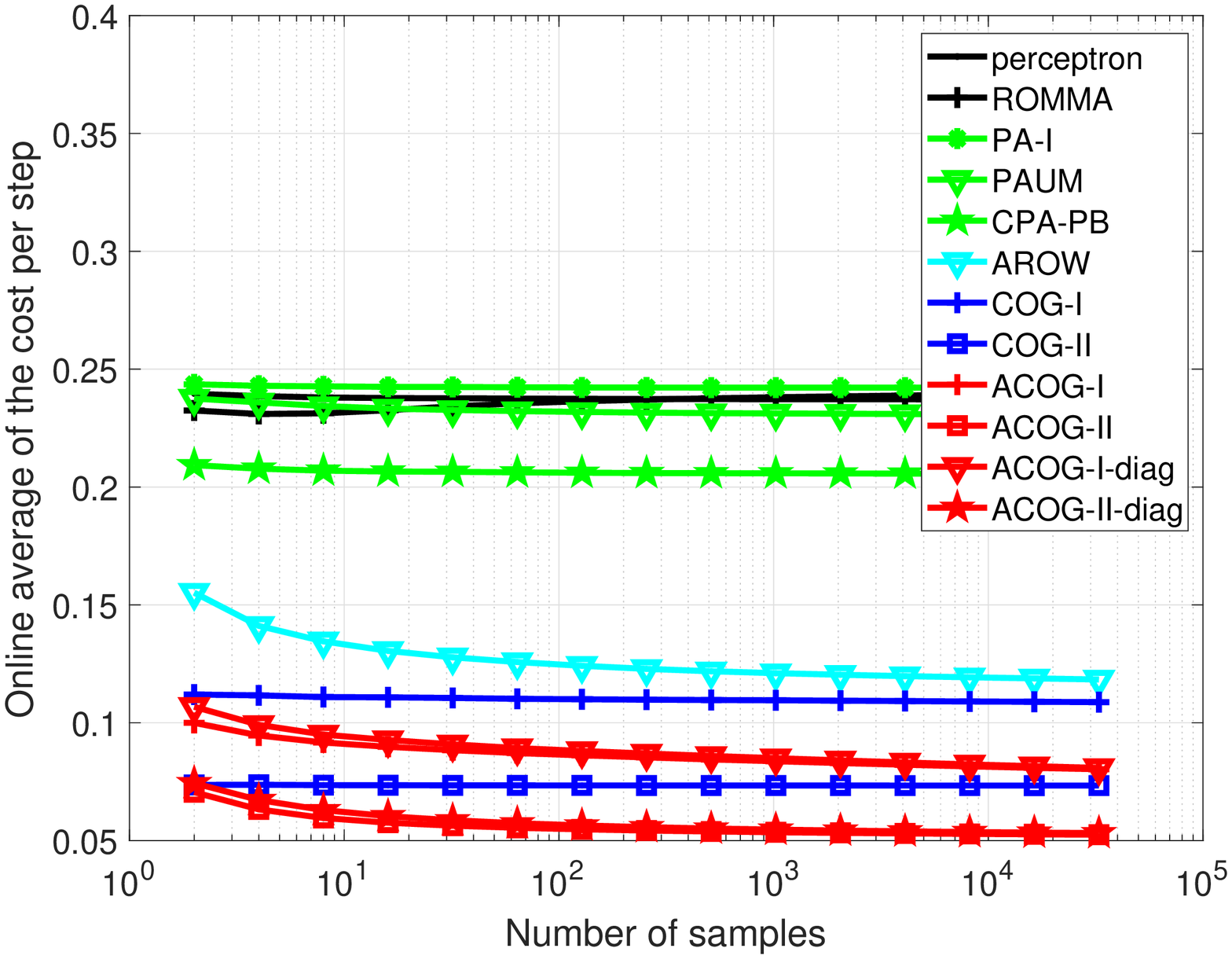}}
      \centerline{(b) covtype}
    \end{minipage}
    \vfill
    \begin{minipage}{0.5\linewidth}
      \centerline{\includegraphics[width=4.5cm]{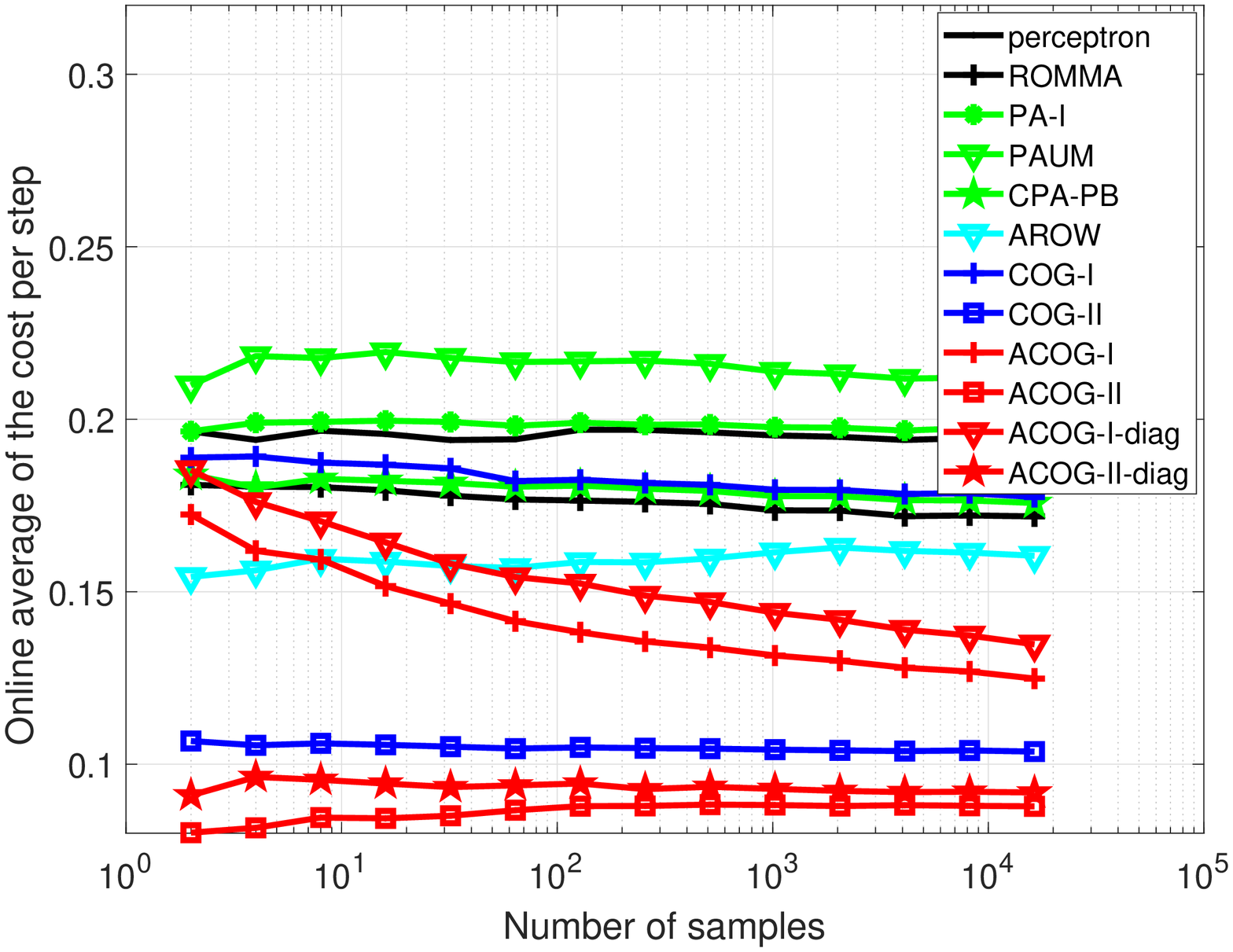}}
      \centerline{(c) german}
    \end{minipage}
    \hfill
    \begin{minipage}{0.5\linewidth}
      \centerline{\includegraphics[width=4.5cm]{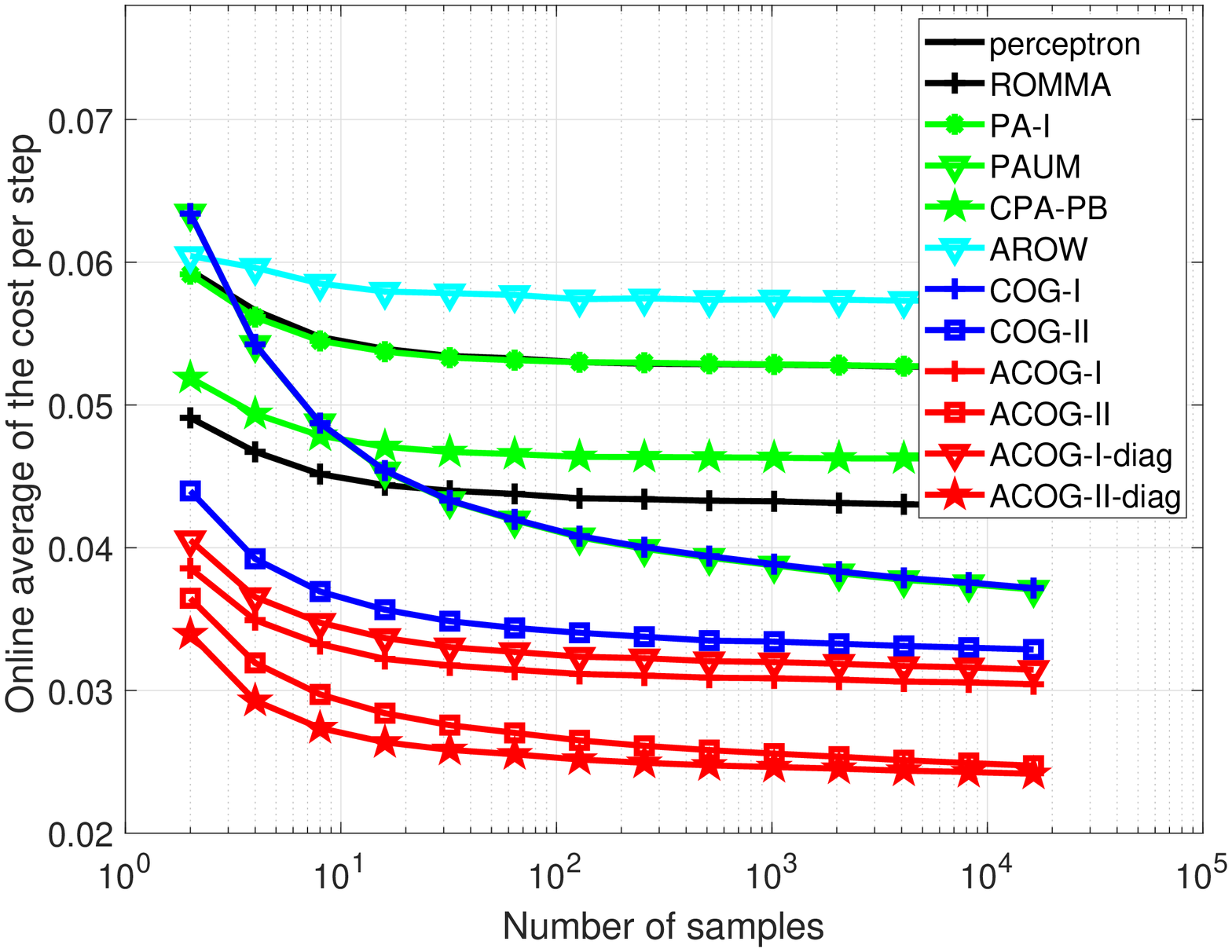}}
      \centerline{(d) ijcnn1}
    \end{minipage}
    \caption{Evaluation of online ``cost`` performance of the proposed algorithms on public datasets.}
    \label{cost}
\end{figure}

\subsubsection{Evaluation of Cost under Varying Weights}

In this subsection, we examine the $cost$ performance under different cost-sensitive weights $c_n$ and $c_p$ for our proposed algorithms. From the results in Fig. 4, we observe that the proposed algorithms outperform almost all other algorithms under different weights. And only on a few datasets, AROW can achieve similar performance with our proposed methods. These discoveries imply that our ACOG algorithms have a wide selection range of weight parameters for online classification tasks.

\begin{figure}
    \begin{minipage}{0.5\linewidth}
      \centerline{\includegraphics[width=4.5cm]{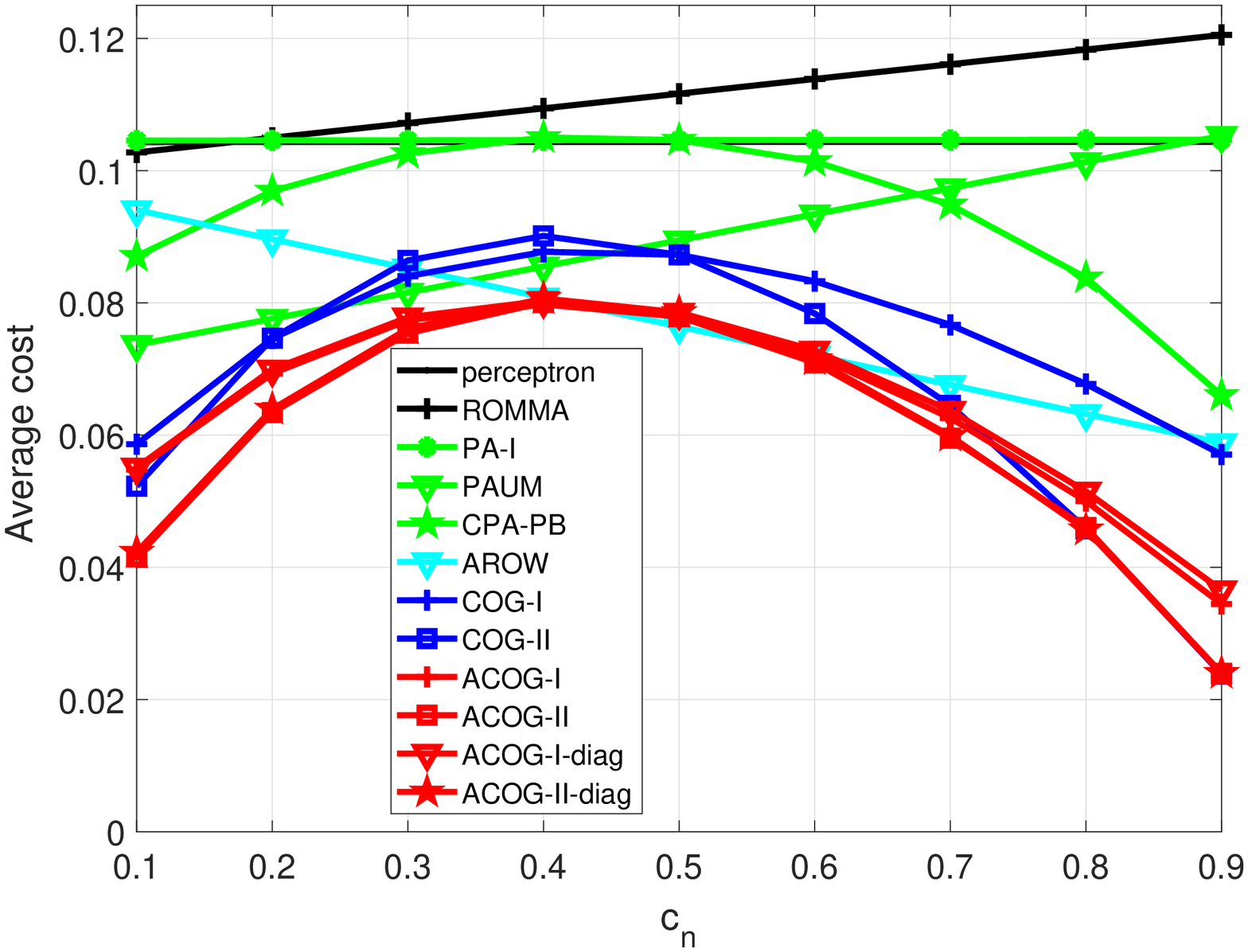}}
      \centerline{(a) a9a}
    \end{minipage}
    \hfill
    \begin{minipage}{0.5\linewidth}
      \centerline{\includegraphics[width=4.5cm]{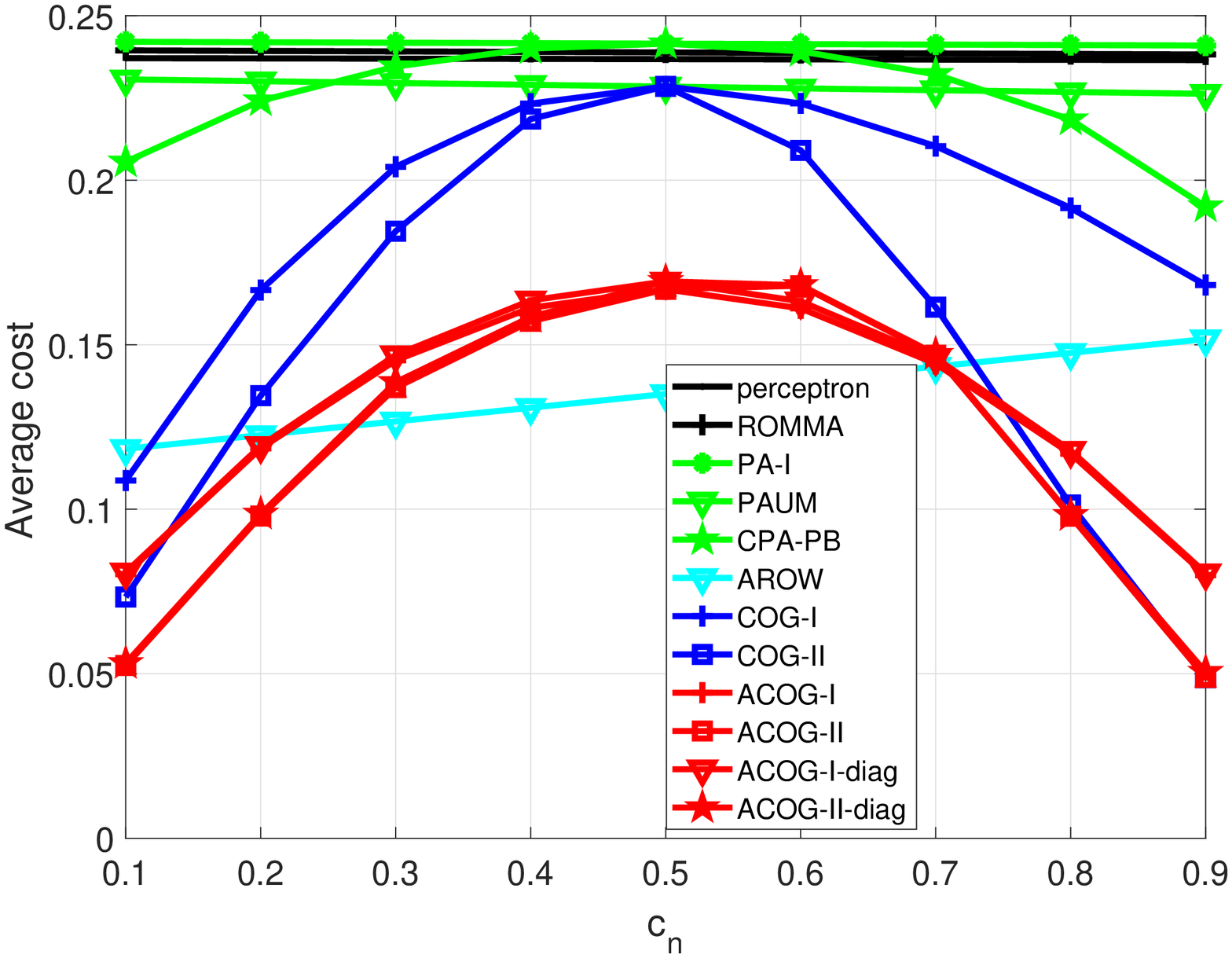}}
      \centerline{(b) covtype}
    \end{minipage}
    \vfill
    \begin{minipage}{0.5\linewidth}
      \centerline{\includegraphics[width=4.5cm]{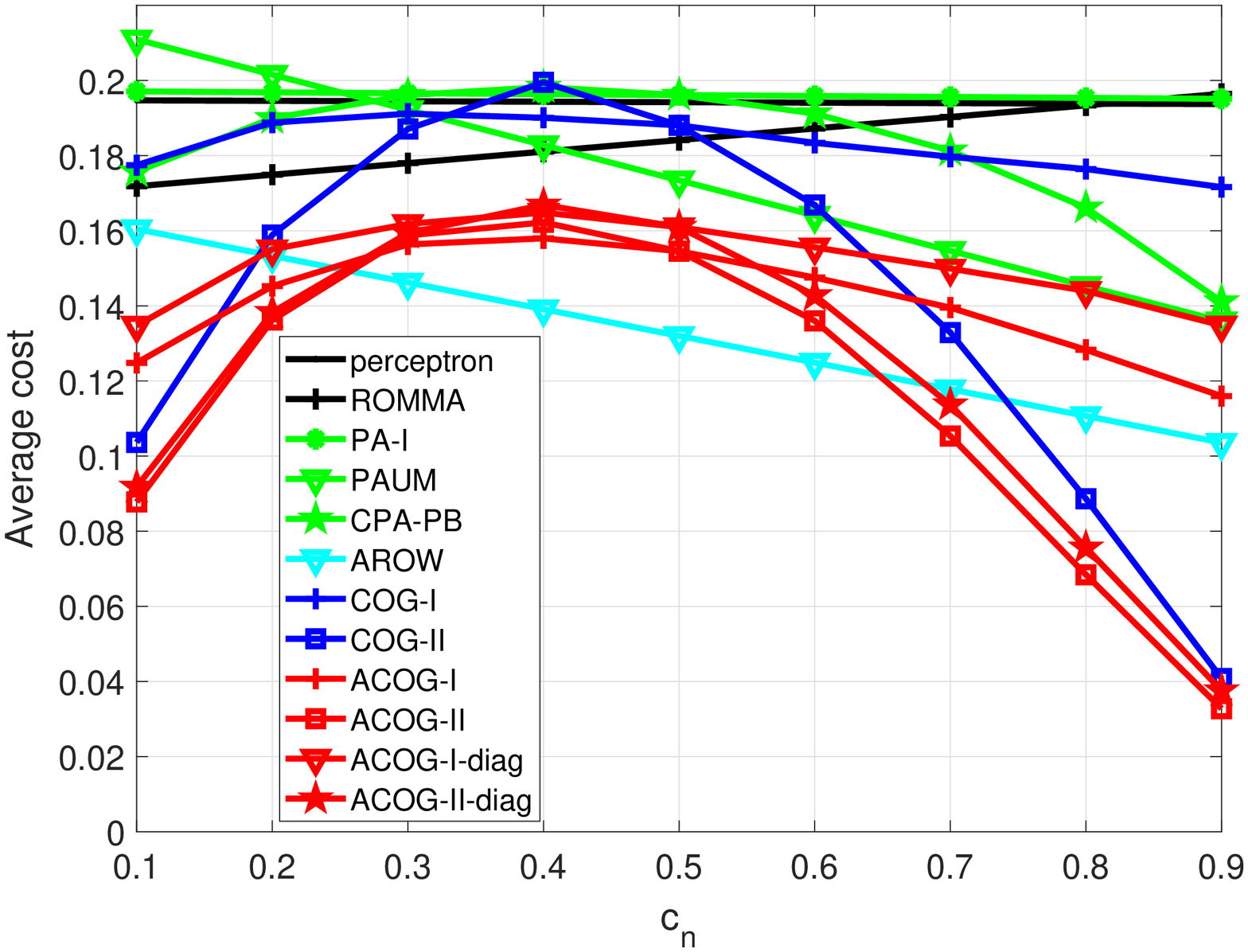}}
      \centerline{(c) german}
    \end{minipage}
    \hfill
    \begin{minipage}{0.5\linewidth}
      \centerline{\includegraphics[width=4.5cm]{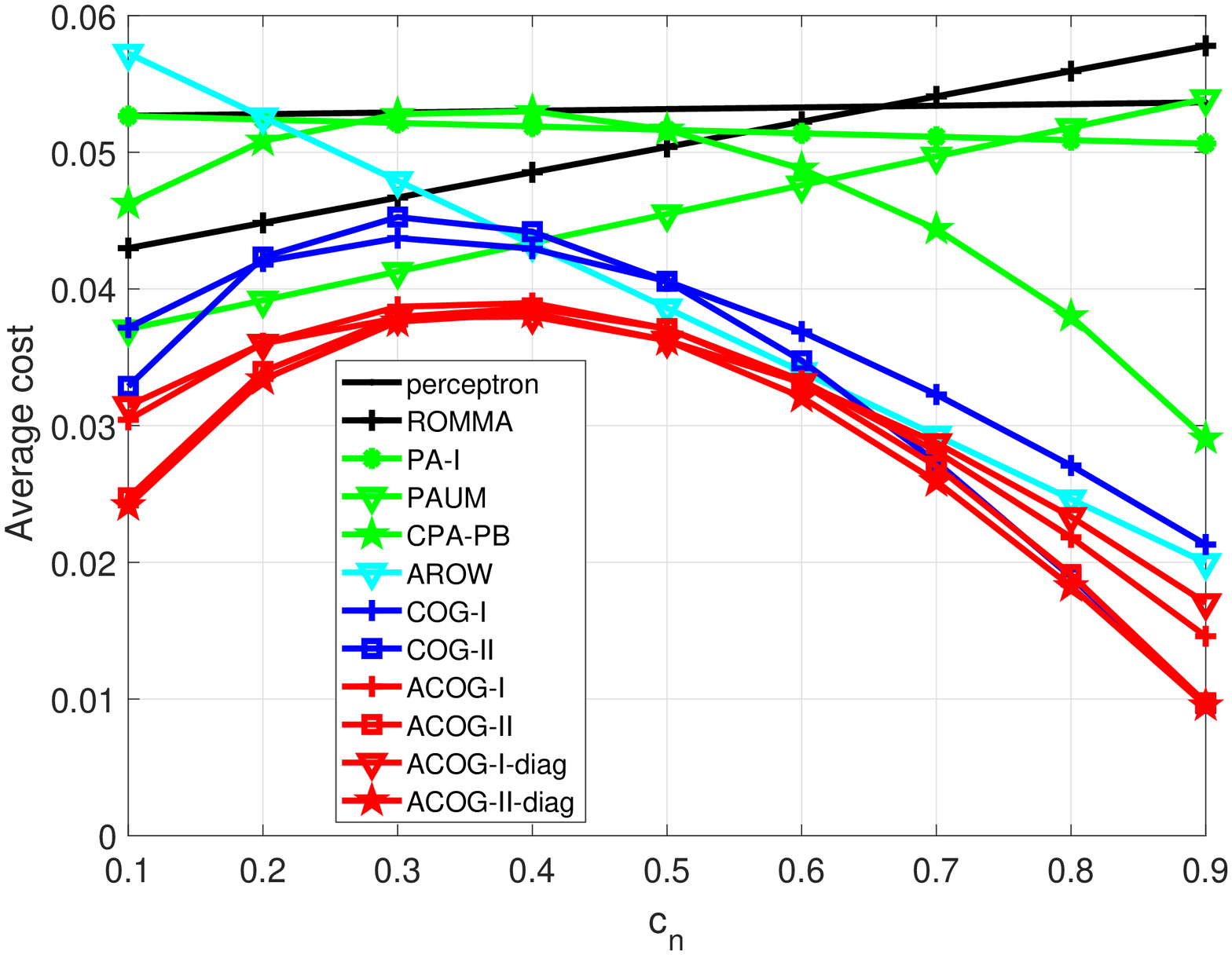}}
      \centerline{(d) ijcnn1}
    \end{minipage}
    \caption{Evaluation of weighted ``cost`` performance under varying weights for False Positives and False Negatives.}
    \label{weight_cost}
    \vspace{-0.1in}
\end{figure}

\vspace{-0.1in}
\subsection{Evaluation of Algorithm properties}
\vspace{-0.05in}
We have evaluated the performance of proposed algorithms in previous experiments, where promising results confirm their great superiority. Next, we are eager to examine their unique properties, including the influence of learning rate, regularized parameter, updating rule, online estimation and generalization ability. These examinations contribute to better understanding and applications of proposed methods. For simplicity, all experiments are based on $sum$ metric, and every experiment only considers one objective or variable, while all other variable settings are fixed and similar with before experiments.

\vspace{-0.05in}
\subsubsection{Evaluation of Learning Rate}
\vspace{-0.05in}
In this subsection, we evaluate the influence of learning rate. In detail, we examine the $sum$ performances of proposed methods with different learning rates $\eta$ from $[10^{-4}, 10^{-3}, ..., 10^{3},10^{4}]$.

In Fig. 5, we find that ACOG algorithms would achieve relatively higher result, when we choose proper learning rate (i.e. relatively higher $\eta$ in general). This is easy to understand because the values of covariance matrix $\Sigma$ are normally small. Specifically, when a misclassification happened at time $t$, we update the predictive vector $\mu$ by $\mu_{t+1} = \mu_{t} + \eta\Sigma_{t+1}g_t$, where $g_t = \partial\ell_t(\mu_t)$. Because the values of covariance matrix $\Sigma$ are normally small, the values of $\Sigma_{t+1}g_t$ thus are small. So if we want to obtain excellent performance, it would be better to choose properly higher learning rates as updating steps.

Moreover, we find the proposed methods with objective function $\ell^{II}(w;(x,y))$ can achieve relatively higher performance than the methods with $\ell^{I}(w;(x,y))$, which means that ACOG-II and ACOG-II$_{diag}$ are more robust to different learning rate $\eta$ and consequently have a wider parameter choice space.

\begin{figure}
    \begin{minipage}{0.5\linewidth}
      \centerline{\includegraphics[width=4.5cm]{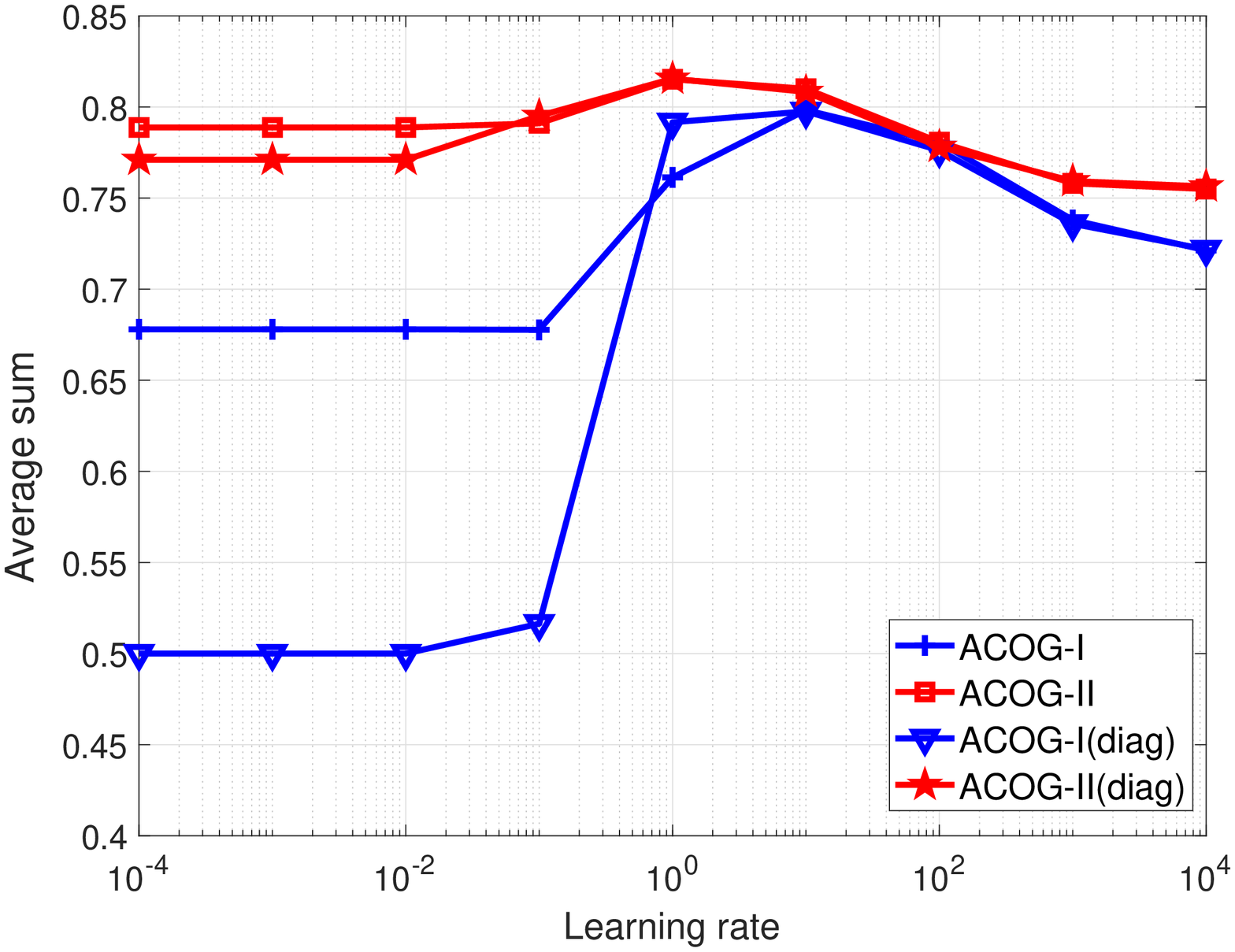}}
      \centerline{(a) a9a}
    \end{minipage}
    \hfill
    \begin{minipage}{0.5\linewidth}
      \centerline{\includegraphics[width=4.5cm]{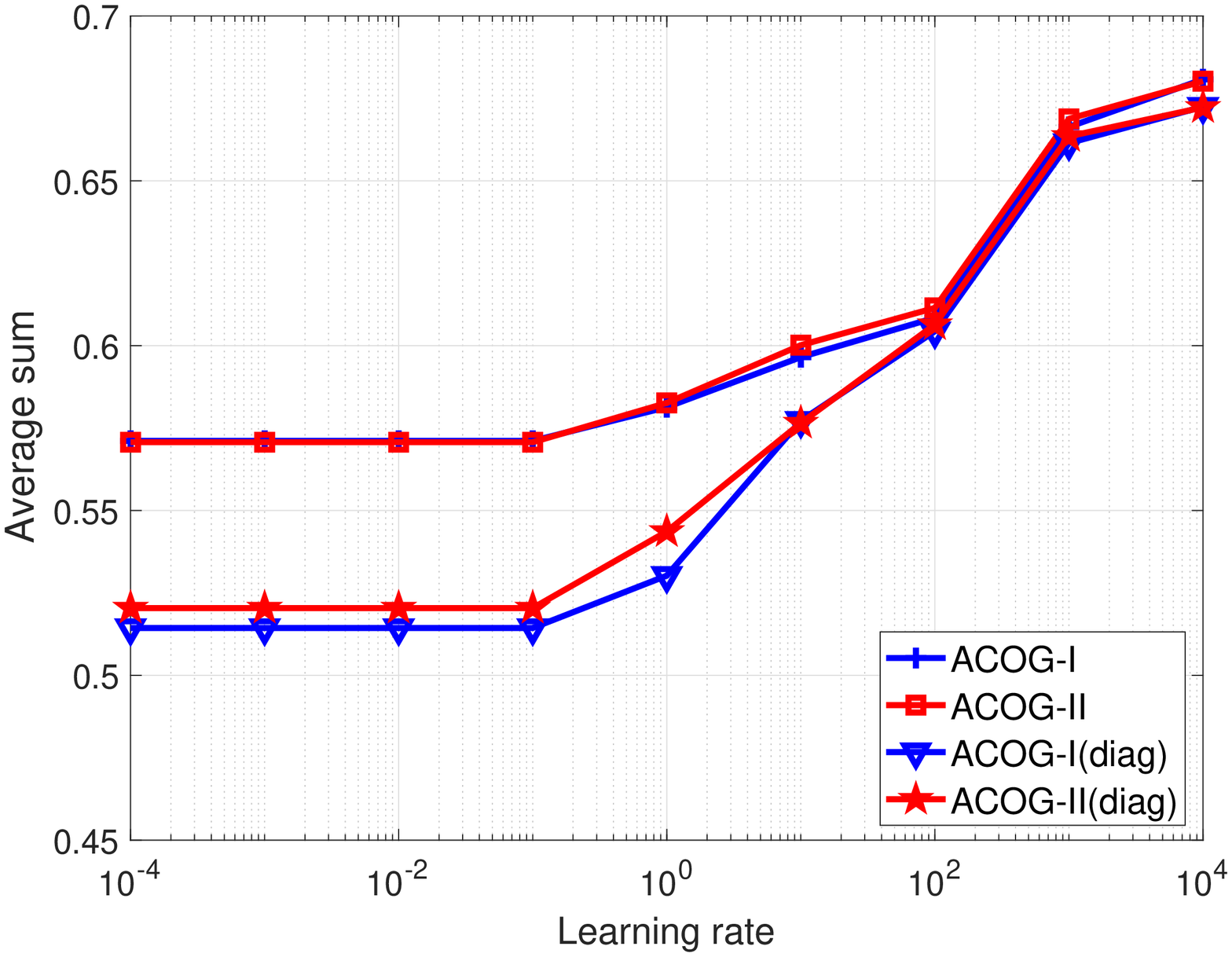}}
      \centerline{(b) covtype}
    \end{minipage}
    \vfill
    \begin{minipage}{0.5\linewidth}
      \centerline{\includegraphics[width=4.5cm]{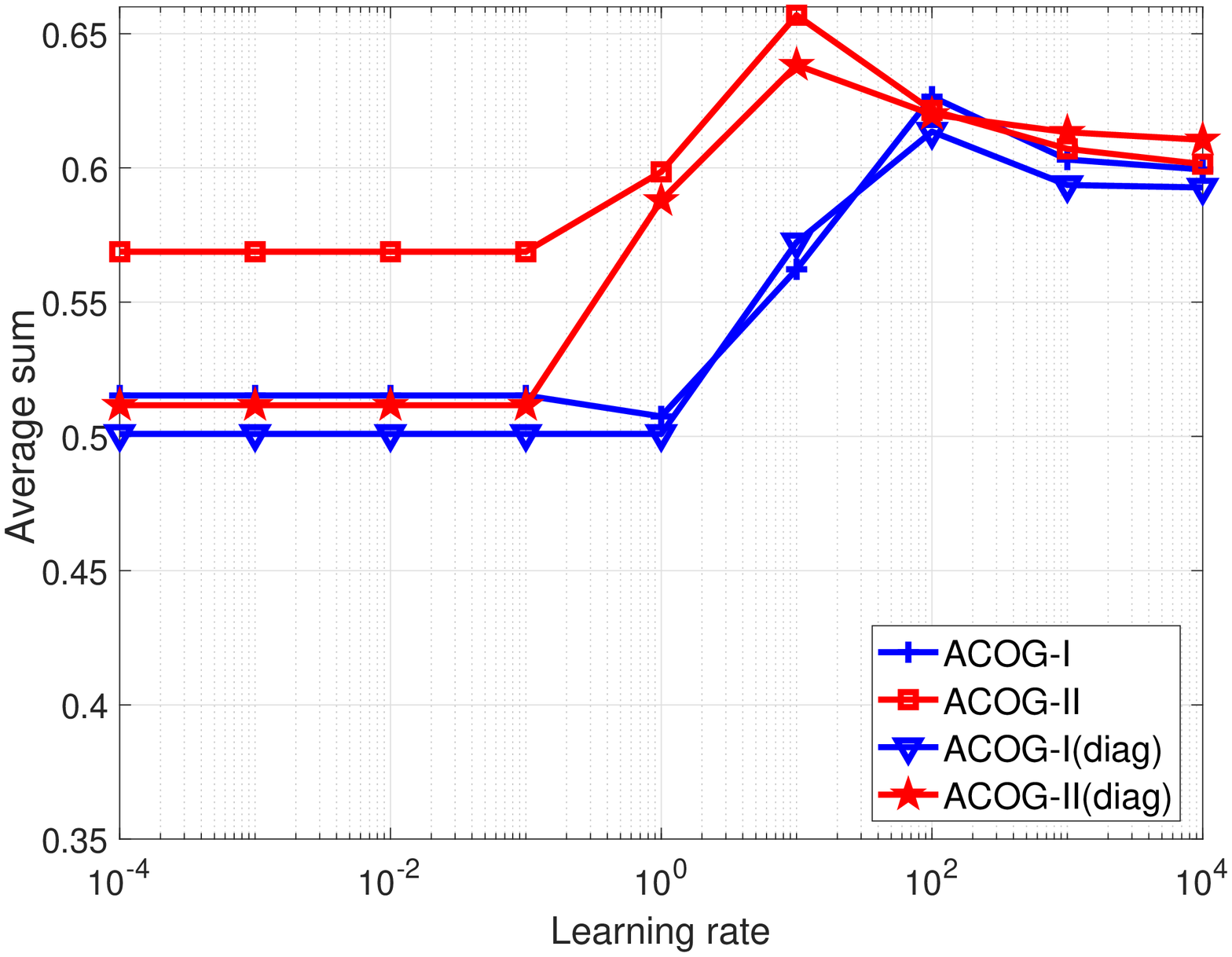}}
      \centerline{(c) german}
    \end{minipage}
    \hfill
    \begin{minipage}{0.5\linewidth}
      \centerline{\includegraphics[width=4.5cm]{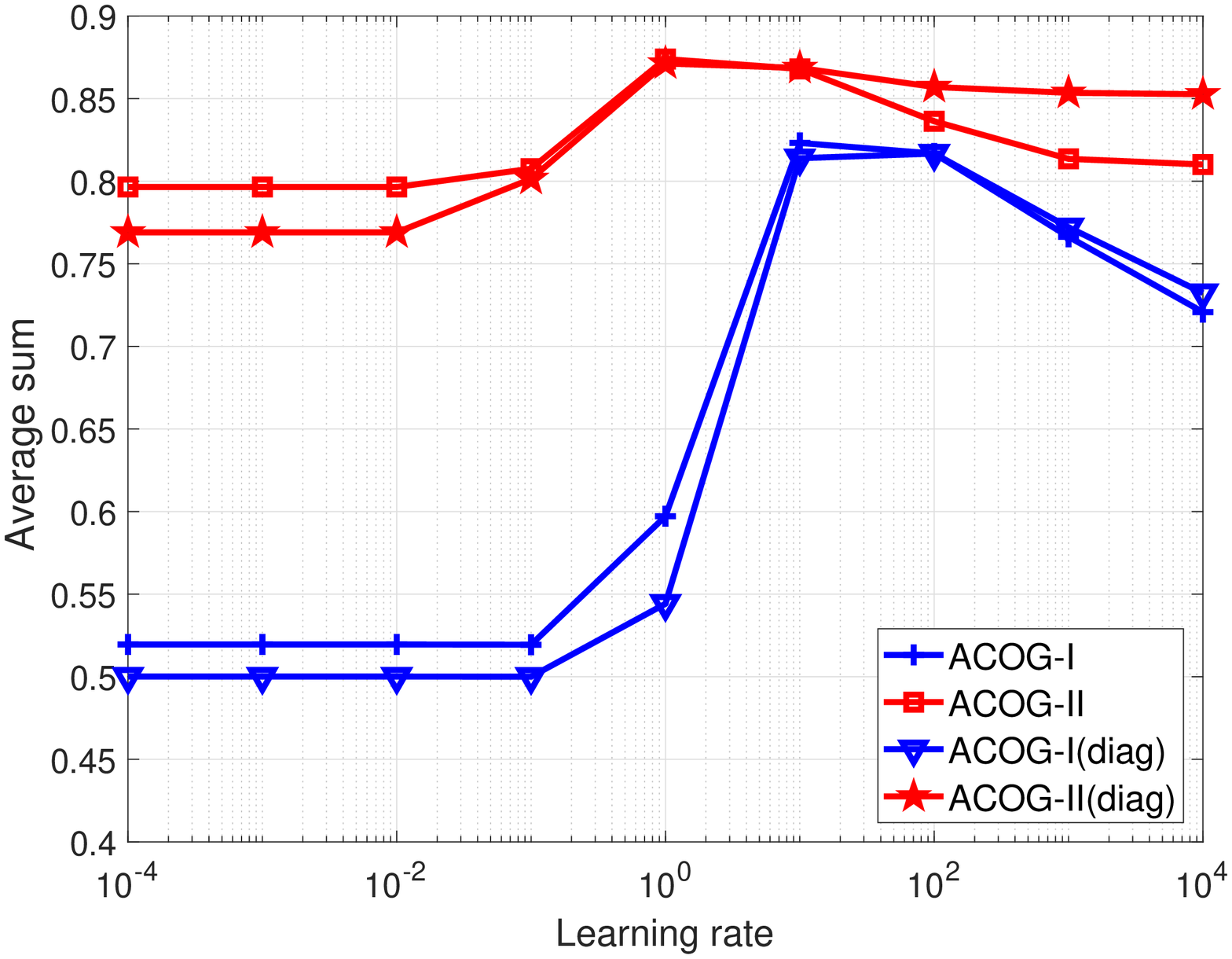}}
      \centerline{(d) ijcnn1}
    \end{minipage}

    \caption{Performance under varying learning rates.}
    \label{lr_sum}
\end{figure}
\vspace{-0.05in}
\subsubsection{Evaluation of Regularized Parameter}
\vspace{-0.05in}
Now, we aim to examine the influence of regularized parameters on our proposed algorithms.

When the learner makes a mistake, we update the covariance matrix $\Sigma$ by $\Sigma_{t+1} = \Sigma_{t} - \frac{\Sigma_{t}x_tx_t^\top\Sigma_{t}}{\gamma+x_t^\top\Sigma_{t}x_t}$ with default regularized parameter $\gamma$ as 1. However, the rationality of this setting is not verified. Thus, we examine the performance of our algorithms with different regularized parameters $\gamma$ from $[10^{-4}, 10^{-3}, ..., 10^{3},10^{4}]$ for $sum$ metrics.

The results in Fig. 6 show that the optimal parameter normally is different according to datasets; while in most cases, the setting $\gamma=1$ can achieve the best or fairly good results. This discovery confirms the practical value of our algorithms with default settings.

\begin{figure}
    \begin{minipage}{0.5\linewidth}
      \centerline{\includegraphics[width=4.5cm]{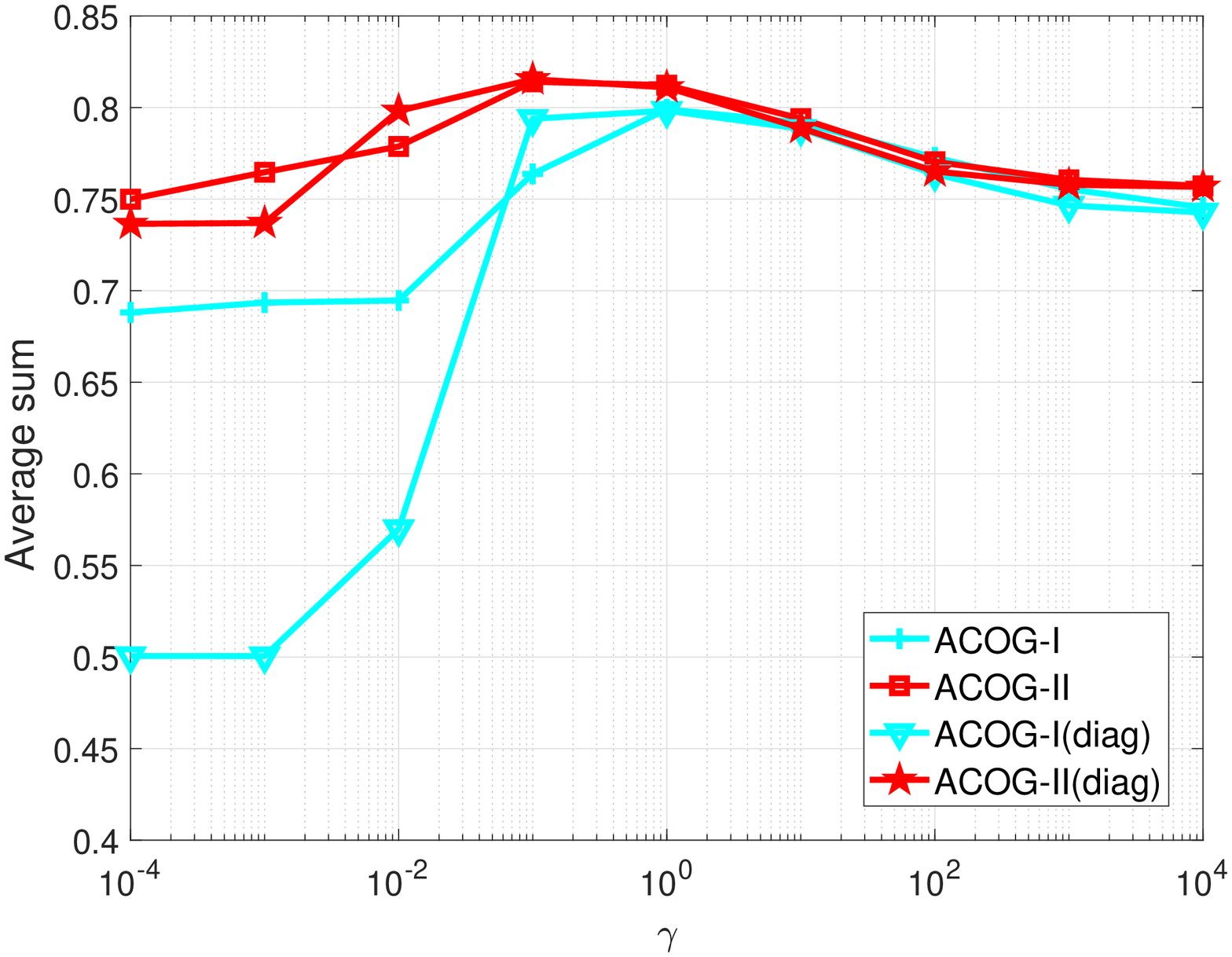}}
      \centerline{(a) a9a}
    \end{minipage}
    \hfill
    \begin{minipage}{0.5\linewidth}
      \centerline{\includegraphics[width=4.5cm]{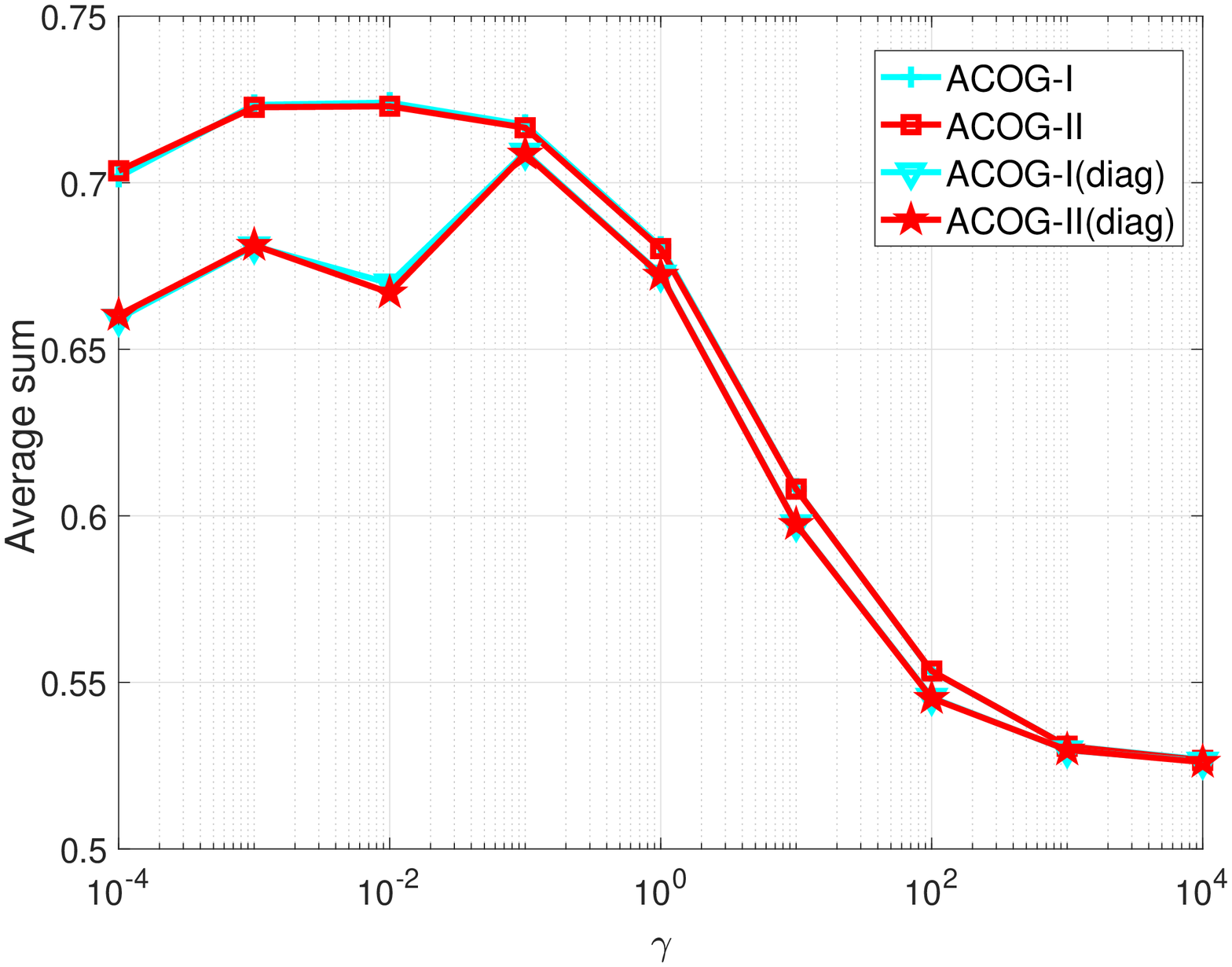}}
      \centerline{(b) covtype}
    \end{minipage}
    \vfill
    \begin{minipage}{0.5\linewidth}
      \centerline{\includegraphics[width=4.5cm]{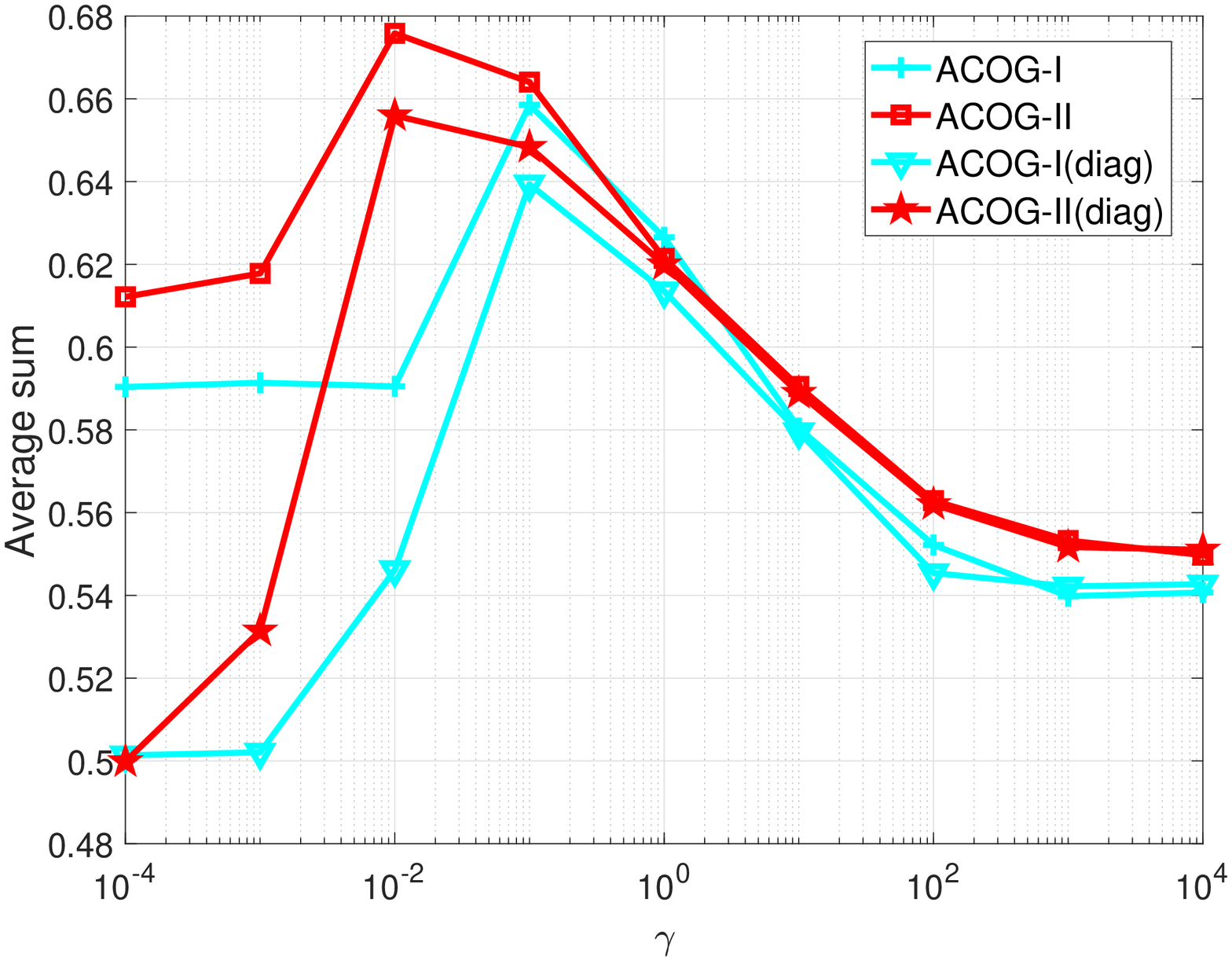}}
      \centerline{(c) german}
    \end{minipage}
    \hfill
    \begin{minipage}{0.5\linewidth}
      \centerline{\includegraphics[width=4.5cm]{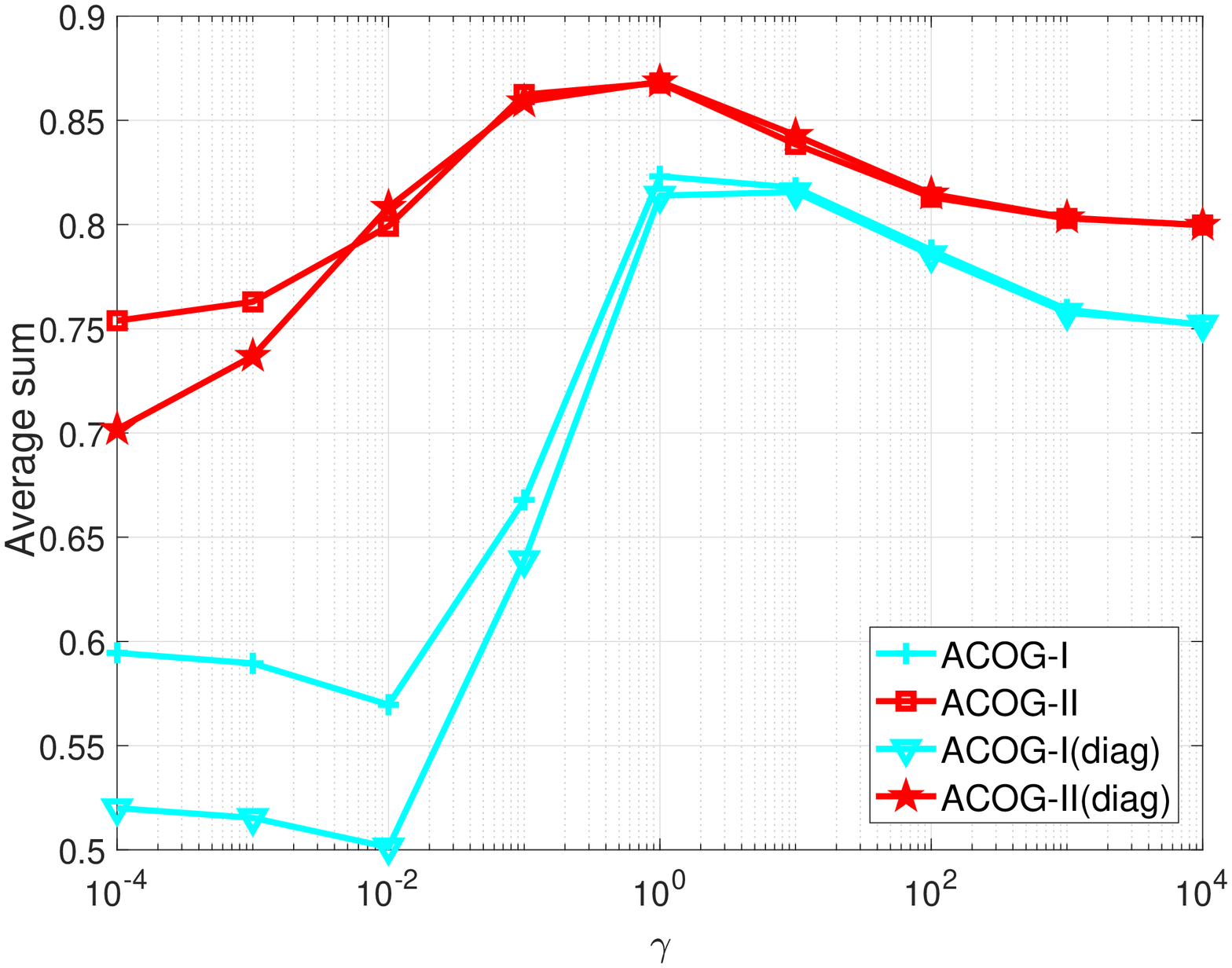}}
      \centerline{(d) ijcnn1}
    \end{minipage}

    \caption{Performance under different regularized parameters.}
    \label{gamma_sum}
    \vspace{-0.1in}
\end{figure}

\vspace{-0.05in}
\subsubsection{Evaluation of Updating Rule}
\vspace{-0.05in}
As mentioned in Section 2, the predictive vector $\mu$ is updated by $\mu_{t+1} = \mu_{t} + \eta\Sigma_{t+1}g_t$, which is different from AROW where the updating rule for $\mu$ relies on the old $\Sigma_{t}$. In this subsection, we would like to evaluate the difference between two updating rules based on $sum$ metrics for proposed methods, where the invariant versions (i.e., green line in Fig. 7) depending on old $\Sigma_t$.

From Fig. 7, we find that although the difference between two updating rules is not obvious, the performance of $\Sigma_{t+1}$ versions slightly exceed $\Sigma_{t}$ versions, which is consistent with our analysis in Section 2.

\begin{figure}
    \begin{minipage}{0.5\linewidth}
      \centerline{\includegraphics[width=4.5cm]{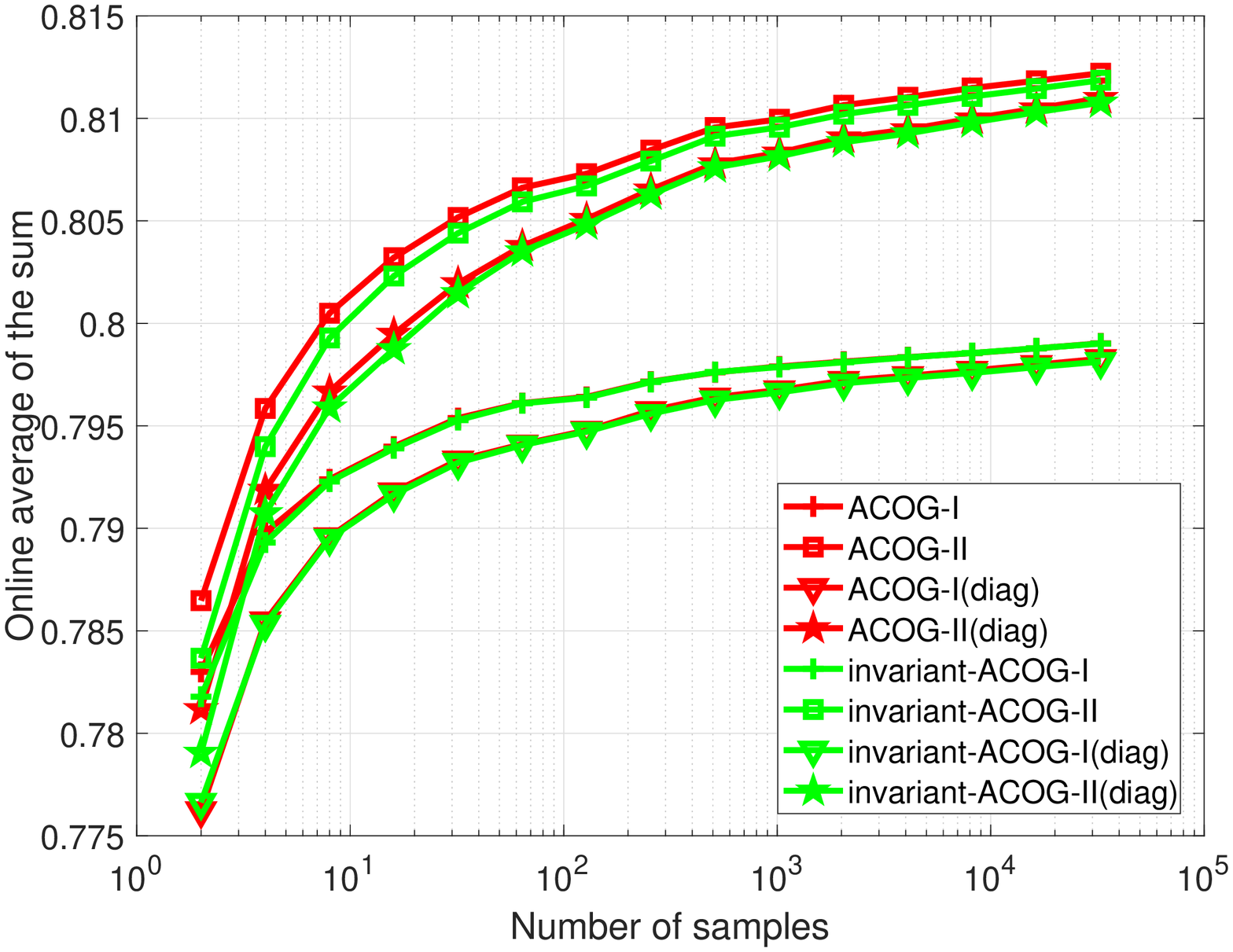}}
      \centerline{(a) a9a}
    \end{minipage}
    \hfill
    \begin{minipage}{0.5\linewidth}
      \centerline{\includegraphics[width=4.5cm]{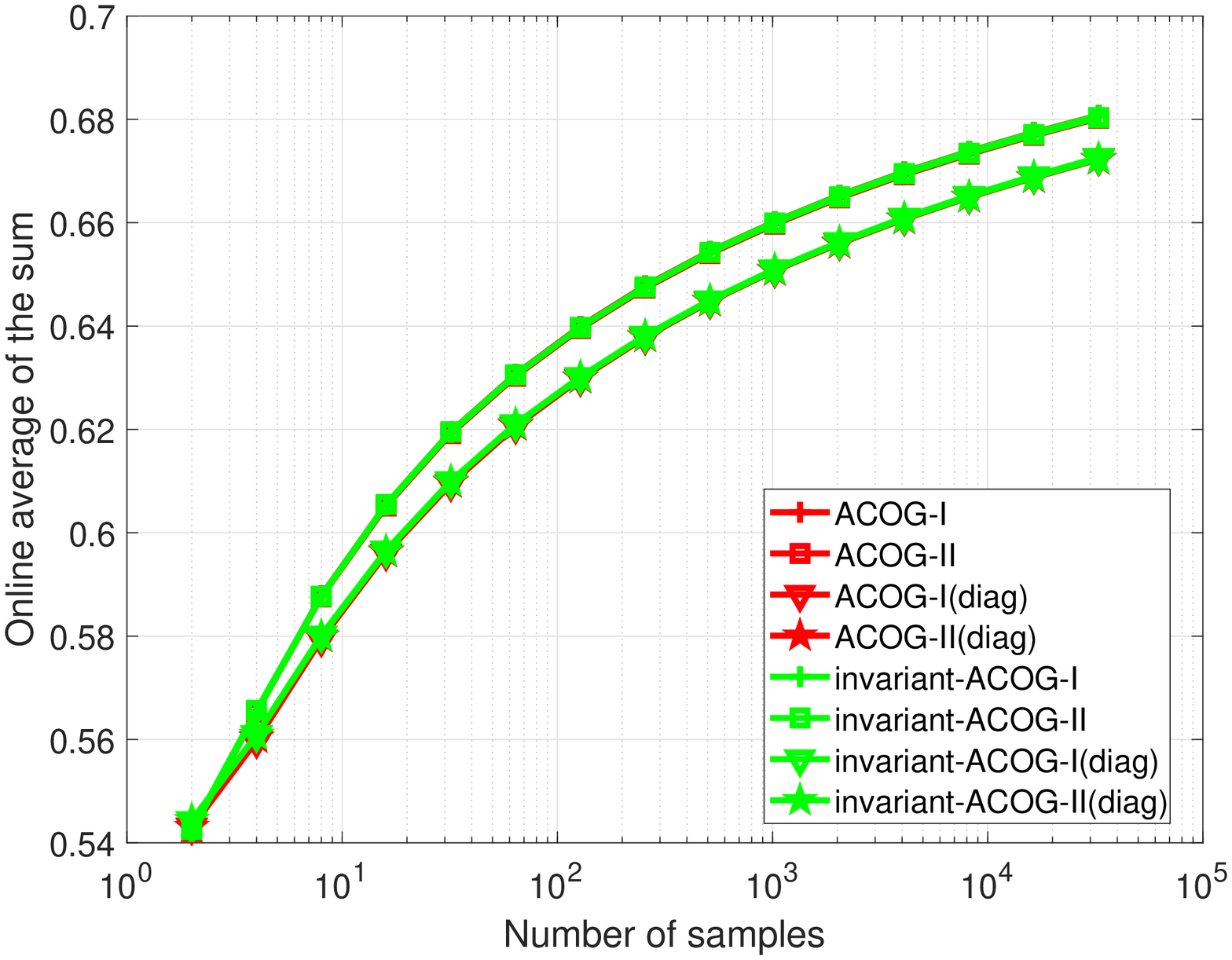}}
      \centerline{(b) covtype}
    \end{minipage}
    \vfill
    \begin{minipage}{0.5\linewidth}
      \centerline{\includegraphics[width=4.5cm]{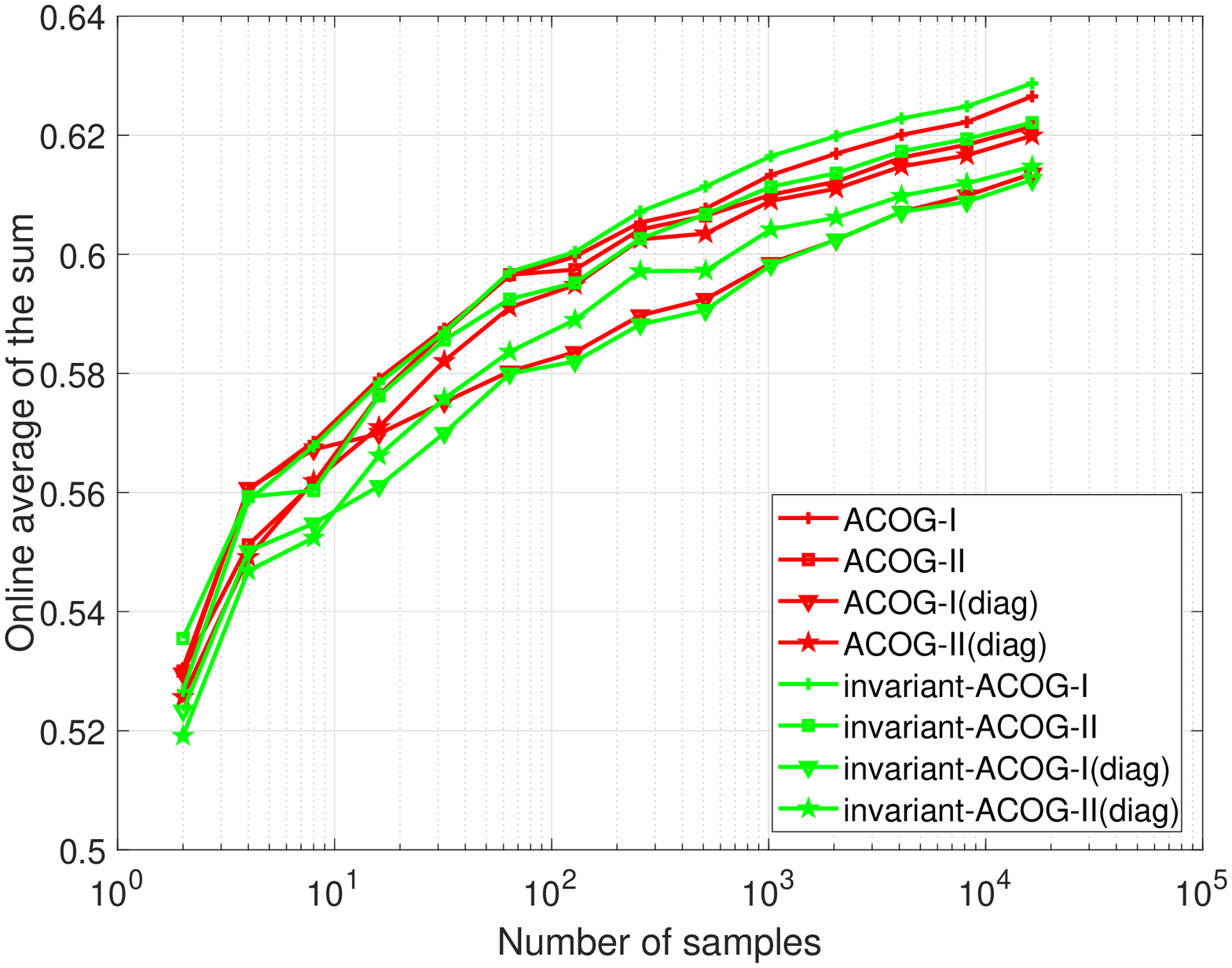}}
      \centerline{(c) german}
    \end{minipage}
    \hfill
    \begin{minipage}{0.5\linewidth}
      \centerline{\includegraphics[width=4.5cm]{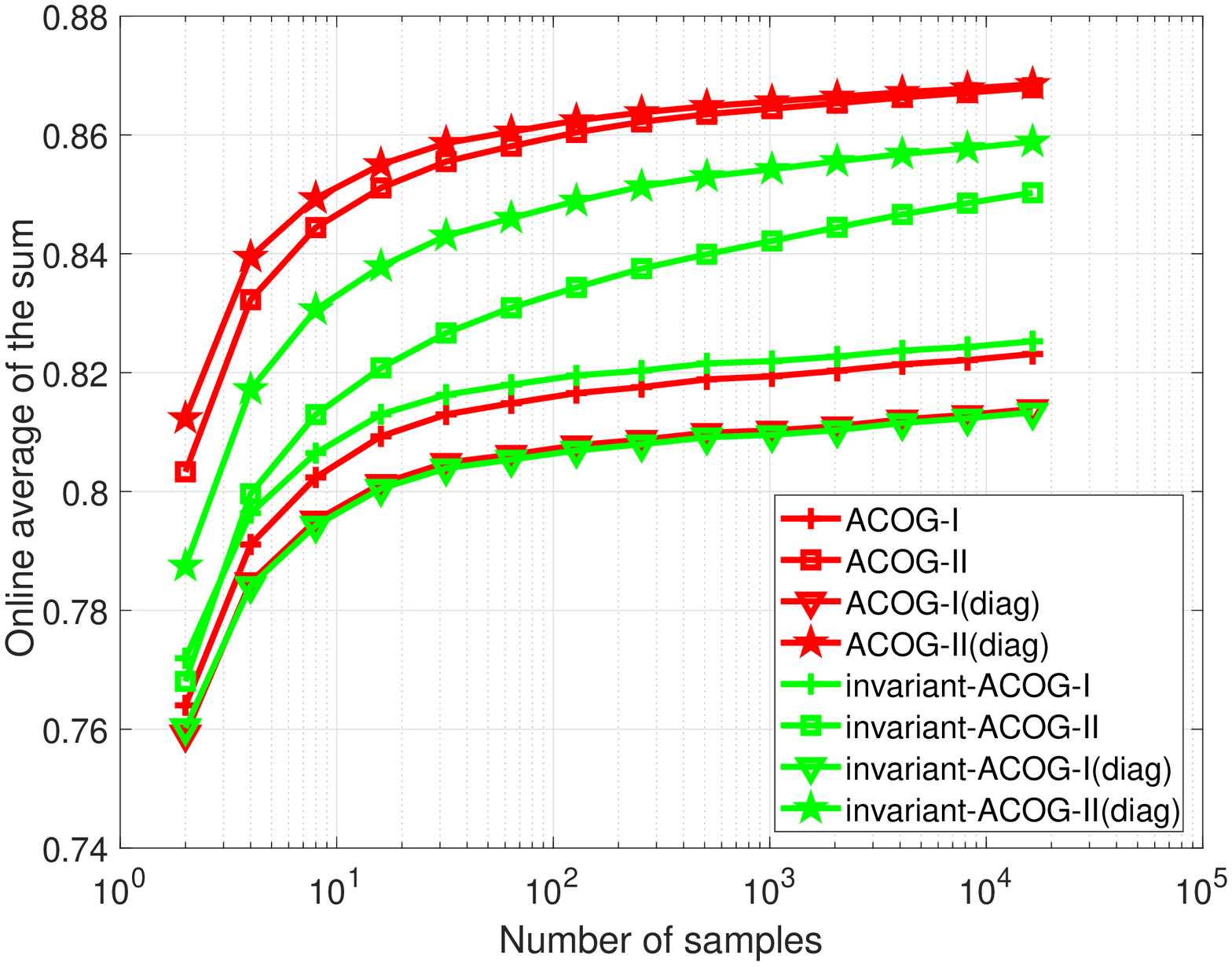}}
      \centerline{(d) ijcnn1}
    \end{minipage}
    \caption{Evaluation of updating rules.}
    \label{update_sum}
    \vspace{-0.11in}
\end{figure}

\vspace{-0.05in}
\subsubsection{Evaluation of Online Estimation of $\frac{T_n}{T_p}$}
\vspace{-0.05in}
In the \textbf{remark} of Algorithm 1, we analyzed the parameter $\rho = \frac{\eta_pT_n}{\eta_nT_p}$ for ACOG$_{sum}$ algorithms, where the main question is that the value of $T_p$ and $T_n$ cannot be obtained in advance on real-world online learning.

Thus, we want to evaluate the influence of online estimation $\frac{T_n}{T_p}$ on $sum$ performance, compared with the original algorithms. We adopt the widely used laplace estimation here, which estimates $\frac{T_n}{T_p}$ by $\frac{t_n+1}{t_p+1}$, where $t_p$ and $t_n$ represent the number of positive samples and negative samples that have been seen, respectively.

Fig. 8 shows the performance of online estimation. We find that the online laplace estimation performs quite similar results with the original one. This discovery validates the practical value of the proposed ACOG$_{sum}$ algorithms.
\begin{figure}
    \begin{minipage}{0.5\linewidth}
      \centerline{\includegraphics[width=4.8cm]{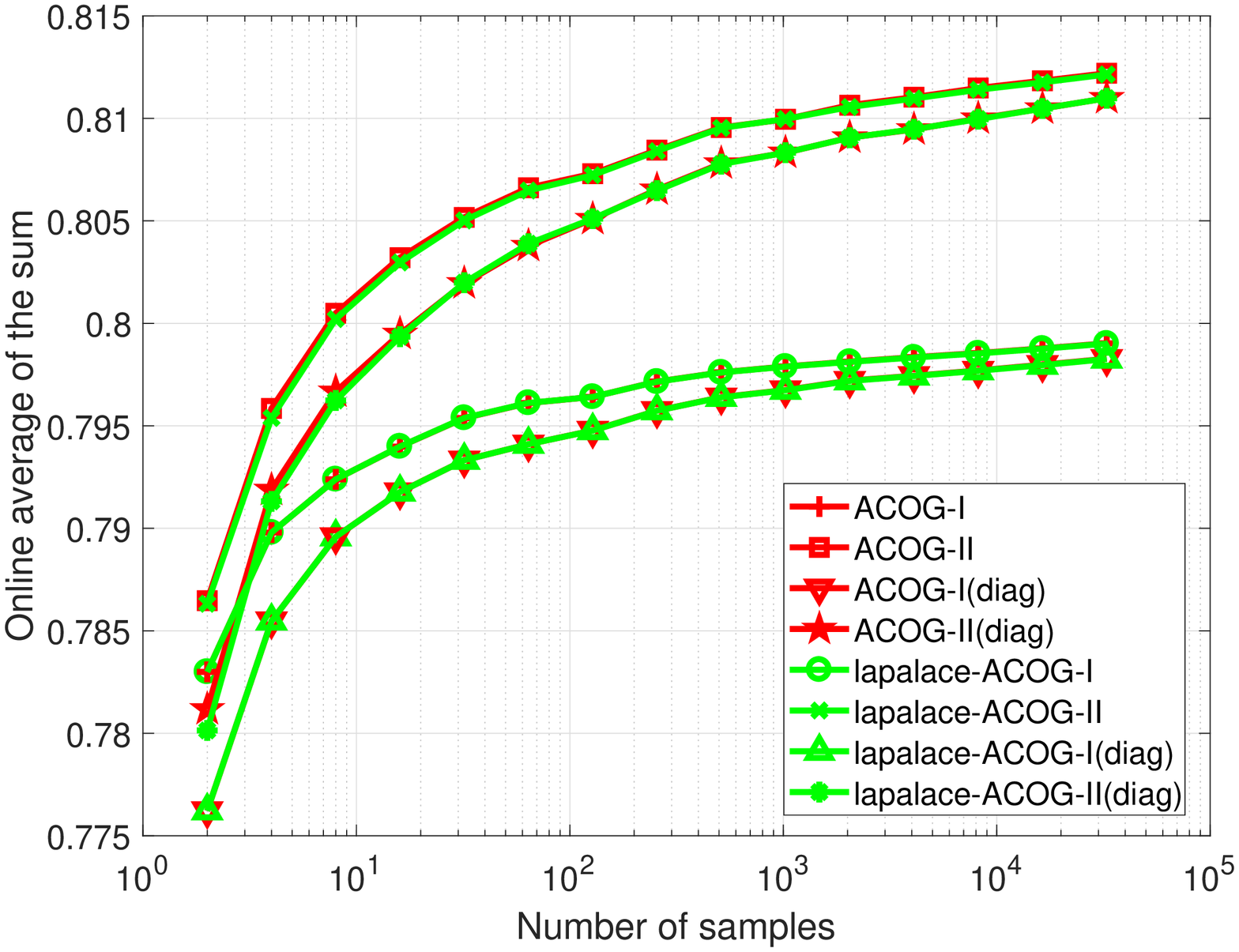}}
      \centerline{(a) a9a}
    \end{minipage}
    \hfill
    \begin{minipage}{0.5\linewidth}
      \centerline{\includegraphics[width=4.8cm]{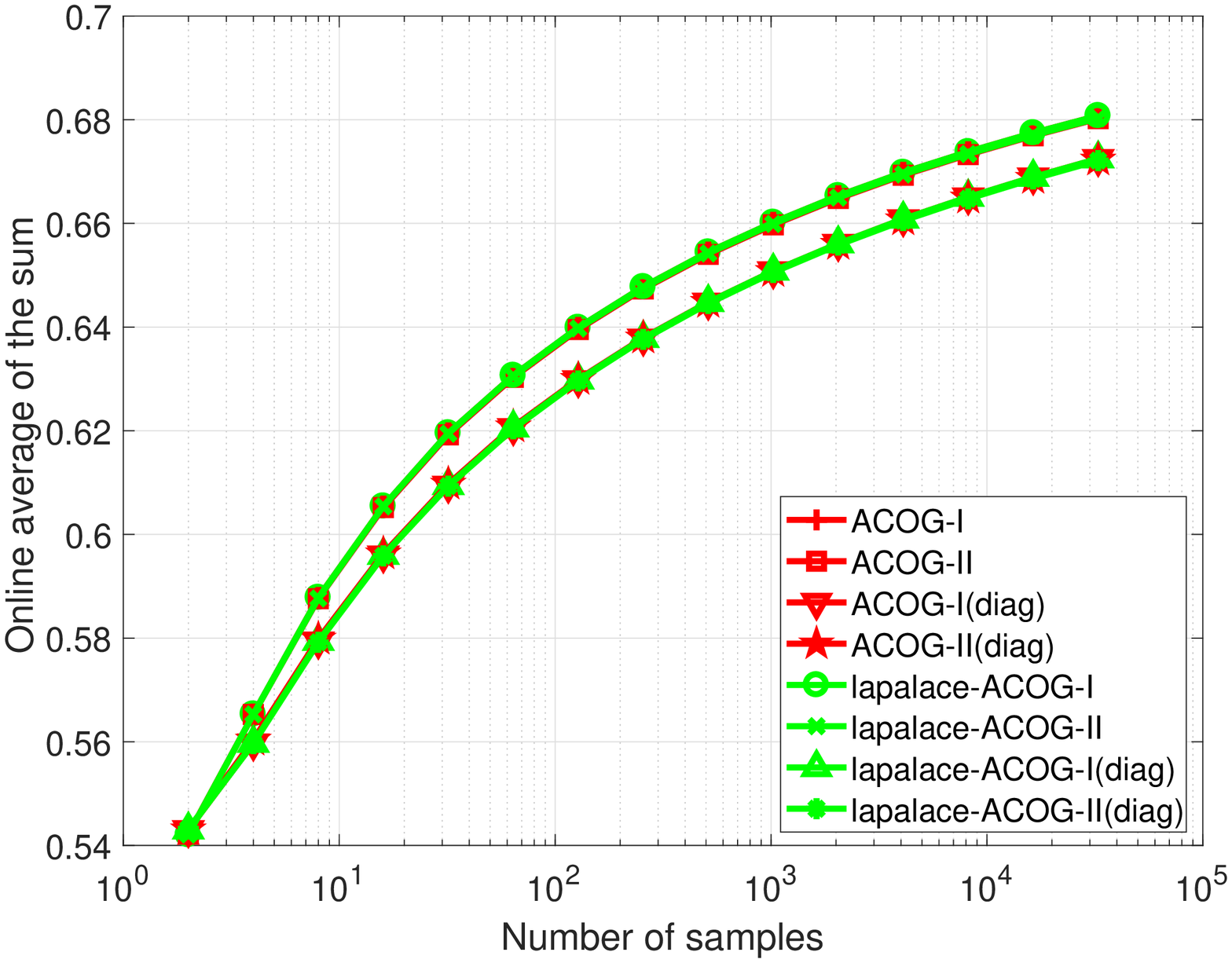}}
      \centerline{(b) covtype}
    \end{minipage}
    \vfill
    \begin{minipage}{0.5\linewidth}
      \centerline{\includegraphics[width=4.8cm]{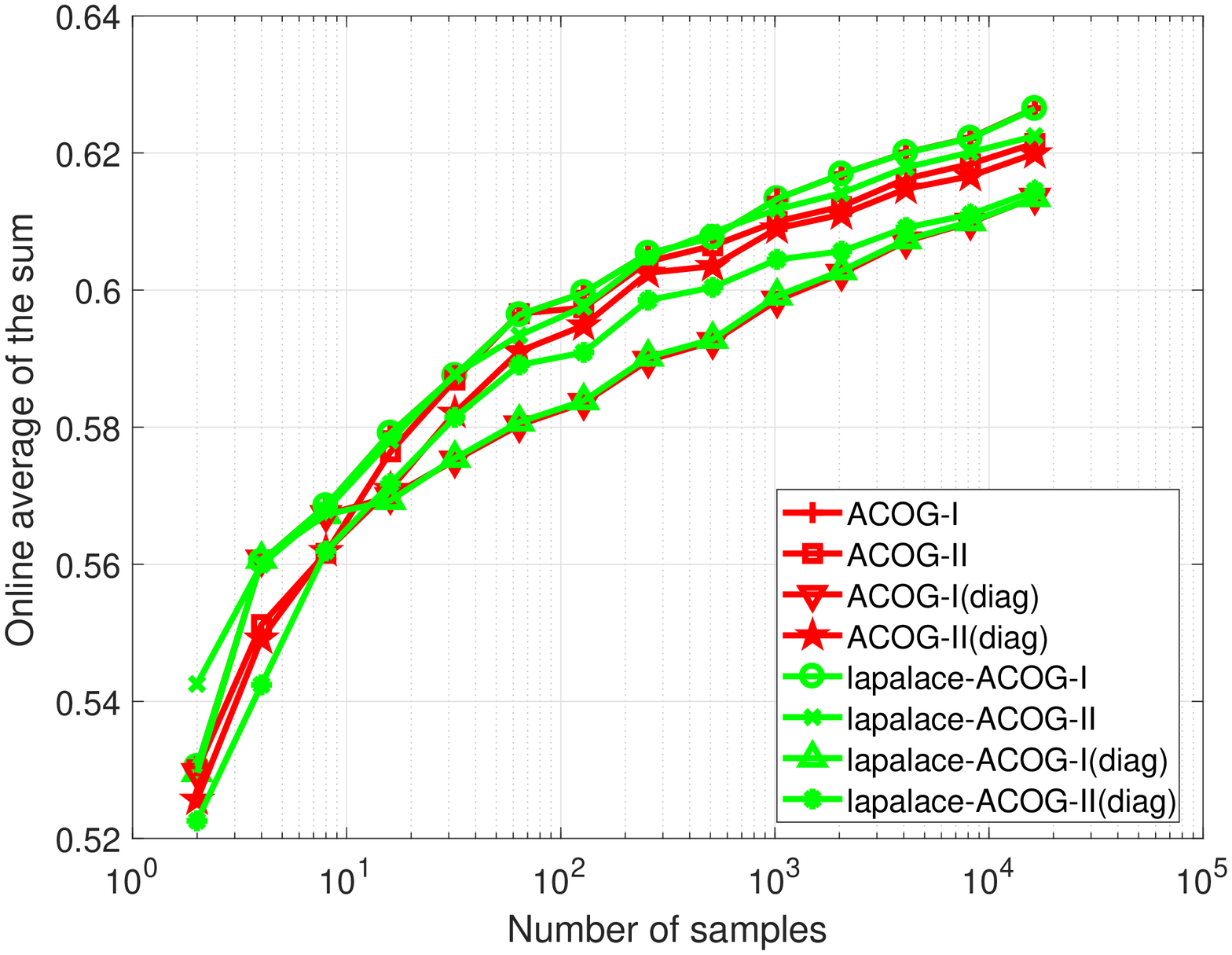}}
      \centerline{(c) german}
    \end{minipage}
    \hfill
    \begin{minipage}{0.5\linewidth}
      \centerline{\includegraphics[width=4.8cm]{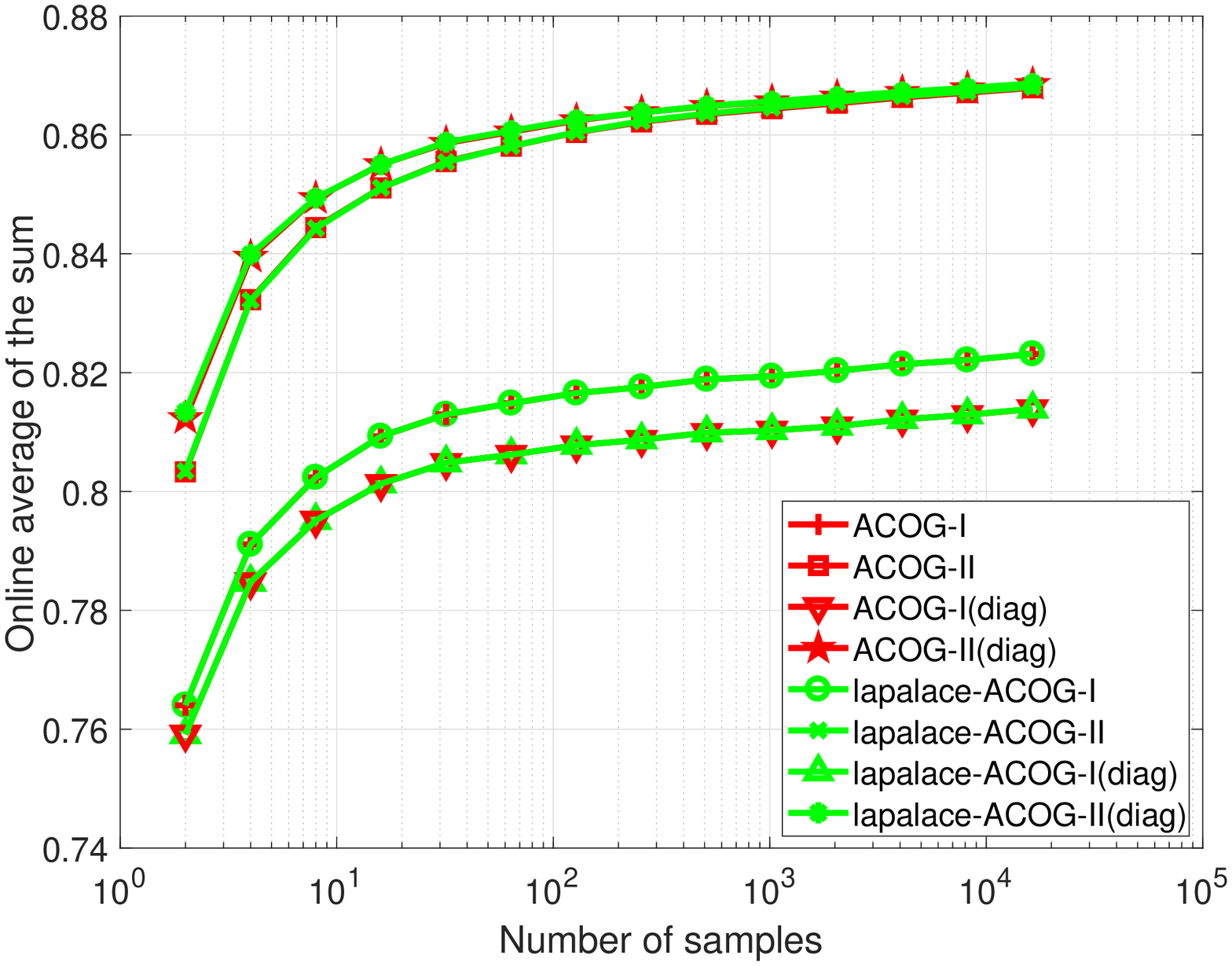}}
      \centerline{(d) ijcnn1}
    \end{minipage}

    \caption{Evaluation of online estimation of $\frac{T_n}{T_p}$.}
    \label{sum}
    \vspace{-0.1in}
\end{figure}

\subsubsection{Evaluation of Generalization Ability}

Then, we evaluate the generalization ability of proposed methods, which may exist problems when converting an online algorithm to a batch training approach. We use 5-fold cross-validation for better validation of the general performance.

Table 3 summary the consequences on $sum$ metrics, in which we discover that our proposed algorithms achieve the best among all algorithms on all datasets. This discovery indicates that our proposed methods have a strong generalized ability and can be regarded as a potentially useful tool to train large-scale cost-sensitive models.

\begin{table}
	\caption{Evaluation of generalization ability with $sum$}
    \vspace{-0.1in}
    \begin{center}
    \scalebox{1}{
	\begin{tabular}{|l|c|c|c|c|}\hline
		Algorithm &a9a &covtype   & german &ijcnn1  \\\hline \hline
        Perceptron   &68.649 	 &51.553  	   &53.737 	   &70.045  \\
        ROMMA        &72.467 	&67.059  	   &58.614 	 	      &76.818  	\\
        PA-I         &71.986 	 &51.283      &51.363       &70.410 		\\
        PAUM         &79.323 	 &53.354          &52.126 	     &82.012 	 	 \\
        CPA$_{PB}$  &73.668 	&51.279 	 &52.768 	 &73.942 	\\
        AROW  &75.961 	 &64.928   &54.575 &67.642 	\\
        COG-I  &79.705 	 &53.354 	&52.258 	 &82.012 	 \\
        COG-II  &78.559 	 &68.897   	 &50.784 	 &82.849 		\\
        ACOG-I  &80.026 	 &\textbf{72.428}		  &62.954   &82.926 	 \\
        ACOG-II  &\textbf{81.630}	   &\textbf{72.632}	  	  &60.928   &\textbf{87.730}	 	\\
        ACOG-I$_{diag}$  &80.118 	&71.051 	 	 &\textbf{64.389}	 &82.334 	\\
        ACOG-II$_{diag}$  &\textbf{81.752}	 &71.311 	  &\textbf{66.036}	  &\textbf{87.628}	\\
        \hline
	\end{tabular}
        }
    \end{center}
    \vspace{-0.15in}
\end{table}

\vspace{-0.05in}
\subsection{Performance and Efficiency of Sketched ACOG}

In the previous experiments, the evaluations of the proposed ACOG algorithms have shown promising results. However, we can find the implementation of ACOG is time consuming when facing high-dimensional datasets, because of the updating step for covariance matrix. As a result, it is difficult for engineers to address the real-world tasks with quite large-scale datasets.

A simple solution to this question is to implement the diagonal version of ACOG, and then enjoy linear time complexity. However, the gain of diagonal ACOG is at the cost of lower performance, because it abandons the correlation information between sample dimensions, which is quite important and indispensable for datasets with strong inner-correlation. Thus, for better trade off between performance and time efficiency, we propose the Sketched ACOG (named SACOG) and its sparse version (named SSACOG).

In this section, we first evaluate our sketched algorithms with several baseline algorithms: (1) ``COG-I`` and ``COG-II``; (2) ``ACOG-I`` and ``ACOG-II``; (3) ``ACOG-I$_{diag}$`` and ``ACOG-II$_{diag}$``, where we adopt 4 relatively high-dimensional datasets from LIBSVM, which are higher than 45 dimensions as list in Table 4. After that, we examine the performance difference between SACOG and SSACOG.

For simplicity, we focus on the case that the sketch size $m$ is fixed as 5 for all sketched algorithms, although our methods can be easily generalized by setting different sketch sizes like \cite{luo2016efficient}. Moreover, the learning rate was selected from $[10^{-5},10^{-4},...,10^{5}]$, where other implementation details are similar with \cite{luo2016efficient}. In addition, all experimental settings for other algorithms are same as previous experiments.

\begin{table}[h]\label{5}
	\caption{\small{Datasets for Evaluation of Sketched Algorithm}}
    \vspace{-0.12in}
    \begin{center}
	\begin{tabular}{|l|c|c|c|}\hline
		Dataset &\#Examples & \#Features & \#Pos:\#Neg  \\\hline \hline
        mushrooms & 8124 & 112 & 1:1.07  \\
        protein&17766&357&1:1.7\\
        usps&7291&256&1:5.11\\
        Sensorless & 58509 &48&1:10\\\hline
	\end{tabular}
    \end{center}
    \vspace{-0.1in}
\end{table}

\vspace{-0.1in}
\subsubsection{Evaluation of Weighted Sum Performance}
\vspace{-0.05in}

In this subsection, we would like to examine the performance and efficiency of our sketched algorithms, where we adopt the sparse version (SSACOG) rather than the original SACOG, which is more appropriate for real-world datasets.

The results are summarized in Fig. 9, Fig. 10 and Table 5 based on two metrics, from which we find that the proposed SSACOG is much faster than ACOG algorithms, while the performance of sketched algorithms is not affected too much and sometimes even better. In addition, the degree of efficiency optimization by sketching technique goes up along with the increase of data dimensions, which is consistent with the common sense.

Note that although the running time of SSACOG is slower than ACOG$_{diag}$, it enjoys higher performance due to the advantage of sufficient second-order information. This confirms the superiority of ACOG with sketching technique.

\begin{figure}
    \begin{minipage}{0.5\linewidth}
      \centerline{\includegraphics[width=4.5cm]{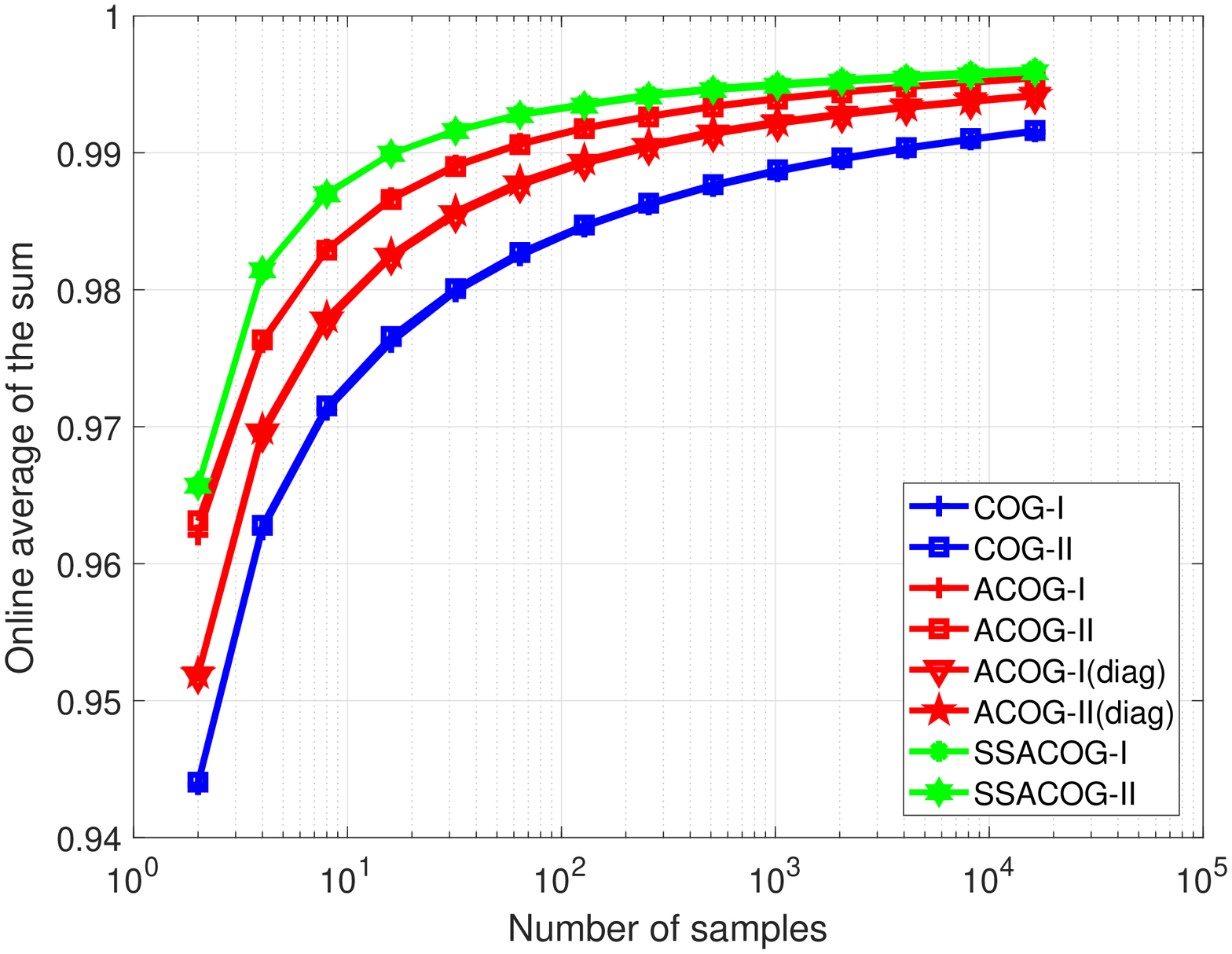}}
      \centerline{(a) mushrooms}
    \end{minipage}
    \hfill
    \begin{minipage}{0.5\linewidth}
      \centerline{\includegraphics[width=4.5cm]{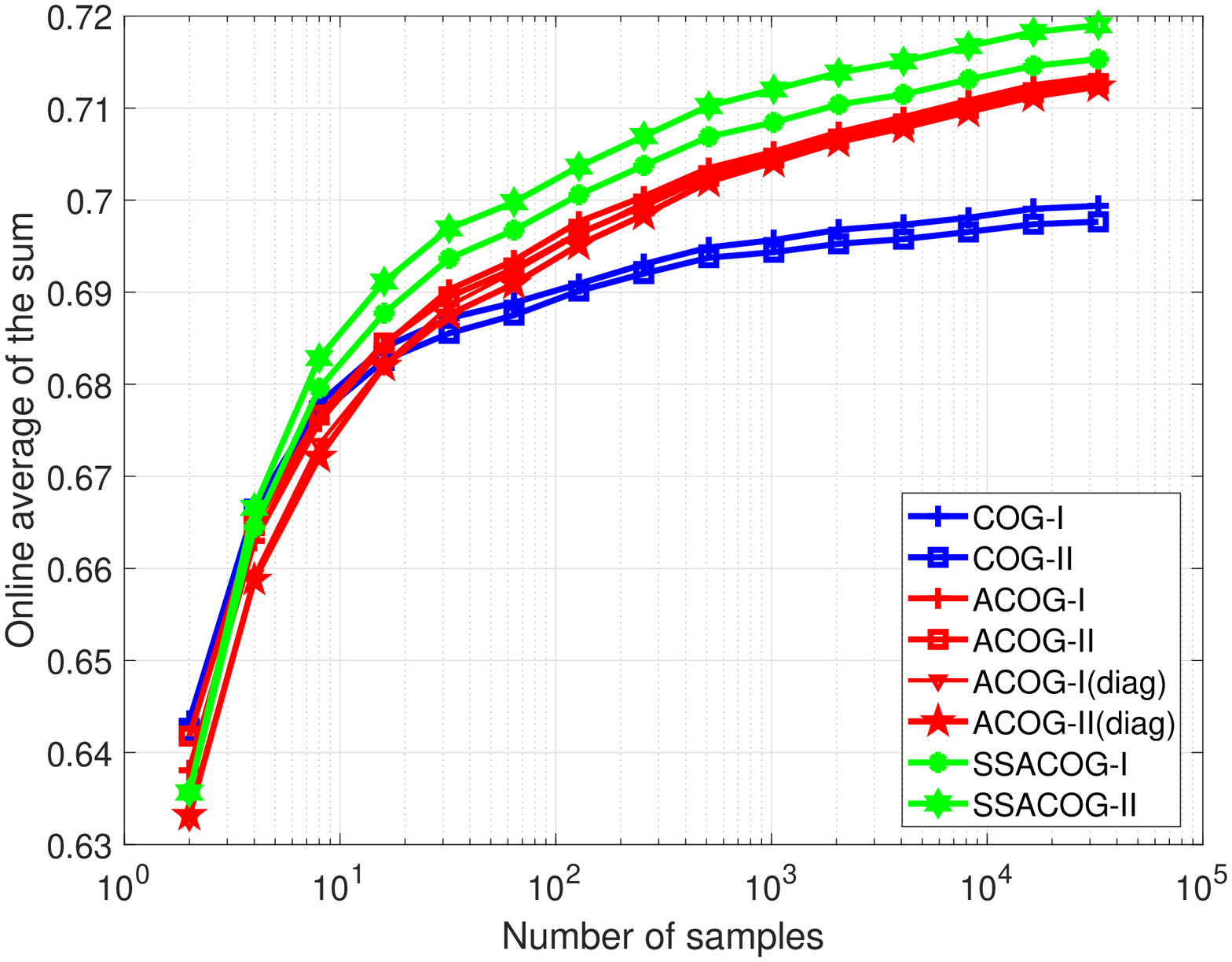}}
      \centerline{(b) protein}
    \end{minipage}
    \vfill
    \begin{minipage}{0.5\linewidth}
      \centerline{\includegraphics[width=4.5cm]{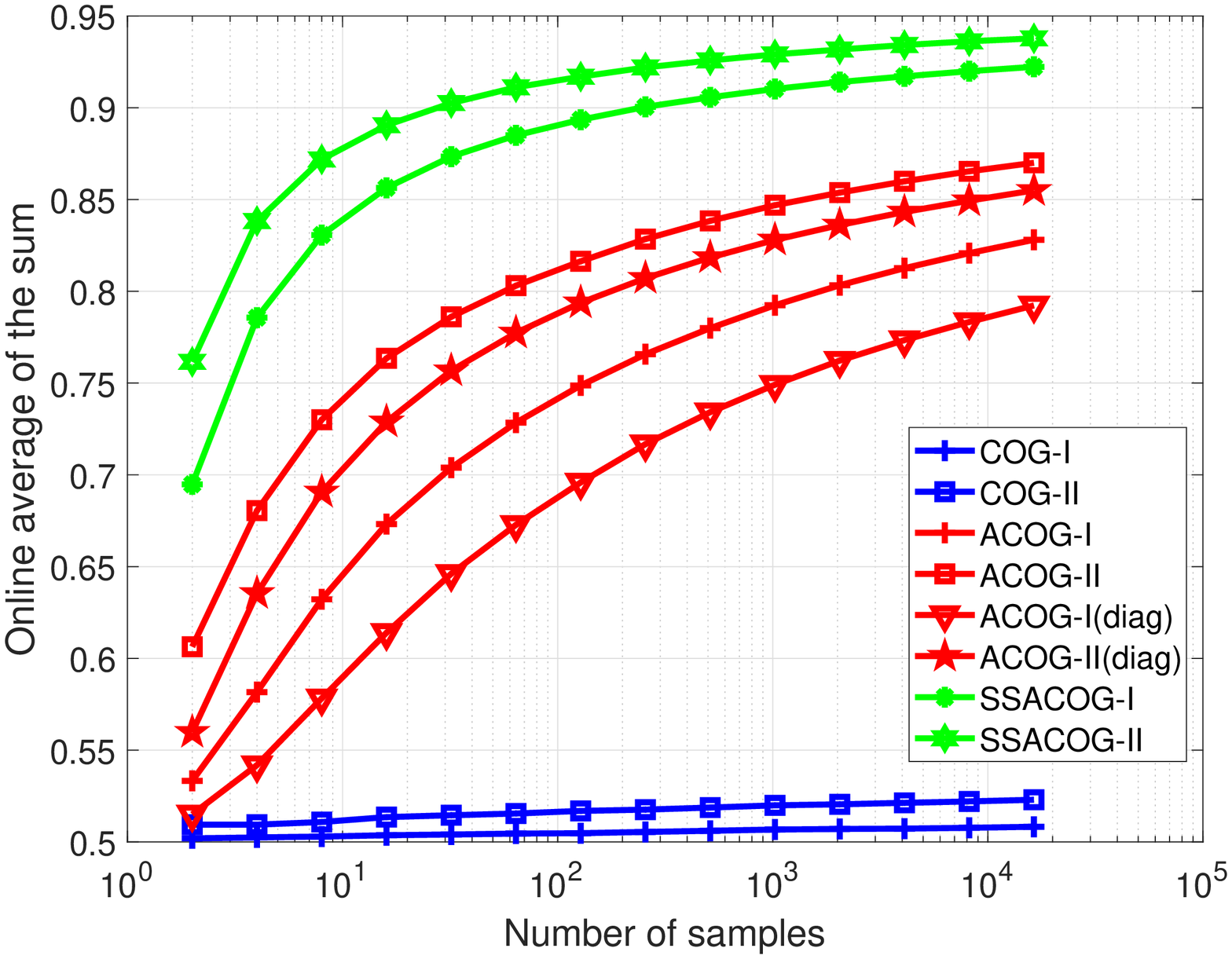}}
      \centerline{(c) Sensorless}
    \end{minipage}
    \hfill
    \begin{minipage}{0.5\linewidth}
      \centerline{\includegraphics[width=4.5cm]{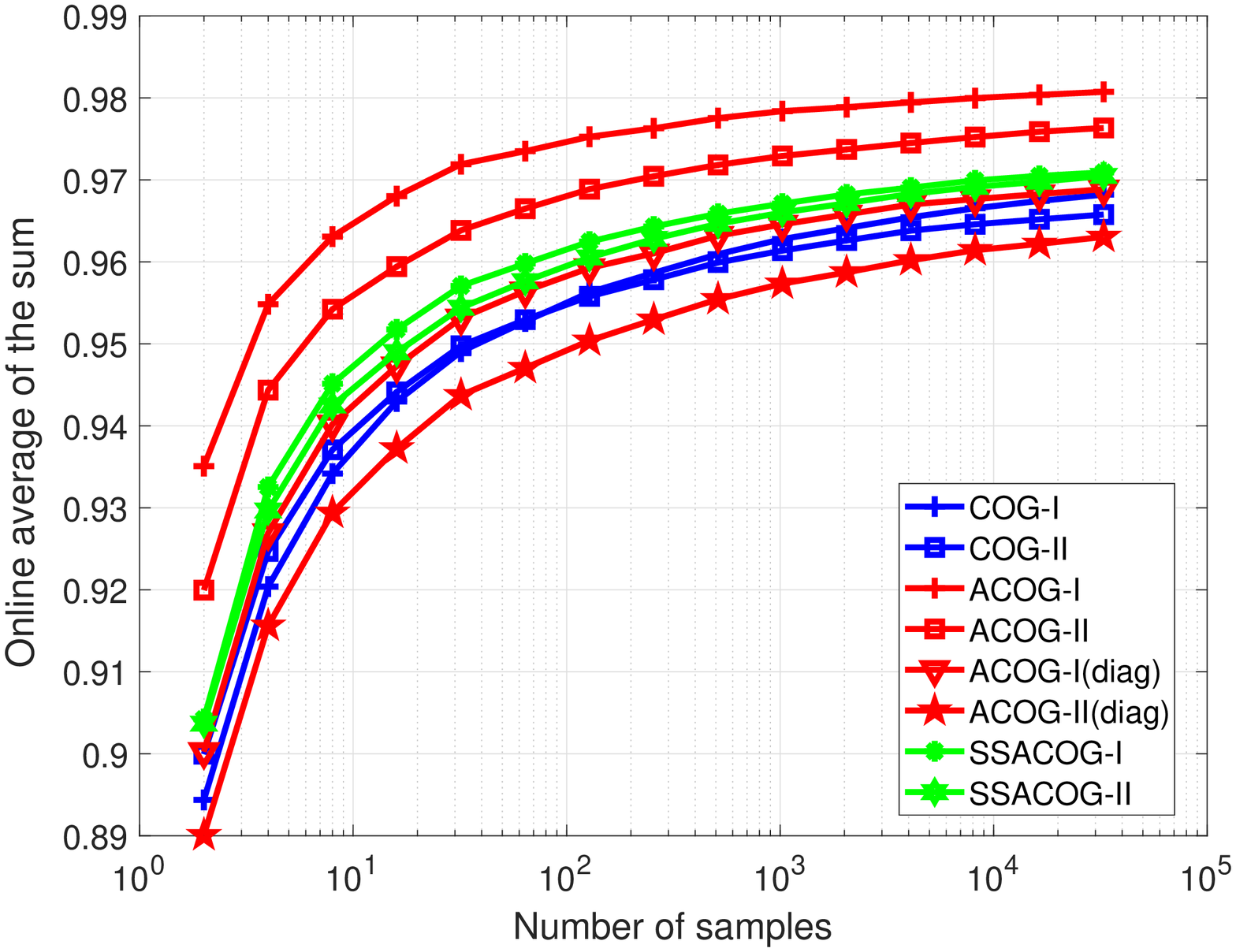}}
      \centerline{(d) usps}
    \end{minipage}
    \vspace{-0.08in}
    \caption{Weighted ``sum`` performance of SACOG.}
    \vspace{-0.12in}
\end{figure}

\begin{figure}[h]
    \begin{minipage}{0.5\linewidth}
      \centerline{\includegraphics[width=4.5cm]{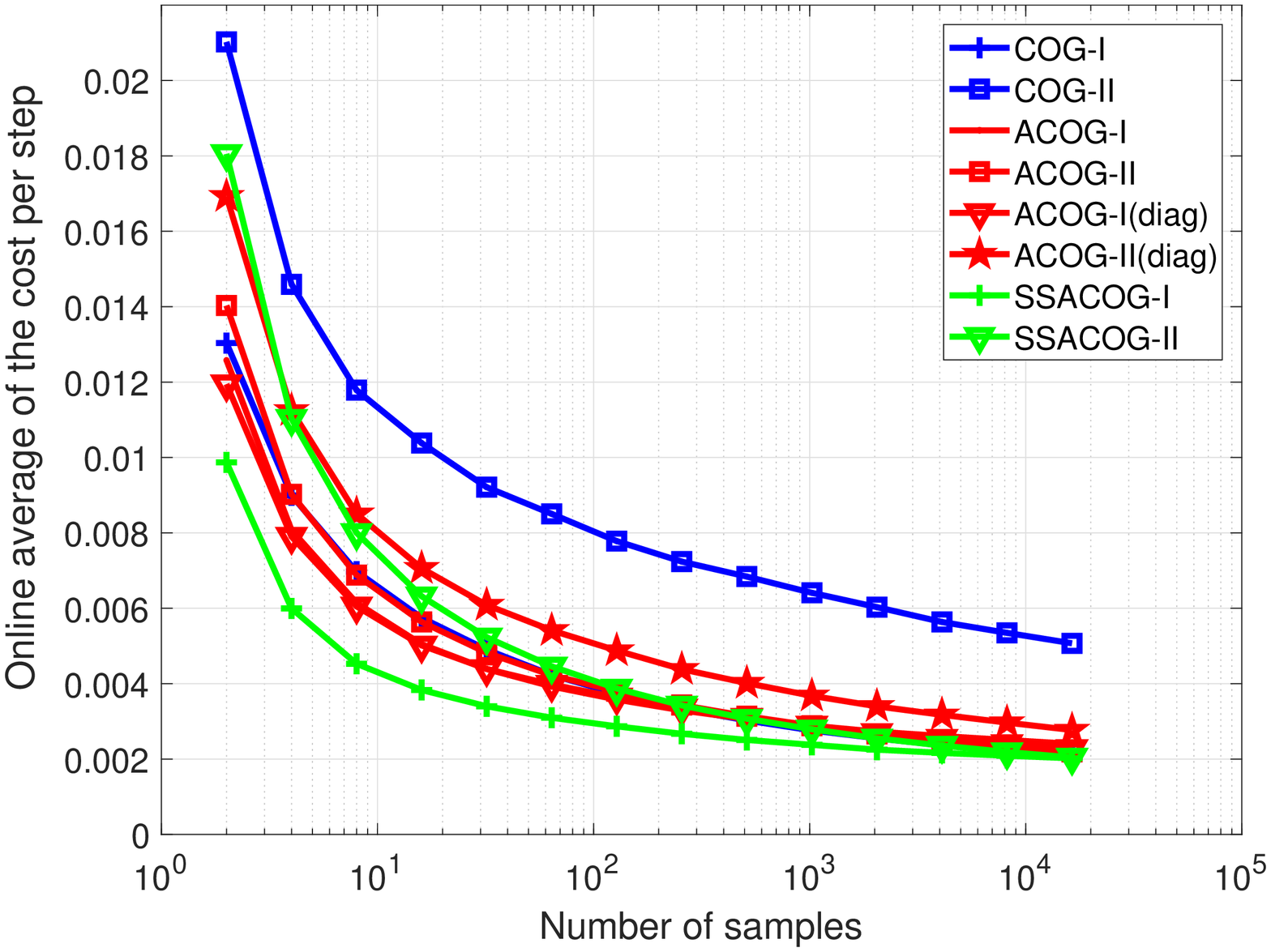}}
      \centerline{(a) mushrooms}
    \end{minipage}
    \hfill
    \begin{minipage}{0.5\linewidth}
      \centerline{\includegraphics[width=4.5cm]{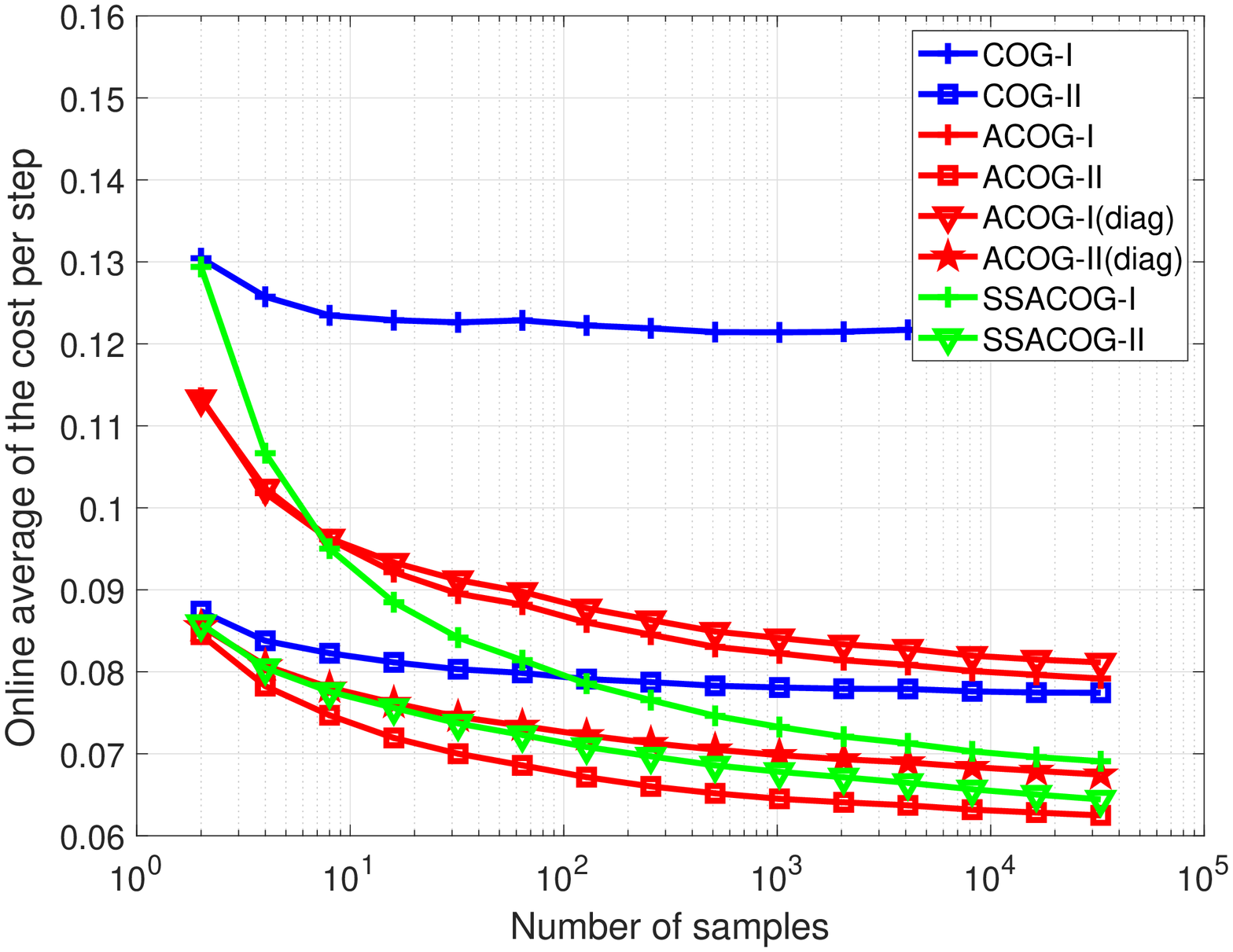}}
      \centerline{(b) protein}
    \end{minipage}
    \vfill
    \begin{minipage}{0.5\linewidth}
      \centerline{\includegraphics[width=4.5cm]{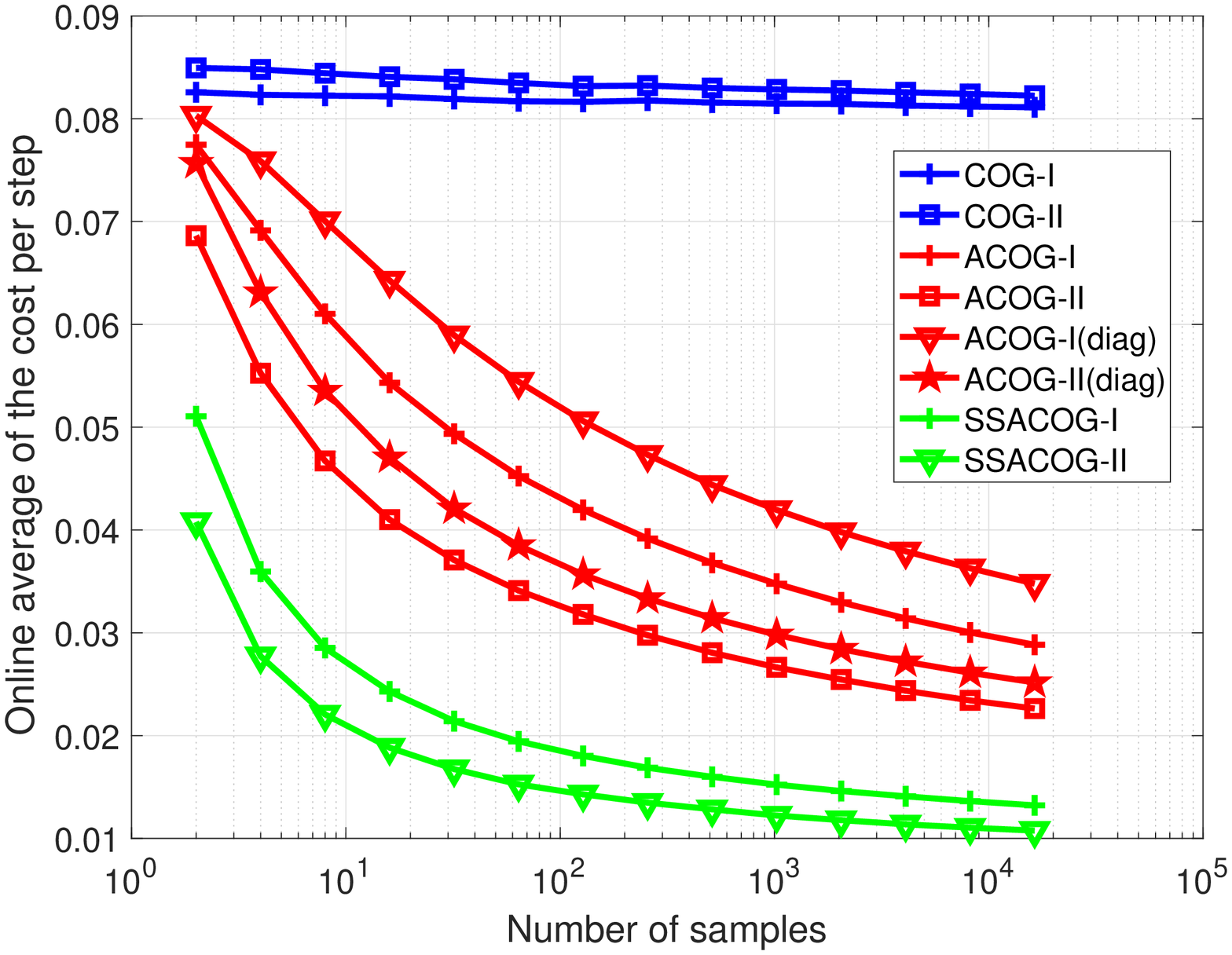}}
      \centerline{(c) Sensorless}
    \end{minipage}
    \hfill
    \begin{minipage}{0.5\linewidth}
      \centerline{\includegraphics[width=4.5cm]{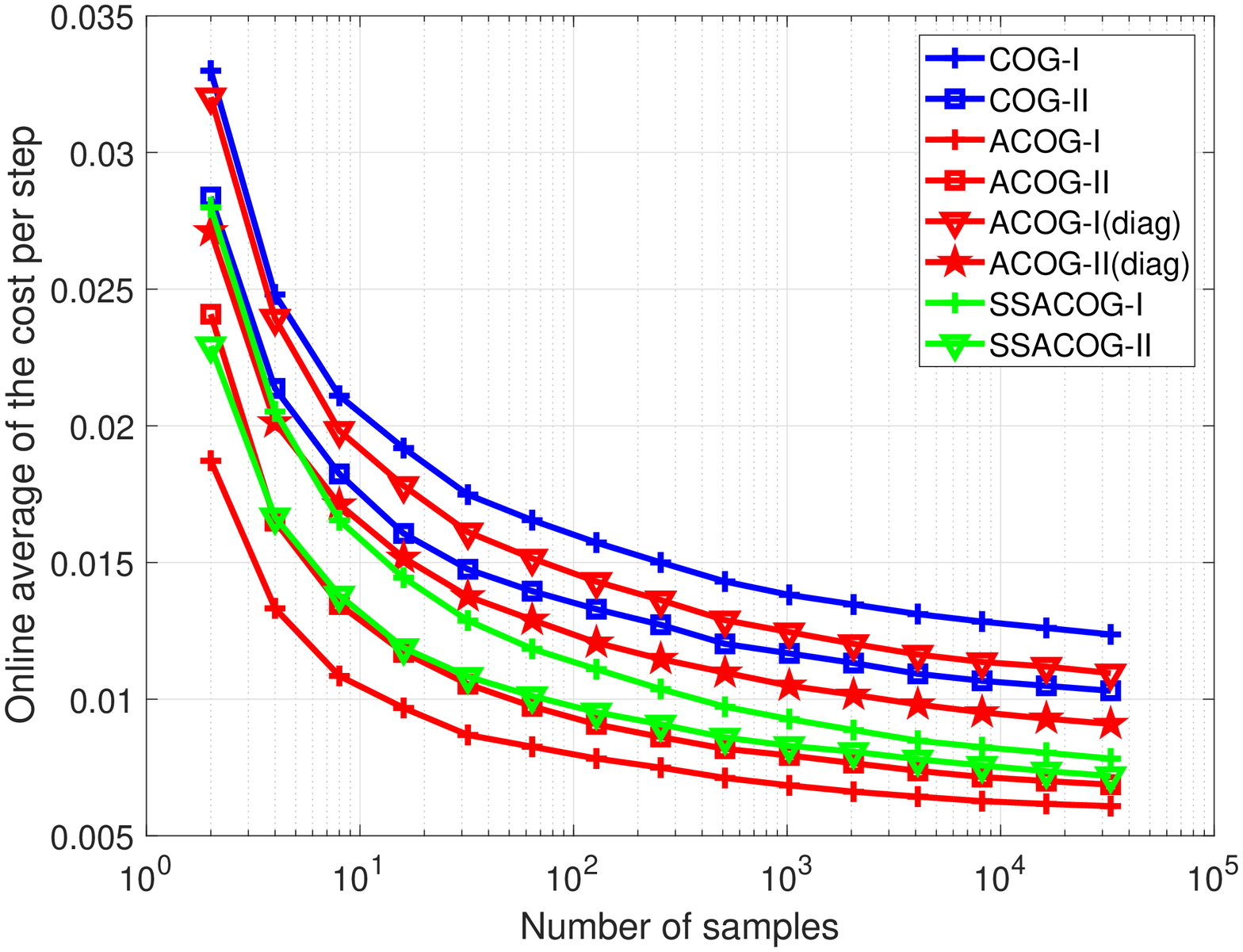}}
      \centerline{(d) usps}
    \end{minipage}
   \vspace{-0.08in}
    \caption{Weighted ``cost`` performance of SACOG.}

\end{figure}

\begin{table*}
	\caption{Evaluation of the Cost-Sensitive Classification Performance of SSACOG}
    \vspace{-0.15in}
    \begin{center}
    \scalebox{0.87}{
	\begin{tabular}{|l|c|c|c|c|c|c|c|c|}\hline
	
        \multirow{2}{*}{Algorithm}&\multicolumn{4}{c|}{$``sum``$ on mushrooms} &  \multicolumn{4}{c|}{$``cost``$ on mushrooms}\cr\cline{2-9}
        &Sum(\%)&Sensitivity(\%)&Specificity  (\%)&Time(s)   & Cost&Sensitivity(\%)&Specificity  (\%)&Time(s)\cr
        \hline \hline

        COG-I  &99.205 	$\pm$ 0.047 	& 99.455 	$\pm$ 0.075 	& 98.956 	$\pm$ 0.095 	 	& 0.019 &\textbf{15.760 	$\pm$ 2.496} 	& 99.823 	$\pm$ 0.070 	& 97.688 	$\pm$ 0.091 	 	& 0.020 	 \\	
        COG-II  &99.211 	$\pm$ 0.057 	& 99.420 	$\pm$ 0.094 	& 99.003 	$\pm$ 0.097 		& 0.019 	&39.180 	$\pm$ 2.283 	& 99.538 	$\pm$ 0.055 	& 94.465 	$\pm$ 0.275 	 	& 0.019 	 \\
        ACOG-I &99.580 	$\pm$ 0.027 	& 99.810 	$\pm$ 0.070 	& 99.350 	$\pm$ 0.076 		& \textbf{0.043} &18.735 	$\pm$ 1.062 	& 99.939 	$\pm$ 0.051 	& 95.802 	$\pm$ 0.443 		& \textbf{0.085}	 \\	
        ACOG-II  &99.572 	$\pm$ 0.033 	& 99.794 	$\pm$ 0.080 	& 99.349 	$\pm$ 0.075 		& \textbf{0.045}	&16.770 	$\pm$ 1.546 	& 99.932 	$\pm$ 0.035 	& 96.373 	$\pm$ 0.412 		& \textbf{0.054} 	 \\
        ACOG-I$_{diag}$  &99.447 	$\pm$ 0.052 	& 99.652 	$\pm$ 0.077 	& 99.243 	$\pm$ 0.087 		& 0.019 	  &17.520 	$\pm$ 1.588 	& 99.933 	$\pm$ 0.045 		& 98.119 	$\pm$ 0.060 	& 0.020 	 \\
        ACOG-II$_{diag}$   &99.457 	$\pm$ 0.052 	& 99.652 	$\pm$ 0.086 	& 99.262 	$\pm$ 0.117 		& 0.019 	  &21.185 	$\pm$ 1.431 	& 99.792 	$\pm$ 0.037 	& 96.601 	$\pm$ 0.167 		& 0.019 	 \\
        SSACOG-I  &\textbf{99.628 	$\pm$ 0.052} 	& 99.798 	$\pm$ 0.066 	& 99.459 	$\pm$ 0.102 		& \textbf{0.038}  &\textbf{15.880 	$\pm$ 1.677} 	& 99.930 	$\pm$ 0.043 	& 96.623 	$\pm$ 0.303 	& \textbf{0.041} 	 \\	
        SSACOG-II  &\textbf{99.606 	$\pm$ 0.050} 	& 99.805 	$\pm$ 0.062 	& 99.408 	$\pm$ 0.093 		&\textbf{0.038}	   &\textbf{15.560 	$\pm$ 3.870} 	& 99.869 	$\pm$ 0.040 	& 97.291 	$\pm$ 1.063 		& \textbf{0.034} 	 \\

        \hline
        \hline
        \multirow{2}{*}{Algorithm}&\multicolumn{4}{c|}{$``sum``$ on protein} &  \multicolumn{4}{c|}{$``cost``$ on protein}\cr\cline{2-9}
        &Sum(\%)&Sensitivity(\%)&Specificity  (\%)&Time(s)   & Cost&Sensitivity(\%)&Specificity  (\%)&Time(s)\cr
        \hline \hline

        COG-I  &69.935 	$\pm$ 0.213 	& 68.114 	$\pm$ 0.343 	& 71.757 	$\pm$ 0.322 		& 0.127 	  &2156.980 	$\pm$ 35.558 	& 75.944 	$\pm$ 0.463 	& 60.071 	$\pm$ 0.270 		& 0.151 	 \\
        COG-II  &69.764 	$\pm$ 0.230 	& 70.005 	$\pm$ 0.392 	& 69.523 	$\pm$ 0.415 		& 0.129 	  &1375.740 	$\pm$ 12.402 	& 90.618 	$\pm$ 0.106 	& 28.559 	$\pm$ 0.607 		& 0.152 	 \\
        ACOG-I &71.340 	$\pm$ 0.214 	& 69.794 	$\pm$ 0.427 	& 72.886 	$\pm$ 0.385 		& \textbf{14.603} 	  &1406.660 	$\pm$ 28.314 	& 87.072 	$\pm$ 0.428 	& 52.671 	$\pm$ 0.576 		& \textbf{16.922} 	 \\
        ACOG-II  &71.265 	$\pm$ 0.235 	& 71.678 	$\pm$ 0.398 	& 70.852 	$\pm$ 0.501 		& \textbf{14.446} 	  &\textbf{1110.075 	$\pm$ 12.274} 	& 94.972 	$\pm$ 0.172 	& 22.753 	$\pm$ 0.853 		& \textbf{13.601} 	 \\
        ACOG-I$_{diag}$  &71.305 	$\pm$ 0.126 	& 69.825 	$\pm$ 0.346 	& 72.785 	$\pm$ 0.257 		& 0.161	  &1441.505 	$\pm$ 19.496 	& 86.500 	$\pm$ 0.269 	& 53.441 	$\pm$ 0.394 		& 0.171 	 \\
        ACOG-II$_{diag}$ &71.233 	$\pm$ 0.150 	& 71.530 	$\pm$ 0.365 	& 70.935 	$\pm$ 0.298 		& 0.158 	  &1198.455 	$\pm$ 11.459 	& 92.585 	$\pm$ 0.143 	& 31.925 	$\pm$ 0.685 		& 0.166 	 \\
        SSACOG-I  &\textbf{71.532 	$\pm$ 0.198} 	& 66.861 	$\pm$ 0.530 	& 76.203 	$\pm$ 0.485 		& \textbf{0.355} 	  &1227.345 	$\pm$ 16.904 	& 90.608 	$\pm$ 0.225 	& 44.148 	$\pm$ 0.342 	       & \textbf{0.393} 	 \\
        SSACOG-II  &\textbf{71.323 	$\pm$ 0.132} 	& 71.725 	$\pm$ 0.305 	& 72.075 	$\pm$ 0.403 		& \textbf{0.352} 	  &1144.680 	$\pm$ 13.087 	& 94.053 	$\pm$ 0.144 	& 26.224 	$\pm$ 0.661 		& \textbf{0.348} 	 \\
        \hline
        \hline
        \multirow{2}{*}{Algorithm}&\multicolumn{4}{c|}{$``sum``$ on Sensorless} &  \multicolumn{4}{c|}{$``cost``$ on Sensorless}\cr\cline{2-9}
        &Sum(\%)&Sensitivity(\%)&Specificity  (\%)&Time(s)   & Cost&Sensitivity(\%)&Specificity  (\%)&Time(s)\cr
        \hline \hline

        COG-I  &50.888 	$\pm$ 0.227 	& 9.637 	$\pm$ 0.473 	& 92.139 	$\pm$ 0.076 		& 0.166 &4741.190 	$\pm$ 21.159 	& 11.155 	$\pm$ 0.403 	& 90.823 	$\pm$ 0.038 		& 0.154 	 \\	
        COG-II  &52.374 	$\pm$ 0.422 	& 52.717 	$\pm$ 0.464 	& 52.032 	$\pm$ 0.387 		& 0.167 	  &4801.600 	$\pm$ 38.579 	& 50.168 	$\pm$ 0.415 	& 54.576 	$\pm$ 0.354 		& 0.149 	 \\
        ACOG-I &83.468 	$\pm$ 0.308 	& 72.935 	$\pm$ 0.620 	& 94.001 	$\pm$ 0.068 		& \textbf{0.503} &1622.600 	$\pm$ 32.312 	& 72.563 	$\pm$ 0.709 	& 94.188 	$\pm$ 0.078 		& \textbf{0.480} 	 \\
        ACOG-II  &87.398 	$\pm$ 0.186 	& 88.088 	$\pm$ 0.284 	& 86.708 	$\pm$ 0.178 		& \textbf{0.486} 	  &1283.350 	$\pm$ 16.768 	& 87.247 	$\pm$ 0.264 	& 87.350 	$\pm$ 0.131 		& \textbf{0.455} 	 \\
        ACOG-I$_{diag}$  &80.044 	$\pm$ 0.314  	& 66.427 	$\pm$ 0.627 	& 93.661 	$\pm$ 0.051 		& 0.169 	  &1956.200 	$\pm$ 33.729 	& 65.995 	$\pm$ 0.668 	& 93.827 	$\pm$ 0.071 		& 0.157 	 \\
        ACOG-II$_{diag}$  &85.968 	$\pm$ 0.124 	& 86.608 	$\pm$ 0.178 	& 85.328 	$\pm$ 0.118 		& 0.173 	   &1422.950 	$\pm$ 16.836 	& 85.783 	$\pm$ 0.227 	& 86.043 	$\pm$ 0.131 		& 0.153 	 \\
        SSACOG-I  &\textbf{92.432 	$\pm$ 0.213} 	& 89.818 	$\pm$ 0.442 	& 95.047 	$\pm$ 0.047 		& \textbf{0.322}  &\textbf{753.695 	$\pm$ 21.185} 	& 89.482 	$\pm$ 0.476 	& 95.296 	$\pm$ 0.066 		& \textbf{0.285} 	 \\
        SSACOG-II  &\textbf{93.913 	$\pm$ 0.129} 	& 94.487 	$\pm$ 0.181 	& 93.339 	$\pm$ 0.123 		& \textbf{0.296} 	  &\textbf{615.625 	$\pm$ 12.280} 	& 94.166 	$\pm$ 0.194 	& 93.676 	$\pm$ 0.096 		& \textbf{0.264} 	 \\

        \hline
        \hline
        \multirow{2}{*}{Algorithm}&\multicolumn{4}{c|}{$``sum``$ on usps} &  \multicolumn{4}{c|}{$``cost``$ on usps}\cr\cline{2-9}
        &Sum(\%)&Sensitivity(\%)&Specificity  (\%)&Time(s)   & Cost&Sensitivity(\%)&Specificity  (\%)&Time(s)\cr
        \hline \hline
        COG-I  &96.820 	$\pm$ 0.165 	& 96.361 	$\pm$ 0.345 	& 97.279 	$\pm$ 0.116 		& 0.039 	   &90.165 	$\pm$ 3.851 	& 92.642 	$\pm$ 0.344 	& 98.179 	$\pm$ 0.060 		& 0.031 	 \\
        COG-II  &96.576 	$\pm$ 0.139 	& 96.516 	$\pm$ 0.226 	& 96.637 	$\pm$ 0.193 		& 0.038 	  &75.135 	$\pm$ 4.338 	& 96.570 	$\pm$ 0.215 	& 93.722 	$\pm$ 0.342 		& 0.030 	 \\
        ACOG-I &\textbf{98.073 	$\pm$ 0.115} 	& 97.822 	$\pm$ 0.242 	& 98.323 	$\pm$ 0.095 		& \textbf{0.271} 	 &\textbf{44.365 	$\pm$ 3.448} 	& 96.671 	$\pm$ 0.321 	& 98.591 	$\pm$ 0.070 		& \textbf{0.151} 	 \\
        ACOG-II  &\textbf{97.633 	$\pm$ 0.148} 	& 97.998 	$\pm$ 0.230 	& 97.268 	$\pm$ 0.176 		& \textbf{0.239} 	  &\textbf{50.100 	$\pm$ 4.321} 	& 98.241 	$\pm$ 0.172 	& 94.883 	$\pm$ 0.488 		& \textbf{0.252} 	 \\
        ACOG-I$_{diag}$  &96.886 	$\pm$ 0.226 	& 95.641 	$\pm$ 0.435 	& 98.131 	$\pm$ 0.076 		& 0.039 	   &79.850 	$\pm$ 4.773 	& 93.526 	$\pm$ 0.423 	& 98.314 	$\pm$ 0.080 		& 0.031 	 \\
        ACOG-II$_{diag}$ &96.305 	$\pm$ 0.182 	& 96.369 	$\pm$ 0.228 	& 96.240 	$\pm$ 0.182 		& 0.040 	 &66.300 	$\pm$ 3.242 	& 96.993 	$\pm$ 0.149 	& 94.425 	$\pm$ 0.295 		& 0.030 	 \\
        SSACOG-I  &97.091 	$\pm$ 0.197	& 96.817 	$\pm$ 0.323 	& 97.365 	$\pm$ 0.125 		& \textbf{0.077} 	   &57.055 	$\pm$ 4.251 	& 95.657 	$\pm$ 0.384 	& 98.296 	$\pm$ 0.076 	 	& \textbf{0.054} 	 \\
        SSACOG-II &97.048 	$\pm$ 0.163  	& 97.010 	$\pm$ 0.237 	& 97.085 	$\pm$ 0.190 		& \textbf{0.074}  &52.420 	$\pm$ 4.009 	& 97.647 	$\pm$ 0.192 	& 95.550 	$\pm$ 0.360 	 	& \textbf{0.054} 	 \\

        \hline

	\end{tabular}}
    \end{center}
    \vspace{-0.1in}
\end{table*}

\begin{table*}
	\caption{Evaluation between SACOG and Sparse SACOG}
    \vspace{-0.15in}
    \begin{center}
    \scalebox{0.9}{
	\begin{tabular}{|l|c|c|c|c|c|c|c|c|}\hline
        \multirow{2}{*}{Algorithm}&\multicolumn{2}{c|}{$``sum``$ on mushrooms} &  \multicolumn{2}{c|}{$``cost``$ on mushrooms} & \multicolumn{2}{c|}{$``sum``$ on protein} &  \multicolumn{2}{c|}{$``cost``$ on protein}\cr\cline{2-9}
        &Sum(\%)&Time(s)   & Cost($10^2$) &Time(s) &Sum(\%)&Time(s)   & Cost($10^2$)&Time(s)\cr
        \hline \hline

		SACOG-I  &99.620 	$\pm$ 0.043 		& 0.072 	&16.020 	$\pm$ 1.796 	&  0.096 	 &71.544 	$\pm$ 0.197	& 3.769 	 &1226.890 	$\pm$ 17.094 	& 3.302 	  \\
        SACOG-II  &99.598 	$\pm$ 0.040 	 	& 0.074 	 &13.790 	$\pm$ 1.852 	&  0.035 	   &71.907 	$\pm$ 0.180      & 3.705 	   &1147.775 	$\pm$ 14.364 	& 2.373 		 \\
        SSACOG-I  &99.628 	$\pm$ 0.052 	& 0.038 	  &15.880 	$\pm$ 1.677 	&  0.039 	 &71.532 	$\pm$ 0.198  	& 0.287 	&1227.345 	$\pm$ 16.904 	& 0.272 		 \\
        SSACOG-II  &99.606 	$\pm$ 0.050 	& 0.038 	&15.560 	$\pm$ 3.870 	& 0.033 	&71.900 	$\pm$ 0.204 	& 0.285 	 &1144.680 	$\pm$ 13.087 	& 0.239 	 \\
        \hline \hline

        \multirow{2}{*}{Algorithm}&\multicolumn{2}{c|}{$``sum``$ on Sensorless} &  \multicolumn{2}{c|}{$``cost``$ on Sensorless}&\multicolumn{2}{c|}{$``sum``$ on usps} &  \multicolumn{2}{c|}{$``cost``$ on usps}\cr\cline{2-9}
        &Sum(\%)&Time(s)   & Cost($10^2$)&Time(s)&Sum(\%)&Time(s)   & Cost($10^2$)&Time(s)\cr
        \hline \hline

        SACOG-I  &92.432 	$\pm$ 0.213  	& 0.232 	 &753.695 	$\pm$ 21.185 	& 0.235 	 &97.146 	$\pm$ 0.149  	& 0.135 	 &55.970 	$\pm$ 3.053 	& 0.078  \\
        SACOG-II  &93.913 	$\pm$ 0.129 	& 0.193 	  &615.625 	$\pm$ 12.280 	& 0.194 	&97.071 	$\pm$ 0.169  	& 0.091 	  &53.155 	$\pm$ 4.815 	& 0.090  \\
        SSACOG-I    &92.432 	$\pm$ 0.213  	& 0.239 	  &753.695 	$\pm$ 21.185 	& 0.225 	&97.091 	$\pm$ 0.197  	& 0.057 	 &57.055 	$\pm$ 4.251 	& 0.052  \\
        SSACOG-II  &93.913 	$\pm$ 0.129 	& 0.214 	   &615.625 	$\pm$ 12.280 	& 0.204 	 &97.048 	$\pm$ 0.163  	& 0.054 	  &52.420 	$\pm$ 4.009 	& 0.053 	 \\
        \hline
	\end{tabular}
        }
    \end{center}
    \vspace{-0.15in}
\end{table*}

\vspace{-0.05in}
\subsubsection{Efficiency Comparison between Sketched ACOG and Sparse Sketched ACOG }

Then, We would like to compare the performance and running time between SACOG and its sparse version SSACOG. The experimental results based on both metrics are summarized in Table 6.

From results, we find that the running time of SSACOG is lower than SACOG. It is consistent with the time complexity analysis of two algorithms in Section 3. For better understanding, we simply give a analysis. Given sketch size $m=5$, the time complexity for SACOG is $O(25d)$ according to the analysis of Section 3, while the time complexity for SSACOG is $O(125+5s)$. One can accelerate the time complexity to $O(5d)$ for SACOG and $O(25+5s)$ for SSACOG by only updating the sketch every $m$ round.

Thus, the time complexity for SACOG is linear in the data dimensionality $d$, and running time for SSACOG is linear in the data non-sparse degree $s$. Then, it is easy to understand the SSACOG would be much faster than SACOG, when the data dimensionality $d$ is high and the data sparsity is strong $s\ll d$.

\section{Application to Online Anomaly Detection}

The proposed adaptive regularized cost-sensitive online classification algorithms can be potentially applied to solve a wide range of real-world applications in data mining. To verify their practical application value, we apply them to tackle several online anomaly detection tasks in this section.

\subsection{Application Domains and Testbeds}

Below, we first exhibit the related domains of anomaly detection problems:

$\bullet$ Finance: The credit card approval problem enjoys a huge demand in financial domains, where our task is to discriminate the credit-worthy customers for the Australian dataset from an Australian credit company.

$\bullet$ Nuclear: We apply our algorithms to the Magic04 dataset with 19020 samples to simulate registration of high gamma particles. The dataset was collected by a ground-based atmospheric Cherenkov gamma telescope. In detail, the ``gamma signal`` samples are considered as the normal class, while the hadron ones are treated as outliers.

$\bullet$ Bioinformatics: We address bioinformatics anomaly detection problems with DNA dataset to recognize the boundaries between exons and introns from a given DNA sequence, where exon/intron boundaries are defined as anomalies and others are treated as normal.

$\bullet$ Medical Imaging: We apply our approaches to address the medical image anomaly detection problem with the KDDCUP08 breast cancer dataset\footnote{http://www.sigkdd.org/kddcup/}. The main goal is to detect the breast cancer from X-ray images, where ``benign`` is assigned as normal and ``malignant`` is abnormal.


To better understand, we summary the detailed information for each dataset in Table 7.
\begin{table}[h]\label{8}
	\caption{\small{Datasets for Online Anomaly Detection}}
    \vspace{-0.13in}
    \begin{center}
    \scalebox{0.95}{
	\begin{tabular}{|l|c|c|c|}\hline
		Dataset &\#Examples & \#Features & \#Pos:\#Neg  \\\hline \hline
        Australian & 690 & 14 & 1:1.25  \\
        Magic04 & 19020  & 10 & 1:1.8  \\
        DNA&2000&180&1:3.31\\
        KDDCUP08&102294&117&1:163.19\\ \hline
	\end{tabular}
    }
    \end{center}
    \vspace{-0.3in}
\end{table}

\subsection{Empirical Evaluation Results}

In this subsection, our algorithms are applied to address real-world anomaly detection tasks with 4 datasets from different domains, where we use the \emph{balanced accuracy} metric to avoid inflated performance evaluations on imbalanced datasets. In addition, we apply our sparse sketched ACOG algorithms (SSACOG) only for two high-dimensional datasets (i.e., DNA and KDDCUP08), because for low-dimensional tasks, the proposed ACOG algorithms are fast enough. Furthermore, all implementation settings are same as Section 4.

Table 8 exhibits the experimental results, from which we can draw several observations. First of all, two cost-sensitive methods (PAUM and CPA$_{PB}$) outperform their regular methods (Perceptron and PA-I) among all datasets. This confirms the superiority of cost-sensitiveness for online learning. Second, COG algorithms outperform all regular first-order algorithms (i.e., first 5 baselines) on almost all datasets, which demonstrates the effectiveness of direct cost-sensitive optimization in online learning.

Moreover, ACOG algorithms and AROW algorithm outperform all other algorithms, where ACOG is the updated version of COG with adaptive regularization using second order information. This infers the online classification that introduces the second-order inner-correlation information can enjoy a huge performance improvement. Furthermore, the performance of ACOG exceeds all other algorithms, which demonstrates the effectiveness of cost-sensitive online optimization using the second order information.

By the way, although the speed of SSACOG is slightly slower than ACOG$_{diag}$, its performance is relatively better. On the other hand, SSACOG is much faster than ACOG with slight performance loss. This implies that the sketching version of ACOG is a good choice to balance the performance and efficiency for handling high-dimensional real-world tasks. Furthermore, if someone only wants to pursue the efficiency, they can regard ACOG$_{diag}$ as a choice.

In conclusion, all promising results confirm the superiority of our proposed algorithms for real-world online anomaly detection problems, where datasets are normally high-dimensional and highly class-imbalanced.

\vspace{-0.1in}
\section{Conclusion}

In this paper, to remedy the weakness of first-order cost-sensitive online learning algorithms, we propose to introduce second-order information into cost-sensitive online classification framework based on adaptive regularization. As a result, a family of second-order cost-sensitive online classification algorithms is proposed, with favourable regret bound and impressive properties.

Moreover, to overcome the time-consuming problem of our second-order algorithms, we further study the sketching method in cost-sensitive online classification framework, and then propose sketched cost-sensitive online classification algorithms, which can be developed as a sparse cost-sensitive online learning approach, with better trade off between the performance and efficiency.

Then for examination of the performance and efficiency, we empirically evaluate our proposed algorithms on many public real-world datasets in extensive experiments. Promising results not only prove the new proposed algorithms successfully overcome the limitation of first-order algorithms, but also confirm their effectiveness and efficiency in solving real-world cost-sensitive online classification problems.

Future works include: (i) further exploration about the in-depth theory of cost-sensitive online learning; (ii) further study about the sparse computation methods in cost-sensitive online classification problems.

\begin{table}[t]
	\caption{Evaluation  for online anomaly detection}
    \vspace{-0.1in}
    \begin{center}
    \scalebox{0.9}{
	\begin{tabular}{|l|c|c|c|c|}\hline
        \multirow{2}{*}{Algorithm}&\multicolumn{2}{c|}{$``sum``$ on Australian} &\multicolumn{2}{c|}{$``sum``$ on Magic04}  \cr \cline{2-5}
        &Sum(\%)&Time(s) &Sum(\%)&Time(s)\cr
        \hline \hline
        Perceptron   &57.863 	$\pm$ 1.327& 0.002 &59.154 	$\pm$ 0.408& 0.030   	 \\
        ROMMA        &58.732 	$\pm$ 3.462& 0.002  &64.025 	$\pm$ 3.277& 0.042  	 \\
        PA-I         &57.103 	$\pm$ 1.595 & 0.002 &58.029 	$\pm$ 0.312 & 0.036 	\\
        PAUM         &62.362 	$\pm$ 0.941 & 0.002  &64.671  	$\pm$ 0.204 & 0.030   	\\
        CPA$_{PB}$  &57.110 	$\pm$ 1.599 & 0.003  &58.448 	$\pm$ 0.360 & 0.043    \\
        AROW  &67.174 	$\pm$ 0.749 & 0.008  &70.896 	$\pm$ 0.190 & 0.154    \\
        COG-I  &65.972 	$\pm$ 0.879 & 0.002 &65.913 	$\pm$ 0.189& 0.030   \\
        COG-II  &67.213 	$\pm$ 0.787  & 0.002  &69.815 	$\pm$ 0.183 & 0.030    \\
        ACOG-I  &68.808 	$\pm$ 0.894  & 0.005   &72.935 	$\pm$ 0.186 & 0.088   	\\
        ACOG-II  &\textbf{69.228 	$\pm$ 0.733} & 0.005  &68.345 	$\pm$ 1.822 & 0.092   \\
        ACOG-I$_{diag}$  &68.464	$\pm$ 0.936 & 0.002    &\textbf{73.268 	$\pm$ 0.158}& 0.033    \\
        ACOG-II$_{diag}$  &68.510 	$\pm$ 0.917 & 0.002 &73.035 	$\pm$ 0.187& 0.033 \\

        \hline \hline

        \multirow{2}{*}{Algorithm} &\multicolumn{2}{c|}{$``sum``$ on DNA}&\multicolumn{2}{c|}{$``sum``$ on KDDCUP08} \cr \cline{2-5}
        &Sum(\%)&Time(s) &Sum(\%)&Time(s) \cr
        \hline \hline
        Perceptron    &84.759 	$\pm$ 0.575 	 	& 0.006  &54.018 	$\pm$ 1.056 	& 0.376	 \\
        ROMMA        &85.782 	$\pm$ 0.553 		& 0.006  &54.342 	$\pm$ 1.581 	& 0.507  	 \\
        PA-I         &87.832 	$\pm$ 0.833 		& 0.005 &54.053 	$\pm$ 0.865 	& 0.414  	\\
        PAUM          &88.560 	$\pm$ 0.737 		& 0.005 &55.161 	$\pm$ 0.424 	& 0.386   	\\
        CPA$_{PB}$   &89.401 	$\pm$ 0.645 	& 0.007 	&57.318 	$\pm$ 0.629 	& 0.458 \\
        AROW    &89.183 	$\pm$ 0.405 		& 0.269 	 &50.611 	$\pm$ 0.422 	& 12.554 \\
        COG-I   &87.886 	$\pm$ 0.812 		& 0.006 	 &54.094 	$\pm$ 1.047 	& 0.355\\
        COG-II    &87.395 	$\pm$ 0.530 	& 0.005 	&69.312 	$\pm$ 0.475 	& 0.359	  \\
        ACOG-I    &\textbf{91.490 	$\pm$ 0.416} 	& 0.104 	&55.088 	$\pm$ 0.936 & 4.531 	\\
        ACOG-II    &90.872 	$\pm$ 0.677 	& 0.234 	&\textbf{71.920 	$\pm$ 2.016} & 5.803 \\
        ACOG-I$_{diag}$  &89.498 	$\pm$ 0.633 	& 0.006 	 &55.293 	$\pm$ 0.852 	& 0.384	  \\
        ACOG-II$_{diag}$  &88.433 	$\pm$ 0.490 	& 0.006 	&71.661 	$\pm$ 1.334 	& 0.397\\
        SSACOG-I  &89.975 	$\pm$ 0.516 	& 0.016  &55.711 	$\pm$ 0.812 	& 0.810 		 \\
        SSACOG-II &90.444 	$\pm$ 0.471     & 0.023 	&70.947 	$\pm$ 1.179 	& 0.842 		 \\
        \hline
	\end{tabular}
        }
    \end{center}
    \vspace{-0.2in}
\end{table}

\section{Acknowledgement}
\blue{This research is partly supported by the National Research Foundation, Prime Minister¡¯s Office, Singapore under its International Research Centres in Singapore Funding Initiative, and partly supported by National Natural Science Foundation of China (NSFC) under Grant 61602185, Fundamental Research Funds for the Central Universities under Grant D2172480.}

%
\vspace{-10ex}
\begin{IEEEbiography}[{\includegraphics[width=1in,height=1.25in,clip,keepaspectratio]{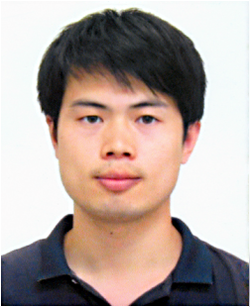}}]{Peilin Zhao}
is currently a researcher with SCUT, China. His research interests are Large-Scale Machine Learning and its applications to Big Data Analytics. Previously, he has worked at Ant Financial, Institute for Infocomm Research (I2R), Rutgers University. He received his PHD degree from Nanyang Technological University and his bachelor degree from Zhejiang University. In his research areas, he has published over 70 papers in top venues, including JMLR, AIJ, ICML, NIPS, KDD, etc. He has been invited as a PC member or reviewer for many international conferences and journals in his area.
\end{IEEEbiography}
\vspace{-10ex}
\begin{IEEEbiography}[{\includegraphics[width=1in,height=1.25in,clip,keepaspectratio]{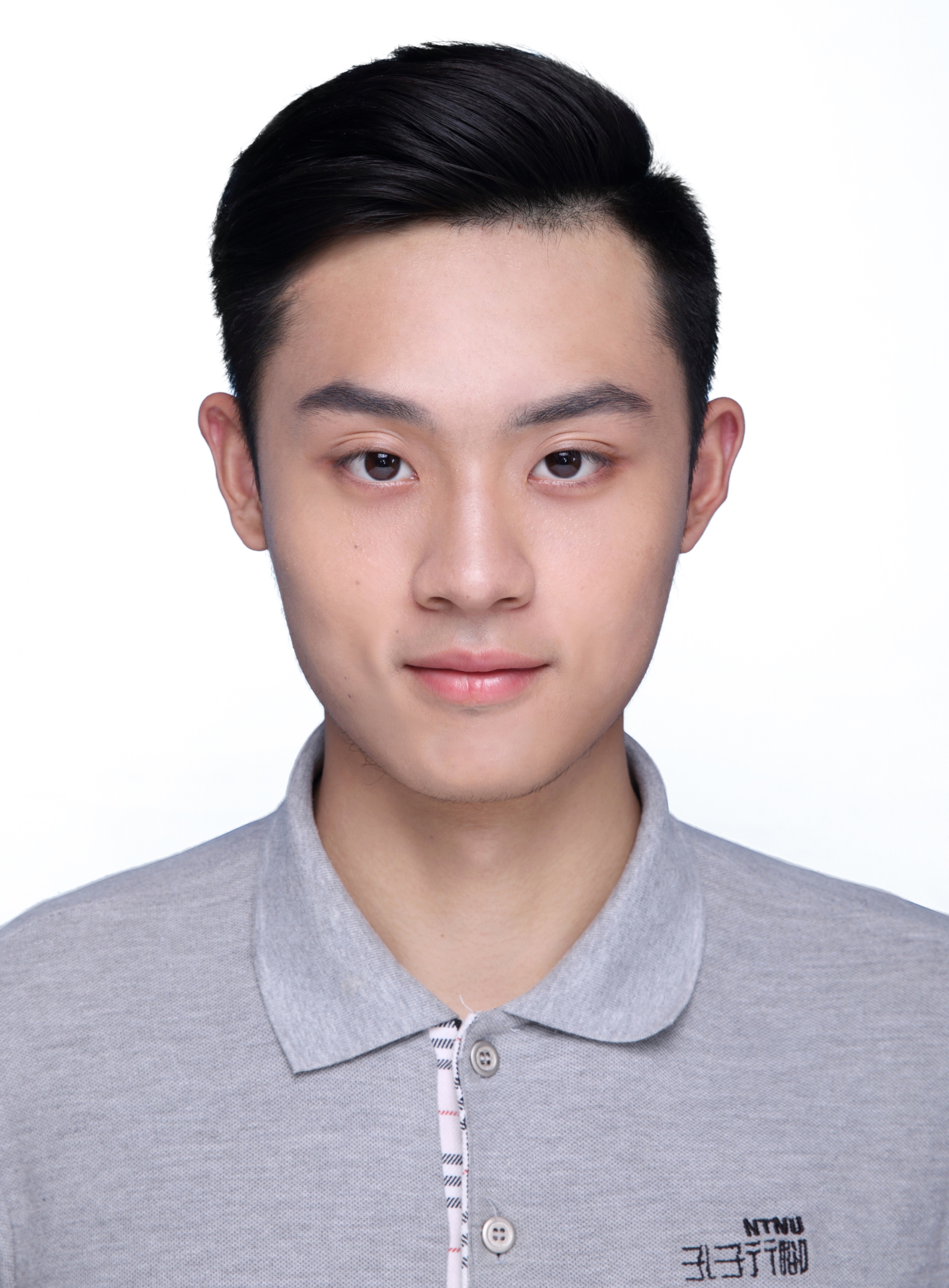}}]{Yifan Zhang}
received the BE degree in electronic commerce from the Southwest University, China, in 2017. He is working toward the ME degree in the School of Software Engineering, South China University of Technology, China. His research interests include Machine Learning, Reinforcement Learning and their applications in Big Data Analytics.
\end{IEEEbiography}
\vspace{-10ex}
\begin{IEEEbiography}[{\includegraphics[width=1in,height=1.25in,clip,keepaspectratio]{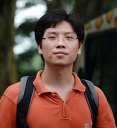}}]{Min Wu}
is a research scientist in Data Analytics Department, Institute for Infocomm Research. He received Ph.D degree from Nanyang Technological University, Singapore, in 2011, and received B.S. degree in Computer Science from University of Science and Technology of China, 2006. His research interests include Graph Mining from Large-Scale Networks, Learning from Heterogeneous Data Sources, Ensemble Learning, and Bioinformatics.
\end{IEEEbiography}
\vspace{-10ex}
\begin{IEEEbiography}[{\includegraphics[width=1in,height=1.25in,clip,keepaspectratio]{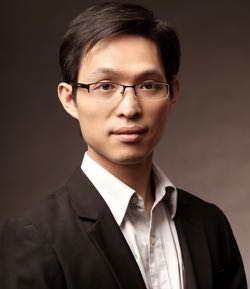}}]{Steven C. H. Hoi}
is an Associate Professor in Singapore Management University (SMU), Singapore. Prior to joining SMU, he was a tenured Associate Professor at Nanyang Technological University (NTU), Singapore. He received his Bachelor degree from Tsinghua University, and his Master and Ph.D degrees from the Chinese University of Hong Kong. His research interests include large-scale machine learning with application to a wide range of real-world applications. He has published over 150 papers in premier conferences and journals, and served as an organizer, area chair, senior PC, TPC member, editors, and referee for many top conferences and premier journals. He is the recipient of the Lee Kong Chian Fellowship Award due to his research excellence.
\end{IEEEbiography}
\vspace{-10ex}
\begin{IEEEbiography}[{\includegraphics[width=1in,height=1.25in,clip,keepaspectratio]{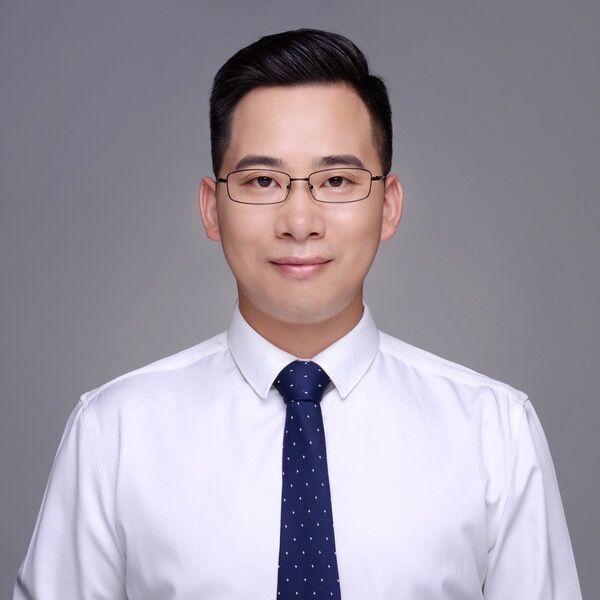}}]{Mingkui Tan}
received the PhD degree in computer science from Nanyang Technological University, Singapore, in 2014. He is a professor in the School of Software Engineering, South China University of Technology. After that, he worked as a senior research associate in the School of Computer Science, University of Adelaide, Australia. His research interests include compressive sensing, big data learning, and large-scale optimization.
\end{IEEEbiography}
\vspace{-10ex}
\begin{IEEEbiography}[{\includegraphics[width=1in,height=1.25in,clip,keepaspectratio]{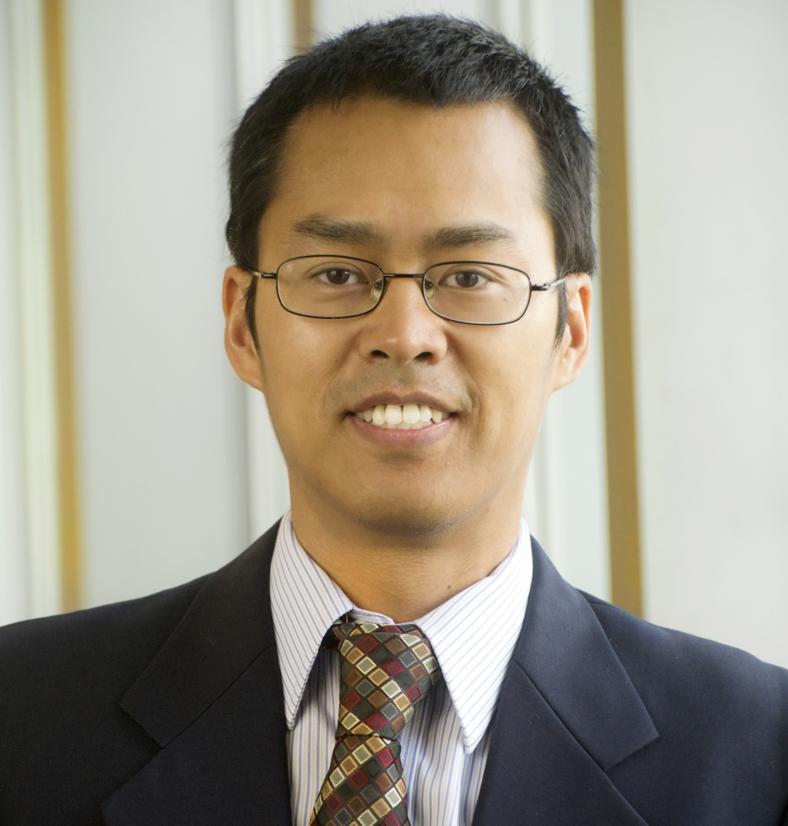}}]{Junzhou Huang}
is an Associate Professor in the Computer Science and Engineering department at the University of Texas at Arlington. He received the B.E. degree from Huazhong University of Science and Technology, China, the M.S. degree from Chinese Academy of Sciences, China, and the Ph.D. degree in Rutgers university. His major research interests include machine learning, computer vision and imaging informatics. He was selected as one of the 10 emerging leaders in multimedia and signal processing by the IBM T.J. Watson Research Center in 2010. He received the NSF CAREER Award in 2016.
\end{IEEEbiography}

%

%
%





\newpage
\title{Supplementary Materials\vspace{1ex} \\ \Large{Adaptive Cost-Sensitive Online Classification}}
\maketitle
\author{Peilin Zhao, Yifan Zhang, Min Wu, Steven C. H. Hoi, Mingkui Tan, and Junzhou Huang
\IEEEcompsocitemizethanks{
\IEEEcompsocthanksitem \blue{P. Zhao, Y. Zhang and M. Tan are with the South China University of Technology, China. E-mail: peilinzhao@hotmail.com; sezyifan@mail.scut.edu.cn; mingkuitan@scut.edu.cn.}
\IEEEcompsocthanksitem M. Wu is with the Institute for Infocomm Research, Singapore. E-mail: wumin@i2r.a-star.edu.sg.
\IEEEcompsocthanksitem S. C. Hoi is with the Singapore Management University, Singapore. E-mail: chhoi@smu.edu.sg.
\IEEEcompsocthanksitem J. Huang is with Tencent AI Lab, China. Email: joehhuang@tencent.com.
\IEEEcompsocthanksitem \blue{Y. Zhang is the co-first author; M. Tan is the corresponding author.}}
}



\IEEEtitleabstractindextext{
\begin{abstract}
This supplemental file provides the proofs of theorems and the related work in our paper of "Adaptive Cost-Sensitive Online Classification".
\end{abstract}

\begin{IEEEkeywords}
Cost-Sensitive Classification; Online Learning; Adaptive Regularization; Sketching Learning.
\end{IEEEkeywords}}



\IEEEdisplaynontitleabstractindextext

\begin{appendices}

\section{Proofs of Theorems}
This section presents the proofs for all the theorems.
\vspace{-0.1in}
\subsection{Proof of Theorem 1}

\textbf{Proof.} \ It is easy to verify $\mu_{t+1} = {\rm arg \min}_\mu h_t(\mu)$, where $h_t(\mu)= \frac{1}{2}||\mu_t-\mu||_{\Sigma^{-1}_{t+1}}^2 + \eta g_t^{\top} \mu$. Since $h_t$ is convex and continuous, through proof by contradiction, one can easily have:
\begin{align}
  \partial & h_t(\mu_{t+1})^\top (\mu-\mu_{t+1}) \nonumber \\
  =& [(\mu_{t+1}-\mu_t)^\top \Sigma_{t+1}^{-1} + \eta g_t^\top] (\mu-\mu_{t+1})\geq 0, \forall \mu.  \nonumber
\end{align}

Re-arrange the inequality will give:
\begin{align}
  (\eta g_t)^\top (\mu_{t+1}-\mu) \leq & (\mu_{t+1}- \mu_t)^\top \Sigma_{T+1}^{-1}(\mu-\mu_{t+1})  \nonumber \\
  = & \frac{1}{2}[||\mu_{t}-\mu||_{\Sigma^{-1}_{t+1}}^2 -||\mu_{t+1}-\mu||_{\Sigma^{-1}_{t+1}}^2 \nonumber \\
  &\ -||\mu_{t}-\mu_{t+1}||_{\Sigma^{-1}_{t+1}}^2].\nonumber
\end{align}

Then, since $\ell_t(\mu)$ is convex function, we have:
\begin{align}
  g_t^\top (\mu_{t+1}-\mu) = & g_t^\top (\mu_{t+1}-\mu+\mu_t-\mu_t)  \nonumber \\
  = & g_t^\top(\mu_{t}-\mu)+g_t^\top(\mu_{t+1}-\mu_t) \nonumber \\
  \geq & \ell_t(\mu_t) - \ell_t(\mu) + g_t^\top(\mu_{t+1}-\mu_t). \nonumber
\end{align}

Next, combining the above two inequalities gives the following key inequality:
\begin{align}
  \ell_t(\mu_t)-\ell_t(\mu) \leq & \frac{1}{2\eta}[||\mu_{t}-\mu||_{\Sigma^{-1}_{t+1}}^2 -||\mu_{t+1}-\mu||_{\Sigma^{-1}_{t+1}}^2  \nonumber \\
   & \ -||\mu_{t}-\mu_{t+1}||_{\Sigma^{-1}_{t+1}}^2] - g_t^\top(\mu_{t+1}-\mu_t). \nonumber
\end{align}

Summing the inequality above over $t =1,2,...T$, gives:
\begin{align}\label{1}
  & \sum_{t=1}^T [\ell_t(\mu_t)-\ell_t(\mu)]   \nonumber \\
   & \leq  \frac{1}{2\eta}\sum_{t=1}^T[||\mu_{t}-\mu||_{\Sigma^{-1}_{t+1}}^2 -||\mu_{t+1}-\mu||_{\Sigma^{-1}_{t+1}}^2]  \nonumber \\
   & \ -\frac{1}{2\eta} \sum_{t=1}^T||\mu_{t}-\mu_{t+1}||_{\Sigma^{-1}_{t+1}}^2] - \sum_{t=1}^Tg_t^\top(\mu_{t+1}-\mu_t).
\end{align}

Now, we bound the right side of inequality by dividing two parts. For the first term, we can bound as:
\begin{align}\label{2}
  & \sum_{t=1}^T [||\mu_{t}-\mu||_{\Sigma^{-1}_{t+1}}^2 -||\mu_{t+1}-\mu||_{\Sigma^{-1}_{t+1}}^2]   \nonumber \\
   & \leq  ||\mu_{1}-\mu||_{\Sigma^{-1}_{2}}^2 + \sum_{t=2}^T[||\mu_{t}-\mu||_{\Sigma^{-1}_{t+1}}^2 -||\mu_{t}-\mu||_{\Sigma^{-1}_{t}}^2]  \nonumber \\
   & = ||\mu_{1}-\mu||_{\Sigma^{-1}_{2}}^2 + \sum_{t=2}^T[||\mu_{t}-\mu||_{(\Sigma^{-1}_{t+1}-\Sigma^{-1}_{t})}^2] \nonumber \\
   & \leq ||\mu_{1}\small{-}\mu||^2 \lambda_{max}(\Sigma_2^{-1})+ \sum_{t=2}^T ||\mu_{t}\small{-}\mu||^2 \lambda_{max}(\Sigma^{-1}_{t+1}\small{-}\Sigma^{-1}_{t}) \nonumber \\
   & \leq ||\mu_{1}-\mu||^2 {\rm Tr}(\Sigma_2^{-1})+ \sum_{t=2}^T ||\mu_{t}-\mu||^2  {\rm Tr}(\Sigma^{-1}_{t+1}-\Sigma^{-1}_{t}) \nonumber \\
   & \leq \max_{t\leq T}||\mu_{t}\small{-}\mu||^2{\rm Tr}(\Sigma_2^{-1}) \small{+} \sum_{t=2}^T \max_{t\leq T}||\mu_{t}\small{-}\mu||^2  {\rm Tr}(\Sigma^{-1}_{t+1}\small{-}\Sigma^{-1}_{t}) \nonumber \\
   & =  \max_{t\leq T}||\mu_{t}\small{-}\mu||^2{\rm Tr}(\Sigma_{T+1}^{-1}),
\end{align}

where $\lambda_{max}(\Sigma)$ is the largest eigenvalue of $\Sigma$. Then, we bound the remain terms. From the updating rule of $\mu$, we have:
\begin{align}
  (\mu_{t+1}-\mu_t)^\top \Sigma_{t+1}^{-1} + \eta g_t^\top =0,   \nonumber
\end{align}

so that:
\begin{align}
   &||\mu_{t}-\mu_{t+1}||_{\Sigma^{-1}_{t+1}}^2   \nonumber \\
   &=(\mu_{t+1}-\mu_t)^\top \Sigma_{t+1}^{-1}\Sigma_{t+1}\Sigma_{t+1}^{-1}(\mu_{t+1}-\mu_t)   \nonumber \\
   &=\eta^2 g_t^\top \Sigma_{t+1}g_t,   \nonumber
\end{align}

and
\begin{align}\label{3}
   g_t^\top (\mu_{t+1}-\mu_t) = -\eta g_t^\top \Sigma_{t+1}g_t.
\end{align}

Combining the two inequalities causes in:

\begin{align}\label{4}
   &-\frac{1}{2\eta} \sum_{t=1}^T||\mu_{t}-\mu_{t+1}||_{\Sigma^{-1}_{t+1}}^2 - \sum_{t=1}^Tg_t^\top(\mu_{t+1}-\mu_t)   \nonumber \\
   &=\sum_{t=1}^T \eta g_t^\top \Sigma_{t+1}g_t-  \sum_{t=1}^T \frac{\eta}{2} g_t^\top \Sigma_{t+1}g_t \nonumber \\
   &= \frac{\eta}{2} \sum_{t=1}^T  g_t^\top \Sigma_{t+1}g_t.
\end{align}

As we know, we have $g_t = L_t y_t x_t$ for ACOG-I, where $L_t = \mathbb{I}_{(\ell_t(\mu_t)>0)}$ means that if $\ell_t(\mu_t)>0$, $L_t =1$; otherwise, $L_t =0$. Now, we can bound $\sum_{t=1}^Tg_t^\top \Sigma_{t+1} g_t$:

\begin{align}\label{5}
   \sum_{t=1}^Tg_t^\top \Sigma_{t+1} g_t = &\sum_{t=1}^T L_t x_t^\top \Sigma_{t+1} x_t=\gamma \sum_{t=1}^T(1-\frac{|\Sigma_t^{-1}|}{|\Sigma_{t+1}^{-1}|})    \nonumber \\
     \leq &  -\gamma \sum_{t=1}^T {\rm log}(\frac{|\Sigma_t^{-1}|}{|\Sigma_{t+1}^{-1}|}) \leq \gamma{\rm log}(|\Sigma_{T+1}^{-1}|) ,
\end{align}

where we used
\begin{align}
   \Sigma_{t+1}^{-1} = \Sigma_{t}^{-1}+\frac{L_t x_tx_t^{\top}}{\gamma} \Rightarrow \frac{L_t}{\gamma}x_t^\top \Sigma_{t+1} x_t = 1-\frac{|\Sigma_t^{-1}|}{|\Sigma_{t+1}^{-1}|}.    \nonumber
\end{align}

Combining Eq. (1-2) and Eq. (4-5), we can get:
\begin{align}
  Regret \leq \frac{1}{2\eta}(D_{\mu})^2{\rm Tr}(\Sigma_{T+1}^{-1})+\frac{\eta \gamma}{2}{\rm log}(|\Sigma_{T+1}^{-1}|), \nonumber
\end{align}

where $D_{\mu} = {\rm max}_t||\mu_t-\mu||$. Then, by setting $\eta=\sqrt{\frac{{\rm max}_{t\leq T}||\mu_t-\mu||^2{\rm Tr}(\Sigma_{T+1}^{-1})}{\gamma{\rm log}(|\Sigma_{T+1}^{-1}|)}}$, we can obtain the bound of ACOG-I.

For ACOG-II, we have $g_t = \rho L_t y_t x_t$, and then have:
\begin{align}\label{6}
   \sum_{t=1}^Tg_t^\top \Sigma_{t+1} g_t \small{=} \rho^2 \sum_{t=1}^T L_t x_t^\top \Sigma_{t+1} x_t \leq  \rho^2 \gamma{\rm log}(|\Sigma_{T+1}^{-1}|).
\end{align}

Combining Eq. (1-2), Eq. (4) and Eq. (6) will give:
\begin{align}
  Regret \leq \frac{1}{2\eta}(D_{\mu})^2{\rm Tr}(\Sigma_{T+1}^{-1})+\frac{\eta \rho^2 \gamma}{2}{\rm log}(|\Sigma_{T+1}^{-1}|), \nonumber
\end{align}

where $D_{\mu} = {\rm max}_t||\mu_t-\mu||$. Then, by setting $\eta=\sqrt{\frac{{\rm max}_{t\leq T}||\mu_t-\mu||^2{\rm Tr}(\Sigma_{T+1}^{-1})}{\rho^2\gamma{\rm log}(|\Sigma_{T+1}^{-1}|)}}$, we obtain ACOG-II's bound.

\subsection{Proof of Theorem 2}

\textbf{Proof.}  For ACOG with both loss function, if $t \in \mathcal{M}_p, \ell_t(\mu_t)\geq \rho$ and $t \in \mathcal{M}_n, \ell_t(\mu_t)\geq 1$. Then we have:
\begin{align}\label{7}
   \rho M_p+M_n \leq \sum_{t=1}^T\ell_t(\mu_t).
\end{align}

According to the definition of $sum$, we can obtain:
\begin{align}
   sum =&\alpha_p \times \frac{T_p-M_p}{T_p}+ \alpha_n \frac{T_n- M_n}{T_n}     \nonumber \\
   = & 1 \small{-} \frac{\alpha_n}{T_n}[\frac{\alpha_pT_n}{\alpha_nT_p}\sum_{y_t=+1}\mathbb{I}_{(y_t\mu \cdot x_t<0)}+\sum_{y_t=-1}\mathbb{I}_{(y_t\mu \cdot x_t<0)}] \nonumber \\
   = & 1 \small{-} \frac{\alpha_n}{T_n}(\frac{\alpha_pT_n}{\alpha_nT_p}M_p+M_n). \nonumber
\end{align}

Setting $\rho=\frac{\alpha_pT_n}{\alpha_nT_p}$, and combining with the regret bound of theorem 1 will conclude the proof.

\subsection{Proof of Theorem 3}

\textbf{Proof.} \ According to the definition of $cost$, we can obtain:
\begin{align}
   cost =&c_p M_p+ c_n M_n     \nonumber \\
   = & c_n[\frac{c_p}{c_n}\sum_{y_t=+1}\mathbb{I}_{(y_t\mu \cdot x_t<0)}+\sum_{y_t=-1}\mathbb{I}_{(y_t\mu \cdot x_t<0)}] \nonumber \\
   = & c_n(\frac{c_p}{c_n}M_p+M_n). \nonumber
\end{align}

Setting $\rho = \frac{c_p}{c_n}$, and combining it with Eq. (7), we obtain:
\begin{align}
   c_n(\rho M_p+M_n) \leq c_n \sum_{t=1}^{T}\ell_t(\mu_t). \nonumber
\end{align}

Then combining with the regret bound of theorem 1 will prove this theorem.

\section{Related Work} \label{Related}
Our work is related with three main categories of studies: (i) Cost-sensitive classification; (ii) Online learning; (iii) Sketching methods.
\vspace{-0.1in}
\subsection{Cost-Sensitive Classification}

Cost-sensitive classification has been widely studied in machine learning and data mining literatures\cite{Liu2006The,Zhou2006Training,Zhu2006Class,Zhao2013Cost,Sahoo2016Cost}. Many real classification problems, such as medical diagnosis and fraud detection, are naturally cost-sensitive. For these problems, the mistake cost of positive samples is much higher than that of negative samples, so that the optimal classifier under equalized cost setting often tends to underperform.

To address this issue, researchers have proposed a series of cost-sensitive metrics that take mistake cost into consideration when measuring classification performance. Well-known examples include the weighted sum of \emph{sensitivity} and \emph{specificity} \cite{Han2011Data,Brodersen2010the} and the weighted \emph{misclassification cost}\cite{Elkan2001The,Akbani2004Applying}. In special, when all class weights are equal to 0.5, the weighted sum is simplified as the well-known \emph{balanced accuracy}\cite{Brodersen2010The}, which is extensively used in real-world anomaly detection.

During the past decades, kinds of batch learning methods have been proposed for cost-sensitive classification in machine learning literatures\cite{Elkan2001The,Domingos1999A, Li2010Cost}. However, quite few studies consider online learning process where data arrives sequentially, except Cost-sensitive Passive Aggressive (CPA)\cite{Crammer2006Online}, Perceptron Algorithm with Uneven Margin (PAUM)\cite{Li2002The} and COG algorithms\cite{wang2012cost,wang2014cost}.

\subsection{Online Learning}

Online learning manages a sequence of samples with time series, which is quite common to see in real-world applications. For example at some moment, an anomaly detector receives a sample signal from natural world and then predicts its estimated class, i.e., normal or anomaly. After that, the detector receives the true class information and figures out the misclassification cost. Finally, the detector would update its model weights based on the suffered loss. Overall, the main goal of online learning is to minimize the cumulative loss over the whole sample sequence\cite{Hoi2014Libol}.

Online learning has been extensively studies in machine learning and data mining communities\cite{Crammer2003Ultraconservative, Cesa2004On,Zhang2017Projection, Zhao2010OTL, Zhao2011Online, Wu2017Onlined,Yan2017Online}, where a great variety of online approaches was proposed, including many first-order online algorithms\cite{Rosenblatt1958the,Crammer2006Online}. One of most famous first-order online algorithms is the perceptron algorithm\cite{Rosenblatt1958the,Freund1999Large}. Specifically, perceptron algorithm updates its model weights by adding the misclassified sample with a constant weight to the current support vector set. Recently, a number of online algorithms have been proposed based on the criterion of maximum margin\cite{Gentile2001A,Kivinen2002Online}. one famous example is the Relaxed Online Maximum Margin algorithm (named ROMMA)\cite{Li2000The}, which repetitively selects the hyperplane to classify all existing samples with the maximum margin. Another well-known method is the Passive Aggressive algorithm (named PA)\cite{Crammer2006Online}, which updates the classifier based on analytical solutions when an example is misclassified or its estimated score does not exceed the predefined margin. Moreover, the Perceptron Algorithm with Uneven Margins (named PAUM)\cite{Li2002The} also enjoys high attention, because it shows strong predictive abilities by producing decision hyperplanes with uneven margins. Many experimental studies show that online algorithms based on large margin are generally more effective than the classic Perceptron algorithm. Despite the difference, these online approaches only update model parameters based on the first order information, such as the gradient of training loss. This constraint could significantly limit the effectiveness of online algorithms.

Recently, a number of studies about second-order online algorithms have been proposed in machine learning and data mining literatures\cite{Cesa2005A, Wang2012Exact,Zhao2011Double, Dredze2008Confidence,Crammer2009Adaptive,Tan2016Learning}, which implies the confidence information of parameters can be adopted to guide the prediction of online algorithms. For example, the Second Order Perceptron algorithm\cite{Cesa2005A} is the first proposed second-order online learning approach based on the whitening transformation with the correlation matrices of previous seen samples. After that, many second-order online learning methods with large margin emerge. One famous method is the Confidence-weighted algorithm\cite{Dredze2008Confidence,Crammer2009Exact}, which maintains a multivariate Gaussian distribution of model parameters to manage the step and direction of model updates\cite{Dredze2008Confidence}. Although CW algorithm has theoretical guarantee of mistake bound\cite{Crammer2009Exact}, it may over fit training samples in certain situations because of the aggressive updating rule. To solve this issue, a modified algorithm, named as Adaptive Regularization of Weights algorithm (named AROW)\cite{Crammer2009Adaptive}, was recently proposed to relax such assumption. In detail, AROW adopts an adaptive regularization for prediction model, based on current confident information when seeing new samples. This regularization derives from minimizing a combination of the confidence penalty of vectors and the Kullback-Leibler divergence between Gaussian distributed weight vectors. As a result, it is robust to the sudden changes by the noise instance in the online learning process. However, although AROW improves the original CW algorithm by handling the noisy and non-separable samples, it is not the exact corresponding soft extension of CW (like PA-I and PA-II relative to PA algorithm). Particularly, the directly added loss and confidence regularization make AROW lose some important property of CW, i.e., Adaptive Margin property\cite{Crammer2009Exact}. Following the similar inspiration of soft margin support vector machine, the Soft Confidence-Weighted algorithm \cite{Wang2012Exact} is proposed to assign adaptive margins for diverse samples via a probability formulation, which improves CW algorithm to obtain extra effectiveness and efficiency. Generally, the second order online algorithms converge faster with more accurate predictions.

It is remarkable that most mentioned online learning algorithms are cost-insensitive, apart from PAUM\cite{Li2002The}, $CPA_{PB}$\cite{Crammer2006Online} and COG\cite{wang2012cost,wang2014cost}.

\vspace{-0.1in}
\subsection{Sketching Method}

In real-world applications, large-scale datasets are quite common. For these datasets, numbers of algorithms are difficult to implement, due to the extremely high computational cost. To address this problem, Sketching methods are designed to provide an efficient computational framework by obtaining compact approximations of large matrices, where the time complexity of reading samples is only linear in an artificial constant number, while the overall running time is linear in the dimensions of samples.

Sketching methods attract extensive attention in machine learning and data mining fields. One famous approach is to generate a sparser matrices, because sparser matrices can be stored more efficiently and be computed faster\cite{Achlioptas2007Fast}.

Another well-known family of methods for sketching is the Random Projection\cite{Vempala2004The, Sarlos2006Improved,Liberty2007Randomized,Achlioptas2003Database,Indyk1998Approximate,Kane2014Sparser}, which relies on the properties of random low dimensional subspaces and strong concentration of measure phenomena. Although Random Projection for Sketching algorithms have theoretical guarantees based on the rank of approximated matrix, it may perform terrible when the rank of approximated matrix is near full-rank\cite{luo2016efficient}.

To address this problem, researchers recently proposed the Frequent Direction Sketch\cite{Ghashami2016Frequent, Liberty2013Simple}, which is a class of deterministic methods that derives from the similarity comparison between matrix sketching problem and item frequency estimation problem. Empirical results show that the frequent direction sketch algorithms produce more accurate sketches than the widely used random projection approaches\cite{Ghashami2016Frequent, Liberty2013Simple}.

However, the regret bound of Frequent Direction Sketch depends on a super-parameter and a square root term, which are not controlled by the sketching algorithms\cite{luo2016efficient}. Thus, to better focus on the dominant part of the spectrum for deterministic sketching, an Oja's sketch algorithm was recently proposed\cite{luo2016efficient} based on Oja's algorithm\cite{Oja1982Simplified,oja1985stochastic}.

To the best of our knowledge, quite few existing sketching methods were applied in cost-sensitive online classification problems. However, This is a valuable research issue, because the computational efficiency is quite important for online learning in real-world applications.

\end{appendices}


\begin{thebibliography}{1}

\bibitem{Rosenblatt1958the}
F. Rosenblatt. The perceptron: A probabilistic model for information storage and organization in the brain. \emph{Psychological Review}, 1958, Vol. 65, No. 6, pp. 386.


\bibitem{Zhao2011Double}
P. Zhao, S. C. Hoi, R. Jin. Double updating online learning. \emph{Journal of Machine Learning Research}, 2011, Vol. 12, pp. 1587-1615.

\bibitem{Wang2012Exact}
J. Wang, P. Zhao, S. C. Hoi. Exact soft confidence-weighted learning. \emph{International Conference on Machine Learning}, 2012, pp. 107-114.

\bibitem{Wu2017Online}
Q. Wu, H. Wu, X. Zhou, M. Tan, Y. Xu, Y. Yan, T. Hao. Online Transfer Learning with Multiple Homogeneous or Heterogeneous Sources. \emph{IEEE Transactions on Knowledge and Data Engineering}, 2017, Vol. 29, No. 7, pp. 1494-1507.




\bibitem{Zhao2013Costd}
P. Zhao, S. C. Hoi. Cost-sensitive online active learning with application to malicious URL detection. \emph{ACM International Conference on Knowledge Discovery and Data Mining}, 2013, pp. 919-927.

\bibitem{Hoi2018Online}
\blue{S. C. Hoi, D. Sahoo, J. Lu, P. Zhao. Online Learning: A Comprehensive Survey, 2018. \emph{arXiv preprint arXiv}:1802.02871.}
%

\bibitem{Ma2011Learning}
J. Ma, L. K. Saul, S. Savage, G. M. Voelker. Learning to detect malicious urls. \emph{ACM Transactions on Intelligent Systems and Technology}, 2011, Vol. 2, No. 3, pp. 30.

\bibitem{Li2013Confidence}
B. Li, S. C. Hoi, P. Zhao, V. Gopalkrishnan. Confidence weighted mean reversion strategy for online portfolio selection. \emph{ACM Transactions on Knowledge Discovery from Data}, 2013, Vol. 7, No. 1, pp.4.

\bibitem{Shalev-Shwartz2011Pegasos}
S. Shalev-Shwartz, Y. Singer, N. Srebro, A. Cotter, Pegasos: Primal estimated sub-gradient solver for svm. \emph{Mathematical Programming}, 2011, Vol. 127, No. 1, pp. 3-30.


\bibitem{Elkan2001The}
C. Elkan. The foundations of cost-sensitive learning. \emph{International Joint Conference on Artificial Intelligence}, 2001, Vol. 17,  pp. 973-978.

\bibitem{Veropoulos1999Controling}
K. Veropoulos, C. Campbell, N. Cristianini. Controlling the sensitivity of support vector machines. \emph{International Joint Conference on Artificial Intelligence}, 1999, pp. 55-60.


\bibitem{He2009Learning}
H. He, E. A. Garcia. Learning from imbalanced data. \emph{IEEE Transactions on Knowledge and Data Engineering}, 2009. Vol. 21, pp. 1263-1284.


%

\bibitem{Han2011Data}
J. Han, J. Pei, M. Kamber. Data mining: Concepts and Techniques. \emph{Elsevier}, 2011.

\bibitem{Brodersen2010the}
K. H. Brodersen, C. S. Ong, K. E. Stephan, J. M. Buhmann. The balanced accuracy and its posterior distribution. \emph{International Conference on Pattern Recognition}, 2010, pp. 3121-3124.

\bibitem{Akbani2004Applying}
R. Akbani, S. Kwek, N. Japkowicz. Applying support vector machines to imbalanced datasets. \emph{European Conference on Machine Learning}, 2004, pp. 39-50.

\bibitem{wang2012cost}
J. Wang, P. Zhao and S. C. H. Hoi. Cost-sensitive online classification. \emph{IEEE Iternational Conference on Data Mining}, 2012, 1140-1145.

\bibitem{wang2014cost}
J. Wang, P. Zhao and S. C. H. Hoi. Cost-sensitive online classification. \emph{IEEE Transactions on Knowledge and Data Engineering}, vol. 26, no. 10, pp. 2425-2438.


\bibitem{Dredze2008Confidence}
M.Dredze, K.Crammer, F.Pereira.Confidence-weighted linear classification.\emph{International Conference on Machine learning}, 2008, 264-271.

\bibitem{Crammer2009Exact}
K. Crammer, M. Dredze, F. Pereira. Exact convex confidence-weighted learning. \emph{In Advances in Neural Information Processing Systems}, 2009, pp. 345-352.

\bibitem{Crammer2009Adaptive}
K. Crammer, A. Kulesza, M. Dredze. Adaptive regularization of weight vectors. \emph{In Advances in Neural Information Processing Systems}, 2009, pp. 414-422.

\bibitem{Zinkevich2003Online}
M. Zinkevich. Online convex programming and generalized infinitesimal gradient ascent. \emph{International Conference on Machine Learning}, 2003, pp. 928-936.

\bibitem{luo2016efficient}
H. Luo, A. Agarwal, N Cesa-Bianchi. Efficient second order online learning by sketching. \emph{In Advances in Neural Information Processing Systems}, 2016, pp. 902-910.

\bibitem{woodruff2014sketching}
Woodruff, P. David. Sketching as a tool for numerical linear algebra. \emph{Foundations and Trends in Theoretical Computer Science}, 2014, Vol. 10, No. 1-2, pp. 1-157.

\bibitem{Krummenacher2016Scalable}
G. Krummenacher, B. McWilliams, Y. Kilcher, J. M. Buhmann, N. Meinshausen. Scalable adaptive stochastic optimization using random projections. \emph{In Advances in Neural Information Processing Systems}, 2016, pp. 1750-1758.

\bibitem{Wang2014High}
D. Wang, P. Wu, P. Zhao, Y. Wu,  C. Miao, S. C. Hoi. High-dimensional data stream classification via sparse online learning. \emph{IEEE International Conference on Data Mining}, 2014, pp. 1007-1012.

\bibitem{zhao2015cost}
P. Zhao, F. Zhuang, M. Wu, X. Li, and S. C. H. Hoi. Cost-sensitive online classification with adaptive regularization and its applications. \emph{IEEE International Conference on Data Mining}, 2015, pp. 649-658.

%
%
%


\bibitem{Zhou2006Training}
Z. H. Zhou, X. Y. Liu. Training cost-sensitive neural networks with methods addressing the class imbalance problem. \emph{IEEE Transactions on Knowledge and Data Engineering}, 2006, Vol. 18, No. 1, pp. 63-77.

\bibitem{Horn1985matrix}
R. Horn. Matrix analysis. \emph{Cambridge University Express}, 1985.
%

\bibitem{Zhao2013Cost}
P. Zhao, S. C. Hoi. Cost-sensitive double updating online learning and its application to online anomaly detection. \emph{SIAM International Conference on Data Mining}, 2013, pp. 207-215.

\bibitem{Sahoo2016Cost}
D. Sahoo, S. C. Hoi, P. Zhao. Cost Sensitive Online Multiple Kernel Classification. \emph{Asian Conference on Machine Learning}, 2016, pp. 65-80.

%
%
%

%






\bibitem{Zhao2010OTL}
P. Zhao, S. C. Hoi. OTL: A framework of online transfer learning. \emph{International Conference on Machine Learning}, 2010, pp. 1231-1238.

\bibitem{Zhao2011Online}
P. Zhao, R. Jin, T. Yang, S. C. Hoi. Online AUC maximization. \emph{International Conference on Machine Learning}, 2011, pp. 233-240.

\bibitem{zhang2015E-tree}
P. Zhang, C. Zhou, P. Wang, B. J. Gao, X. Zhu, L. Guo. E-tree: An efficient indexing structure for ensemble models on data streams. \emph{IEEE Transactions on Knowledge and Data Engineering}, 2015, Vol. 27, No. 2, pp. 461-474.

\bibitem{zhang2015online}
Q. Zhang, P.Zhang, G. Long, W. Ding, C. Zhang, X. Wu. Online learning from trapezoidal data streams. \emph{IEEE Transactions on Knowledge and Data Engineering}, 2016, Vol. 28, No. 10, pp. 2709-2723.

%

\bibitem{Yan2017Online}
Y. Yan, Q. Wu, M. Tan, M. K. Ng, H. Min, I. W. Tsang. Online Heterogeneous Transfer by Hedge Ensemble of Offline and Online Decisions. \emph{IEEE Transactions on Neural Networks and Learning Systems}, 2017.

\bibitem{Crammer2006Online}
K. Crammer, O. Dekel, J. Keshet, S. Shalev-Shwartz, Y. Singer. Online passive-aggressive algorithms. \emph{Journal of Machine Learning Research}, 2006, pp. 551-585.

\bibitem{Zhang2017Strategy}
Y. Zhang, G. Shu, Y. Li. Strategy-updating depending on local environment enhances cooperation in prisoner¡¯s dilemma game. \emph{Applied Mathematics and Computation}, 2017, Vol. 301, pp. 224-232.


\bibitem{Freund1999Large}
Y. Freund, R. E. Schapire. Large margin classification using the perceptron algorithm. \emph{Machine learning}, 1999, Vol. 37, pp. 277-296.

%

\bibitem{Li2000The}
Y. Li, P. M. Long. The relaxed online maximum margin algorithm. \emph{In Advances in Neural Information Processing Systems}, 2000, 498-504.

\bibitem{Li2002The}
Y. Li, H. Zaragoza, R. Herbrich, J. Shawe-Taylor, J. Kandola. The perceptron algorithm with uneven margins. \emph{International Conference on Machine learning}, 2002, pp. 379-386.


%


%
%
%
%
%

%



\bibitem{Oja1982Simplified}
E. Oja. Simplified neuron model as a principal component analyzer. Journal of Mathematical biology, 1982, Vol. 15, No. 3, pp. 267-273.

\bibitem{oja1985stochastic}	
E. Oja, J. Karhunen. On stochastic approximation of the eigenvectors and eigenvalues of the expectation of a random matrix. \emph{Journal of Mathematical Analysis and Applications}, 1985, vol. 106, pp. 69-84.



%

\bibitem{Abernethy2008Optimal}
J. Abernethy, P. L. Bartlett, A. Rakhlin, A. Tewari. Optimal strategies and minimax lower bounds for online convex games. \emph{Annual Conference on Computational Learning Theory}, 2008.



\bibitem{Hardt2014The}
M. Hardt, E. Price, The noisy power method: A meta algorithm with application.  \emph{In Advances in Neural Information Processing Systems}, 2014, pp. 2861-2869.


%
%
%
%
%
\end{thebibliography}

\vspace{-0.1in}
\end{document}